\documentclass{article}

\usepackage[T1]{fontenc}    
\usepackage[hyphens]{url}
\usepackage[inline]{enumitem}
\usepackage[numbers,compress]{natbib}
\usepackage[utf8]{inputenc}
\usepackage[x11names]{xcolor}
\usepackage{amsfonts}
\usepackage{amsmath}
\usepackage{amssymb}
\usepackage{amsthm}
\usepackage{bbold}
\usepackage{bm}
\usepackage{booktabs}
\usepackage{cancel}
\usepackage{environ}
\usepackage{graphicx}
\usepackage{listings}
\usepackage{microtype}
\usepackage{nicefrac}
\usepackage{pgfplots}
\usepackage{subcaption}
\usepackage{tikz}
\usepackage{times}
\usepackage{wrapfig}

\usepackage{hyperref}

\usepackage[accepted]{icml2020}

\definecolor{strings}{rgb}{.624,.251,.259}
\definecolor{keywords}{rgb}{.224,.451,.686}
\definecolor{comment}{rgb}{.322,.451,.322}

\lstdefinelanguage{python}{
  keywords=[3]{Normal, Bernoulli, Beta, Categorical, Dirichlet,
  Exponential, MultivariateNormalFull, RandomVariable,
  DirichletProcess, Empirical, PointMass, Gamma,
  MAP, Inference, KLqp, HMC, SGLD, KLpq,
  VariationalInference, MonteCarlo, ConjugateInference, GANInference,
  rnn_cell, dirichlet_process, cond, body, generative_network,
  discriminative_network, Dense,
  evaluate, ppc, copy, dot, get_session},
  morecomment=[l]{\#},
  morecomment=[s]{"""}{"""},
  morestring=[b]',
  morestring=[b]",
  alsoletter={<>=-+/*},
  sensitive=true
}

\renewcommand{\texttt}[1]{\lstinline[basicstyle=\fontsize{9pt}{9.25pt}\selectfont\ttfamily]{#1}}

\DeclareMathOperator*{\bx}{{\mathbf{x}}}
\DeclareMathOperator*{\bM}{{\mathbf{m}}}

\DeclareMathOperator*{\bs}{{\mathbf d}}
\DeclareMathOperator*{\bsopt}{{{\mathbf d}^*}}
\DeclareMathOperator*{\btheta}{{\bm{\theta}}}
\DeclareMathOperator*{\bthetaref}{{\bm{\theta}}_{\normalfont \text{ref}}}

\DeclareMathOperator*{\R}{{\mathbb{R}}}

\makeatletter
\newsavebox{\measure@tikzpicture}
\NewEnviron{scaletikzpicturetowidth}[1]{%
  \def\tikz@width{#1}%
  \def\tikzscale{1}\begin{lrbox}{\measure@tikzpicture}%
  \BODY
  \end{lrbox}%
  \pgfmathparse{#1/\wd\measure@tikzpicture}%
  \edef\tikzscale{\pgfmathresult}%
  \BODY
}
\makeatother
\tikzstyle{every node}=[font=\scriptsize]

\icmltitlerunning{Likelihood-free Markov chain Monte Carlo with Amortized Approximate Ratio Estimators}

\begin{document}

\pdfinfo{
/Title (Likelihood-free Markov chain Monte Carlo with Amortized Approximate Ratio Estimators)
/Author (Joeri Hermans, Volodimir Begy, Gilles Louppe)
}

\twocolumn[
\icmltitle{Likelihood-free MCMC with Amortized Approximate Ratio Estimators}



\icmlsetsymbol{equal}{*}

\begin{icmlauthorlist}
\icmlauthor{Joeri Hermans}{ulg}
\icmlauthor{Volodimir Begy}{uwien}
\icmlauthor{Gilles Louppe}{ulg}
\end{icmlauthorlist}

\icmlaffiliation{ulg}{University of Li\`ege, Belgium}
\icmlaffiliation{uwien}{University of Vienna, Austria}

\icmlcorrespondingauthor{Joeri Hermans}{joeri.hermans@doct.uliege.be}

\icmlkeywords{Approximate Inference, Bayesian Statistics, Likelihood-free Inference, Machine Learning, Neural Networks}

\vskip 0.3in
]



\printAffiliationsAndNotice{}  

\begin{abstract}
  Posterior inference with an intractable likelihood is becoming an increasingly common task in scientific domains which rely on sophisticated computer simulations. Typically, these forward models do not admit tractable densities forcing practitioners to make use of approximations.
  This work introduces a novel approach to address the intractability of the likelihood and the marginal model. We achieve this by learning a flexible amortized estimator which approximates the likelihood-to-evidence ratio. We demonstrate that the learned ratio estimator can be embedded in \textsc{mcmc} samplers to approximate likelihood-ratios between consecutive states in the Markov chain, allowing us to draw samples from the intractable posterior. Techniques are presented to improve the numerical stability and to measure the quality of an approximation. The accuracy of our approach is demonstrated on a variety of benchmarks against well-established techniques. Scientific applications in physics show its applicability.
\end{abstract}

\section{Introduction}
\label{sec:introduction}
Domain scientists are generally interested in the posterior
\begin{equation}
p(\btheta\vert\bx) = \frac{p(\btheta)p(\bx\vert\btheta)}{p(\bx)}
\end{equation}
which relates the parameters $\btheta$ of a model or theory to observations $\bx$. Although Bayesian inference is an ideal tool for such settings, the implied computation is generally not. Often the marginal model $p(\bx)=\int p(\btheta) p(\bx|\btheta) d\btheta$ is intractable, making posterior inference using Bayes' rule impractical. Methods such as Markov chain Monte Carlo (\textsc{mcmc})~\cite{metropolis,hastings1970monte} bypass the dependency on the marginal model by evaluating the ratio of posterior densities between consecutive states in the Markov chain. This allows the posterior to be approximated numerically, provided that the likelihood $p(\bx\vert\btheta)$ and the prior $p(\btheta)$ are tractable.
We consider the equally common and more challenging setting, the so-called likelihood-free setup, in which the likelihood cannot be evaluated in a reasonable amount of time or has no tractable closed-form expression. However, drawing samples from the forward model is possible.

\paragraph{Contributions} We introduce a Bayesian inference algorithm for scientific applications where \begin{enumerate*}[label=(\roman*)]\item a forward model is available, \item the likelihood is intractable, and \item accurate approximations are important to do science.\end{enumerate*} Central to this work is a novel amortized likelihood-to-evidence ratio estimator which allows for the direct estimation of the posterior density function
for arbitrary model parameters $\btheta\sim p(\btheta)$ and observations $\bx\sim p(\bx\vert\btheta)$. 
We exploit this ability to amortize the estimation of acceptance ratios in \textsc{mcmc}, enabling us to draw posterior samples. Finally, we develop a necessary diagnostic to probe the quality of the approximations in intractable settings.


\section{Background}
\label{sec:background}
\subsection{Markov chain Monte Carlo}
\label{sec:mcmc}
\textsc{mcmc} methods are generally applied to sample from a posterior probability distribution with an intractable marginal model, but for which point-wise evaluations of the likelihood are possible~\cite{metropolis,hastings1970monte,mackay2003information}.
Posterior samples are drawn from the target distribution by collecting dependent states $\btheta_{0:T}$ of a Markov chain. The mechanism for transitioning from $\btheta_t$ to the next state $\btheta'$ depends on the algorithm at hand. However, the acceptance of a transition $\btheta_t \xrightarrow{} \btheta'$, for $\btheta'$ sampled from a proposal mechanism $q(\btheta'\vert\btheta_t)$, is usually determined by evaluating some form of the
posterior ratio
\begin{align}
  \label{eq:mcmc_posterior_ratio}
\frac{p(\btheta'|\bx)}{p(\btheta_t|\bx)} &= \frac{p(\btheta')p(\bx|\btheta') \,/\, p(\bx)}{p(\btheta_t)p(\bx|\btheta_t) \,/\, p(\bx)} = \frac{p(\btheta')p(\bx|\btheta')}{p(\btheta_t)p(\bx|\btheta_t)}.
\end{align}
We observe that \begin{enumerate*}[label=(\roman*)]
  \item the normalizing constant $p(\bx)$ cancels out within the ratio, thereby bypassing its intractable evaluation, and \item
  how the likelihood ratio is central in assessing the quality of a candidate state $\btheta'$ against state $\btheta_t$.
\end{enumerate*}
\paragraph{Metropolis-Hastings} Metropolis-Hastings (\textsc{mh})~\cite{metropolis,hastings1970monte} is a straightforward implementation of Equation~\ref{eq:mcmc_posterior_ratio} in which the proposal mechanism  $q(\btheta'\vert\btheta_t)$ is typically a tractable distribution. These components are combined to compute the acceptance probability $\rho$ of a transition $\btheta_t\rightarrow\btheta'$:
\begin{equation}
  \label{eq:acceptance_ratio_mh}
  \rho = \min\left(1,~\frac{p(\btheta')p(\bx\vert\btheta')}{p(\btheta_t)p(\bx\vert\btheta_t)}\frac{q(\btheta'\vert\btheta_t)}{q(\btheta_t\vert\btheta')}\right).
\end{equation}
The choice of an appropriate transition distribution is important to maximize the effective sample size (sampling efficiency) and to reduce the autocorrelation.

\paragraph{Hamiltonian Monte Carlo} Hamiltonian Monte Carlo (\textsc{hmc})~\cite{hmc,hmcoriginal,hmcsummary}
improves upon the sampling efficiency of Metropolis-Hastings by reducing the
autocorrelation of the Markov chain.
This is achieved by modeling the density $p(\bx \vert \btheta)$ as a potential energy function
\begin{equation}
  U(\btheta) \triangleq -\log p(\bx \vert \btheta),
\end{equation}
and attributing some kinetic energy,
\begin{equation}
    K(\bM) \triangleq \frac{1}{2}{\bM}^2
\end{equation}
with momentum $\bM \sim p(\bM)$ to the current state $\btheta_t$. A new state $\btheta'$ can be proposed by simulating
the Hamiltonian dynamics of $\btheta_t$. This is achieved by leapfrog integration of $\nabla_{\btheta}~U(\btheta)$ over a fixed number of steps
with initial momentum $\bM$. Afterwards, the acceptance ratio
\begin{equation}
  \min\left(1,~\exp\left(U({\btheta}') - U({\btheta}_t) + K({\bM}') - K(\bM)\right)\right)
\end{equation}
is computed to assess the quality of the candidate state $\btheta'$.
\subsection{Approximate likelihood ratios}
\label{sec:approximate_lr}
The most powerful test-statistic to compare two hypotheses $\btheta_0$ and $\btheta_1$ for an observation $\bx$ is the likelihood ratio~\cite{neymanpearson}
\begin{equation}
  r(\bx\vert{\btheta}_0,{\btheta}_1) \triangleq \frac{p(\bx\vert{\btheta}_0)}{p(\bx\vert{\btheta}_1)}.
\end{equation}
\citet{approxlr} have shown that it is possible to express the test-statistic through a change of variables $\bs(\cdot)\colon \R^d \mapsto [0, 1]$.
This observation can be used in a supervised learning setting to train a classifier $\bs(\bx)$ to distinguish samples $\bx \sim p(\bx \vert \btheta_{0})$ with class label $y=1$
from $\bx \sim p(\bx \vert \btheta_{1})$ with class label $y=0$.
The decision function modeled by the optimal classifier $\bsopt(\bx)$ is in this case
\begin{equation}
  \label{eq:ratio}
  \bsopt(\bx) = p(y=1\vert\bx) = \frac{p(\bx \vert \btheta_{0})}{p(\bx \vert \btheta_{0}) + p(\bx \vert \btheta_{1})},
\end{equation}
thereby obtaining the likelihood ratio as
\begin{equation}
  r(\bx\vert~{\btheta}_0,{\btheta}_1) = \frac{\bsopt(\bx)}{1 - \bsopt(\bx)}.
\end{equation}
In the literature, this is known as the likelihood ratio trick (\textsc{lrt})~\cite{approxlr, 2016arXiv161003483M, classifierabc,dutta2016likelihood,tran2017hierarchical,brehmer2018mining} and is especially prominent in the area of Generative Adversarial Networks (\textsc{gan}s)~\cite{goodfellow2014generative,uehara2016generative,turner2018metropolis,azadi2018discriminator}.

Often we are interested in computing the likelihood ratio between many arbitrary hypotheses. Training $\bs(\bx)$ for every possible pair of hypotheses becomes impractical. A solution proposed by~\cite{approxlr,Baldi:2016fzo} is to parameterize the classifier $\bs$ with $\btheta$ (typically by injecting $\btheta$ as a feature) and train $\bs(\bx, \btheta)$ to distinguish between samples from $p(\bx\vert\btheta)$ and samples from an arbitrary but fixed reference hypothesis $p(\bx\vert\bthetaref)$.
In this setting, the decision function modeled by the optimal classifier~\cite{approxlr} is
\begin{equation}
    \label{eq:ratio_2}
    \bsopt(\bx, \btheta) = \frac{p(\bx \vert \btheta)}{p(\bx \vert \btheta) + p(\bx \vert \bthetaref)},
\end{equation}
thereby defining the likelihood-to-reference ratio as
\begin{equation}\label{eq:ratio_3}
  r(\bx\vert\btheta)\triangleq r(\bx\vert\btheta,\bthetaref) = \frac{\bsopt(\bx,\btheta)}{1-\bsopt(\bx,\btheta)}.
\end{equation}
Subsequently, the likelihood ratio between arbitrary hypotheses $\btheta_0$ and $\btheta_1$ can then be expressed as
\begin{equation}
  r(\bx\vert~{\btheta}_0,{\btheta}_1) = \frac{r(\bx\vert~{\btheta}_0)}{r(\bx\vert~{\btheta}_1)}.
\end{equation}

\section{Method}
\label{sec:alr_mcmc}
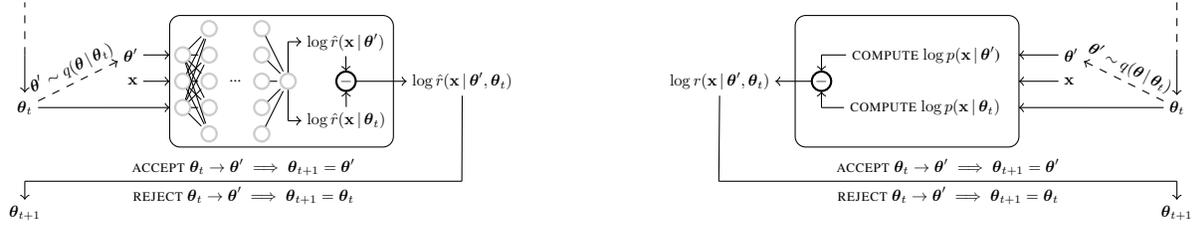
\begin{figure*}[t]
   \centering
   \usetikzlibrary{arrows}
   \def\layersep{1cm}
   \begin{subfigure}{.48\textwidth}
     \centering
   \begin{tikzpicture}[shorten >= 1pt, ->, node distance=\layersep,scale=0.35, every node/.style={scale=.61}]
   \tikzstyle{neuron} = [circle, minimum size=0.25cm, draw=black!20, line width=0.3mm, fill=white]
   \draw[rounded corners] (-1,-0.5) rectangle (7.5,-5.5);
   \path[->, shorten >= 0pt] (-2,-2) edge (-1,-2);
   \path[->, shorten >= 0pt] (-2,-3) edge (-1,-3);
   \node[left] at (-2,-2) (thetahat) {$\btheta'$};
   \node[left] at (-2,-3) {$\bx$};
   \node[right] at (4,-1.5) (R_next) {$\log\hat{r}(\bx\vert\btheta')$};
   \node[right] at (4,-4.5) (R_current) {$\log\hat{r}(\bx\vert\btheta_t)$};
   \node[right] at (8, -3) (R_out) {$\log\hat{r}(\bx\vert\btheta',\btheta_t)$};
   \node[circle, draw=black, thick, inner sep=0, minimum size=12pt] at (5.7,-3) (minus) {$-$};
   \path[-, shorten >= 0pt] (3.5, -1.5) edge (3.5, -3);
   \path[-, shorten >= 0pt] (3.5, -1.5) edge (3.5, -4.5);
   \path[->, shorten >= 0pt] (3.5, -1.5) edge (R_next);
   \path[->, shorten >= 0pt] (3.5, -4.5) edge (R_current);
   \path[->, shorten >= 0pt] (R_next) edge (minus);
   \path[->, shorten >= 0pt] (R_current) edge (minus);
   \path[->, shorten >= 0pt] (minus) edge (R_out);
   \path[-, shorten >= 0] (R_out) edge (10.1, -6.8);
   \path[-, shorten >= 0] (10.1, -6.8) edge node[midway, style={anchor=south,auto=false}] {\textsc{accept $\btheta_t\rightarrow\btheta'\implies \btheta_{t + 1} = \btheta'$}} node[midway, style={anchor=north,auto=false}] {\textsc{reject $\btheta_t\rightarrow\btheta'\implies \btheta_{t + 1} = \btheta_t$}} (-6.5, -6.8);
   \node[below] at (-6.5, -7.5) (thetanext) {$\btheta_{t+1}$};
   \path[->, shorten >= 0] (-6.5, -6.8) edge (thetanext);
   \node at (-6.5, -4) (mc_thetat) {$\btheta_t$};
   \path[->, shorten >=0, dashed] (mc_thetat) edge node[midway, color=black, style={sloped,anchor=south,auto=false}] {$\btheta'\sim q(\btheta\vert\btheta_t)$} (thetahat);
   \path[->, shorten >=0] (mc_thetat) edge (-1,-4);
   \path[->] (-6.5, -2) edge (mc_thetat);
   \path[-,dashed, shorten >= 0] (-6.5,-2) edge (-6.5,0);
   \foreach \name / \y in {1,...,3}
       \node[neuron] (f-I-\name) at (-0.5,-1-\y) {};
   \foreach \name / \y in {1,...,5}
       \node[neuron] (f-H1-\name) at (-0.5cm+\layersep,-\y cm) {};
   \foreach \name / \y in {1,...,5}
       \node[neuron] (f-H2-\name) at (-0.5cm+3*\layersep,-\y cm) {};
   \node[neuron] (f-O) at (-0.5cm+4*\layersep,-3cm) {};
   \foreach \source in {1,...,3}
       \foreach \dest in {1,...,5}
           \path[-, black] (f-I-\source) edge (f-H1-\dest);
   \foreach \source in {1,...,5}
       \path[-, black] (f-H2-\source) edge (f-O);
   \node[black] at (1.5,-3) {...};
   \end{tikzpicture}
   \caption{\textsc{aalr-mcmc} does not have to evaluate the likelihood, but instead computes an approximation of the likelihood ratio.}
   \end{subfigure}
   \hfill
   \begin{subfigure}{.48\textwidth}
     \centering
     \begin{tikzpicture}[shorten >= 1pt, ->, node distance=\layersep,scale=0.35, every node/.style={scale=.61}]
       \path[-, shorten >= 0pt, dashed] (0,0) edge (0, -2);
       \node at (0, -4) (mc_thetat) {$\btheta_t$};
       \path[->, shorten >= 0pt] (0,-2) edge (mc_thetat);
       \node[right] at (-4.5,-2) (thetahat) {$\btheta'$};
       \path[->, shorten >=0, dashed] (mc_thetat) edge node[midway, color=black, style={sloped,anchor=south,auto=false}] {$\btheta'\sim q(\btheta\vert\btheta_t)$} (thetahat);
       \draw[rounded corners] (-6,-0.5) rectangle (-14.5,-5.5);
       \node[right] at (-4.5,-3) (bx) {$\bx$};
       \path[->, shorten >= 0] (thetahat) edge (-6, -2);
       \path[->, shorten >= 0] (bx) edge (-6,-3);
       \path[->, shorten >= 0] (mc_thetat) edge (-6,-4);
       \node[left] at (-6.5,-2) (likelihoodhat) {\textsc{compute} $\log p(\bx\vert\btheta'$)};
       \node[left] at (-6.5,-4) (likelihoodcurrent) {\textsc{compute} $\log p(\bx\vert\btheta_t)$};
       \node[circle, inner sep=0, minimum size=12pt, thick, draw=black] at (-13.5,-3) (minus) {$-$};
       \path[->, shorten >= 0] (-13.5, -2) edge (minus);
       \path[->, shorten >= 0] (-13.5, -4) edge (minus);
       \path[-, shorten >= 0] (likelihoodhat) edge (-13.5, -2);
       \path[-, shorten >= 0] (likelihoodcurrent) edge (-13.5,-4);
       \node[left] at (-15.25,-3) (r_out) {$\log r(\bx\vert\btheta',\btheta_t)$};
       \path[->, shorten >= 0] (minus) edge (r_out);
       \path[-, shorten >= 0] (r_out) edge (-17.4,-6.8);
       \path[-, shorten >= 0] (-17.4,-6.8) edge node[midway, style={anchor=south,auto=false}] {\textsc{accept $\btheta_t\rightarrow\btheta'\implies \btheta_{t + 1} = \btheta'$}} node[midway, style={anchor=north,auto=false}] {\textsc{reject $\btheta_t\rightarrow\btheta'\implies \btheta_{t + 1} = \btheta_t$}} (0,-6.8);
       \node[below] at (0,-7.5) (final) {$\btheta_{t+1}$};
       \path[->, shorten >=0] (0,-6.8) edge (final);
     \end{tikzpicture}
     \caption{Vanilla \textsc{mcmc} computes the likelihood(s) whenever a transition needs to be assessed.}
   \end{subfigure}
   \caption{Overview showing {\it(a)} the proposed method \textsc{aalr-mcmc} and {\it(b)} traditional \textsc{mcmc} when evaluating the transition from the current state $\btheta_t$ to a candidate state $\btheta' \sim q(\btheta\vert\btheta_t)$. Both methods rely
     on the acceptance ratio as a test-statistic to evaluate the quality of the proposed transition $\btheta_t\rightarrow\btheta'$.
     \textsc{aalr-mcmc} does not depend on the evaluation of the (intractable) likelihood. Rather, it relies on an amortized estimator (Section~\ref{sec:improve_approx_lr}) to approximate the likelihood ratio $r(\bx\vert\btheta',\btheta_t)$.}
   \label{fig:method}
\end{figure*}
We propose a method to draw samples from a posterior with an intractable likelihood and marginal model. As noted earlier, \textsc{mcmc} samplers rely on the likelihood ratio to compute the acceptance ratio. We propose to remove the dependency on the intractable likelihoods $p(\bx\vert\btheta')$ and $p(\bx\vert\btheta_t)$ by directly modeling their ratio using an amortized ratio estimator $\hat{r}(\bx\vert\btheta',\btheta_t)$. We call this method amortized approximate likelihood ratio \textsc{mcmc} (\textsc{aalr-mcmc}). Figure~\ref{fig:method} provides a schematic overview of the proposed method.

\paragraph{Likelihood-free Metropolis-Hastings} Adapting \textsc{mh} to the likelihood-free setup is achieved by replacing the computation of the intractable likelihood ratio in Equation~\ref{eq:acceptance_ratio_mh} with $\hat{r}(\bx\vert\btheta',\btheta_t)$. The algorithm remains otherwise unchanged. We summarize the likelihood-free Metropolis-Hastings sampler in Appendix~\ref{sec:algorithms}.

\paragraph{Likelihood-free Hamiltonian Monte Carlo} The first step in making \textsc{hmc} likelihood-free,
is by showing that $U({\btheta}_t) - U(\btheta')$ reduces to the log-likelihood ratio,
\begin{align}
  \begin{split}
  U({\btheta}_t) - U({\btheta}') &= \log p(\bx \vert\btheta{}') - \log p(\bx \vert\btheta{}_t) \\
  &= \log r(\bx\vert\btheta{}', {\btheta}_t).
  \end{split}
\end{align}
To simulate the Hamiltonian dynamics of $\btheta_{t}$, we require a likelihood-free definition of $\nabla_{\btheta}~U(\btheta)$. Within our framework, $\nabla_{\btheta}~U(\btheta)$ can be expressed as
\begin{equation}
  \label{eq:gradU}
  \nabla_{\btheta}~U(\btheta) = -
  \frac{\displaystyle
    \nabla_{\btheta}~r(\bx\vert\btheta)
  }{\displaystyle
    r(\bx\vert\btheta)
  }.
  \end{equation}
This form can be recovered by a differentiable $\bsopt(\bx, \btheta)$, as expanding $r(\bx\vert\btheta)$ in Equation~\ref{eq:gradU} yields
\begin{equation}
  -\frac{\nabla_{\btheta}~r(\bx\vert\btheta)}{r(\bx\vert\btheta)}=-\nabla_{\btheta}~\log p(\bx\vert\btheta).
\end{equation}
Having likelihood-free alternatives for $U(\btheta) - U(\btheta')$ and $\nabla_{\btheta}~U(\btheta)$, we can
replace these components in \textsc{hmc} to obtain a likelihood-free \textsc{hmc} sampler. This procedure is summarized in Appendix~\ref{sec:algorithms}. While likelihood-free \textsc{hmc} does not rely on the intractable likelihood, it still depends on the computation of $\nabla_{\btheta}~\hat{r}(\bx\vert\btheta)$ to recover $\nabla_{\btheta}~U(\btheta)$. This can be a costly operation depending on the size of the ratio estimator. Similar to \textsc{hmc}, the sampler requires careful tuning to maximize the sampling efficiency.

\subsection{Improving the ratio estimator $\hat{r}$}
\label{sec:improve_approx_lr}
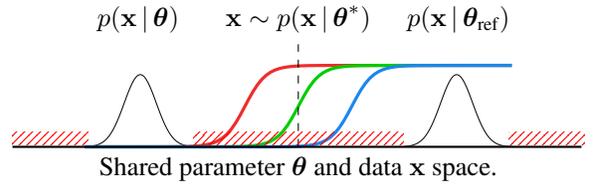
\begin{figure}
  \centering
  \usetikzlibrary{patterns}
  \begin{tikzpicture}[shorten >= 1pt, ->, node distance=\layersep,scale=0.2, every node/.style={scale=1}]
  \pgfmathdeclarefunction{gauss}{2}{\pgfmathparse{1/(#2*sqrt(2*pi))*exp(-((x-#1)^2)/(2*#2^2))}}
\begin{axis}[no markers, domain=-20:20, samples=100,
axis lines*=left,
height=7cm, width=30cm,
at={(-1.72\linewidth,0)},
axis line style={draw=none},
xtick=\empty, ytick=\empty,
enlargelimits=false, clip=false, axis on top,
every axis plot/.append style={ultra thick},
grid = major]
\addplot[black, domain=-20:-10] {gauss(-14.8,1.5)} \closedcycle;
\addplot[black, domain=10:20] {gauss(14.8,1.5)} \closedcycle;
\addplot%
        [
            -,
            color=Firebrick2,
            line width=6pt,
            mark=none,
            samples=100,
            domain=-20:20,
        ]
        (x,{0.3/((1+exp(-x - 5)))});
\addplot%
        [
            -,
            color=Green3,
            line width=6pt,
            mark=none,
            samples=100,
            domain=-20:20,
        ]
        (x,{0.3/((1+exp(-x)))});
\addplot%
        [
            -,
            color=DodgerBlue2,
            line width=6pt,
            mark=none,
            samples=100,
            domain=-20:20,
        ]
        (x,{0.3/((1+exp(-x + 5)))});
\end{axis}
  \draw[-,thick] (-19,0) edge node[midway, style={anchor=north,auto=false}] {Shared parameter $\btheta$ and data $\bx$ space.} (19,0);
  \node[above] at (0,7) (topmiddle) {$\bx\sim p(\bx\vert\btheta^*)$};
  \draw[-, dashed] (0, 0) edge (topmiddle);
  \node[above] at (-10.64, 7) (topleft) {$p(\bx\vert\btheta)$};
  \node[above] at (10.64,7) (topright) {$p(\bx\vert\bthetaref)$};
  \path [pattern=north east lines,thick,pattern color=red] (-19,.1) rectangle (-14,1);
  \path [pattern=north east lines,thick,pattern color=red] (19,.1) rectangle (14,1);
  \path [pattern=north east lines,thick,pattern color=red] (-7,.1) rectangle (7,1);
  \end{tikzpicture}
  \caption{Consider having access to an optimal classifier $\bsopt(\bx,\btheta)$ modeling $r(\bx\vert\btheta)$ with $\bx\sim p(\bx\vert\btheta^*)$. This ratio is undefined for $\bx$ as neither $p(\bx\vert\btheta)$ nor $p(\bx\vert\bthetaref)$ puts numerically non-negligible density on $\bx$. This implies that $\hat{r}(\bx\vert\btheta)$ and its decision function $\bsopt(\bx,\btheta)$ can take on arbitrary values in regions not covered by $p(\bx\vert\btheta)$ or $p(\bx\vert\bthetaref)$ (striped areas) because no such training data exists. The red, green and blue lines depict optimal decision functions as they all minimize the criterion which captures the ability to classify between samples from $p(\bx\vert\btheta)$ and $p(\bx\vert\bthetaref)$. However, the functions have different approximations of $\hat{r}(\bx\vert\btheta)$.}
  \label{fig:failure_carl}
\end{figure}

Simply relying on the amortized likelihood-to-reference ratio estimator $\hat{r}$ 
does not yield satisfactory results, even when considering simple toy problems.
Experiments indicate that the choice of the mathematically arbitrary reference hypothesis $\bthetaref$ does have a significant effect on the approximated likelihood ratios in practice. Other independent studies~\cite{dutta2016likelihood} observe similar issues and also conclude that the reference hypothesis $\bthetaref$ is a sensitive hyper-parameter which requires careful tuning for the problem at hand.
We find that poor inference results occur in the absence of support between $p(\bx\vert\btheta)$ and $p(\bx\vert\bthetaref)$, as illustrated in Figure~\ref{fig:failure_carl}. In this example, the evaluation of the approximate ratio $\hat{r}$ for an observation $\bx\sim p(\bx\vert\btheta^*)$ is undefined when the observation $\bx$ does not have density in $p(\bx\vert\btheta)$ and $p(\bx\vert\bthetaref)$, or either of the densities is numerically negligible.
Therefore, the decision function modeled by the optimal classifier $\bs(\bx,\btheta)$ outside of the space covered by $p(\bx\vert\btheta)$ and $p(\bx\vert\bthetaref$) is undefined.
Practically, this implies that the ratio $\hat{r}(\bx\vert\btheta)$ can take on an arbitrary value which is detrimental to the inference procedure because multiple solutions for $\bsopt(\bx,\btheta)$ exist.

To overcome the issues associated with a fixed reference hypothesis, we propose to train the parameterized classifier to distinguish dependent sample-parameter pairs $(\bx,\btheta) \sim p(\bx, \btheta)$ with class label $y=1$ from independent sample-parameter pairs $(\bx,\btheta) \sim p(\bx)p(\btheta)$ with class label $y=0$. This modification results in the optimal classifier 
\begin{equation}
    \bsopt(\bx, \btheta) = \frac{p(\bx, \btheta)}{p(\bx, \btheta) + p(\bx)p(\btheta)},
\end{equation}
and thereby in the likelihood-to-evidence ratio 
\begin{equation}
  \frac{\bsopt(\bx,\btheta)}{1-\bsopt(\bx,\btheta)} = \frac{p(\bx,\btheta)}{p(\bx)p(\btheta)} = \frac{p(\bx\vert\btheta)}{p(\bx)} = r(\bx\vert\btheta).
\end{equation}
This formulation ensures that the likelihood-to-evidence ratio will always be defined everywhere it needs to be evaluated, as the joint $p(\bx,\btheta)$ is consistently supported by the product of marginals $p(\bx)p(\btheta)$.

We summarize the procedure for learning the classifier $\bsopt(\bx,\btheta)$ and the corresponding ratio estimator $\hat{r}(\bx\vert\btheta)$ in Algorithm~\ref{algo:optimization}.
The algorithm amounts to the minimization of the binary cross-entropy (\textsc{bce}) loss of a classifier $\bs_\phi$.
We provide a proof in Appendix~\ref{sec:proof} that demonstrates that it results in the optimal discriminator $\bsopt$.
\begin{algorithm}[t]
  \caption{Optimization of $\bs_\phi(\bx, \btheta)$.}
  \label{algo:optimization}
  \begin{tabular}{ l l }
    {\it Inputs:} & Criterion $\ell$ (e.g., \textsc{bce}) \\
                  & Implicit generative model $p(\bx \vert \btheta)$ \\
                  & Prior $p(\btheta)$ \\
    {\it Outputs:} & Parameterized classifier $\bs_\phi(\bx, \btheta)$ \\
    {\it Hyperparameters:} & Batch-size $M$ \\
  \end{tabular}
  \\
  \begin{algorithmic}[1]
    \While{\bf not converged}
    \State Sample $\btheta \gets \{\btheta_m \sim p(\btheta)\}_{m=1}^{M}$
    \State Sample ${\btheta}^{'} \gets \{{\btheta}_m^{'} \sim p(\btheta)\}_{m=1}^{M}$
    \State Simulate $\bx \gets \{{\bx}_m \sim p(\bx\vert\btheta_m)\}_{m=1}^{M}$
    \State $\mathcal{L} \gets \ell(\bs_\phi(\bx,\btheta),~1) + \ell(\bs_\phi({\bx},{\btheta}{'}),~0)$
    \State $\phi \gets \textsc{optimizer}(\phi,~\nabla_\phi\mathcal{L})$
    \EndWhile
    \State \Return{$\bs_\phi$}
  \end{algorithmic}
\end{algorithm}

Although the usage of the marginal model instead of an arbitrary reference hypothesis vastly improves the accuracy of $\hat{r}(\bx\vert\btheta)$, obtaining the likelihood-to-evidence ratio $\hat{r}(\bx\vert\btheta)$ by transforming the output of $\bs(\bx,\btheta)$ can still be susceptible to numerical errors. This may happen in the saturating regime where the classifier $\bs(\bx,\btheta)$ is able to (almost) perfectly discriminate samples from $p(\bx\vert\btheta)$ and $p(\bx)$. We prevent this issue by extracting $\log \hat{r}(\bx\vert\btheta)$ from the neural network before applying the sigmoidal projection in the output layer, since $\log\hat{r}(\bx\vert\btheta)$ is the logit of $\bs(\bx,\btheta)$. This choice also mitigates a vanishing gradient when computing  $\nabla_{\btheta}\log\hat{r}(\bx\vert\btheta)$ or $\nabla_{\bx}\log\hat{r}(\bx\vert\btheta)$.

Finally, approximating the likelihood-to-evidence ratio also enables the direct estimation of the posterior density as $\hat{p}(\btheta\vert\bx) = p(\btheta)\hat{r}(\bx\vert\btheta)$. This is useful in low-dimensional model parameter spaces, where scanning is a reasonable strategy.

\subsection{Receiver operating curve diagnostic}
\label{sec:roc}
\begin{figure}[t!]
  \centering
  \includegraphics[width=\linewidth]{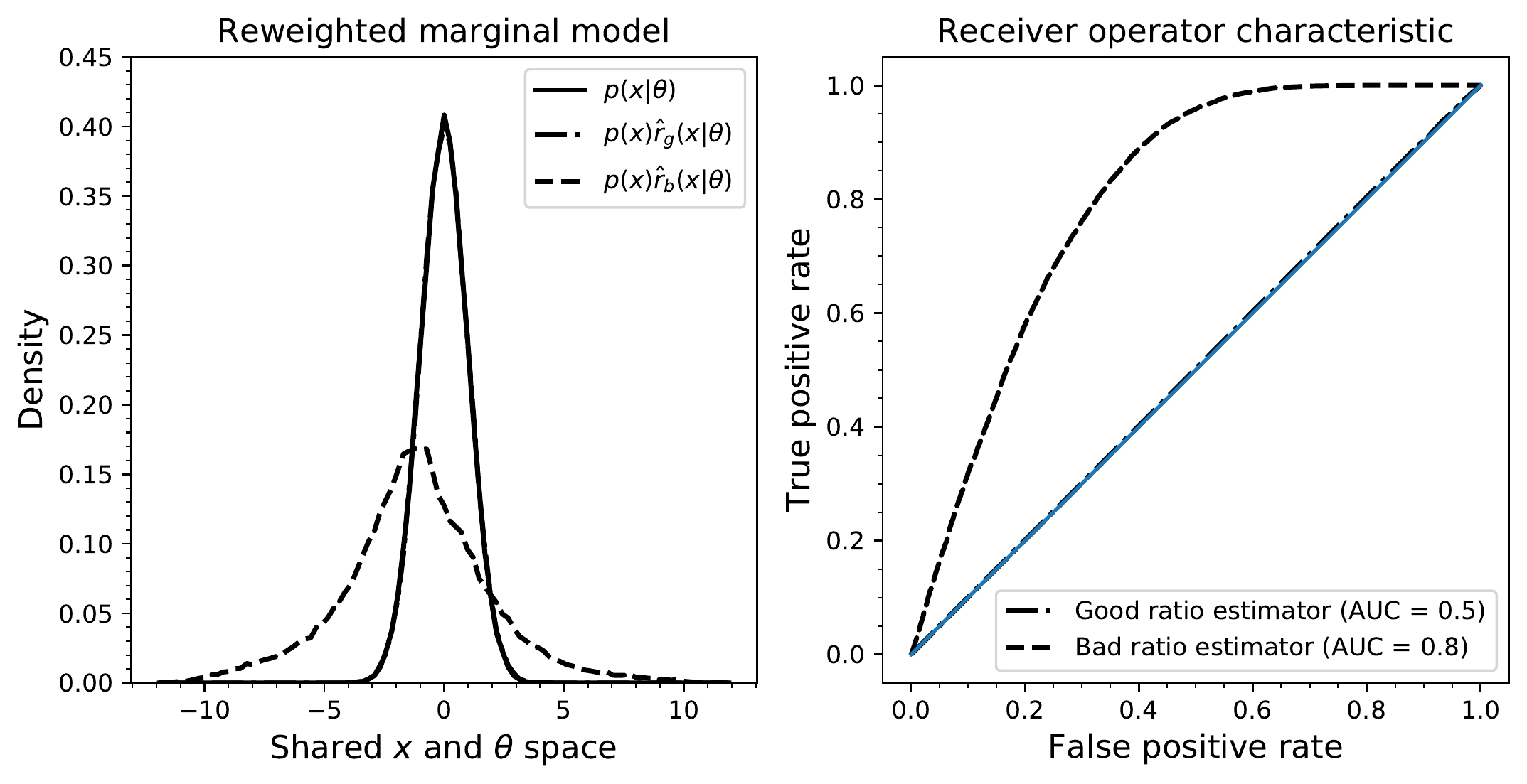}
  \caption{This figure demonstrates the diagnostic presented in Section~\ref{sec:roc}. We train two ratio estimators. The first approximates the ratio $r(\bx\vert\btheta)$ well, while the other does not. We denote these estimators as $\hat{r}_g(\bx\vert\btheta)$ and $\hat{r}_b(\bx\vert\btheta)$ respectively. The test diagnostic is applied to a single test hypothesis $\btheta = 0$. {\it (Left):} Marginal model reweighted using $\hat{r}_g(\bx\vert\btheta)$ and $\hat{r}_b(\bx\vert\btheta)$. It is clear that $\hat{r}_b(\bx\vert\btheta)$ does not properly approximate $r(\bx\vert\btheta)$, as the reweighted marginal model is distinguishable from the test hypothesis $p(\bx\vert\btheta = 0)$. {\it (Right):} A classifier is trained to distinguish between samples from the test hypothesis and the reweighted marginal models. The \textsc{roc} curve indicates that the classifier could not extract any predictive features for samples $\bx\sim p(\bx)$ reweighted by $\hat{r}_g(\bx\vert\btheta)$, indicating a good approximation of $r(\bx\vert\btheta)$ by $\hat{r}_g(\bx\vert\btheta)$.
  }
  \label{fig:diagnostic_roc}
\end{figure}

Likelihood-free computations are challenging to verify as the likelihood is by definition intractable.
A robust strategy is necessary to verify the quality of the approximation before making any scientific conclusion based on a likelihood-free approach.
Inspired by \citet{approxlr}, we identify issues in our ratio-estimator $\hat{r}(\bx\vert\btheta)$ by evaluating the identity $p(\bx\vert\btheta) = p(\bx)\hat{r}(\bx\vert\btheta)$. If $\hat{r}(\bx\vert\btheta)$ is exact, then a classifier should not be able to distinguish between samples from $p(\bx\vert\btheta)$ and the reweighted marginal model $p(\bx)\hat{r}(\bx\vert\btheta)$. The discriminative performance of the classifier can be assessed by means of a \textsc{roc} curve. A diagonal \textsc{roc} (\textsc{auc} = 0.5) curve indicates that a classifier is insensitive and $\hat{r}(\bx\vert\btheta) = r(\bx\vert\btheta)$. This result can also be obtained if the classifier is not powerful enough to extract any predictive features. Figure~\ref{fig:diagnostic_roc} provides an illustration of this diagnostic.
\section{Related work}
\label{sec:related_work}
Algorithms such as \textsc{abc}~\cite{abc0,abc2,abc1,marin2012approximate} tackle the problem of Bayesian inference by collecting proposal states $\btheta\sim p(\btheta)$ whenever an observation $\bx$ produced by the forward model $\bx\sim p(\bx\vert\btheta)$ resembles an observation $\bx_o$. Formally, a proposal state $\btheta$ is accepted whenever a compressed observation $\sigma(\bx)$ (low-dimensional summary statistic) satisfies $d(\sigma(\bx),\sigma(\bx_o)) < \epsilon$ for some distance function $d$ and acceptance threshold $\epsilon$. \emph{The resulting approximation of the posterior will only be exact whenever the summary statistic is sufficient and $\epsilon \to 0$}~\cite{abc1}. Several procedures have been proposed to improve the acceptance rate by guiding simulations based on previously accepted states~\cite{smcabc,abcmcmc,abcmcmc2}. Other works investigated learning summary statistics~\cite{fearnhead2012constructing,dire,jiang2017learning}. Contrary to these methods, \textsc{aalr-mcmc} does not actively use the simulator during inference and learns a direct mapping from data and parameter space to likelihood-to-evidence ratios.

Other approaches take the perspective to cast inference as an optimization
problem~\cite{neal1998view,hoffman2013stochastic}. In variational inference,
a parameterized posterior over parameters of interest
is optimized~\cite{salimansmarkov}. Amortized variational inference~\cite{amortized,amortizeddeep}
expands on this idea by using generative models to capture inference mappings.
Recent work in~\cite{avo} proposes a novel form of variational inference
by introducing an adversary in combination with \textsc{reinforce}-estimates~\cite{reinforce,policygradient} to optimize a parameterized prior. Others have investigated meta-learning to learn parameter updates~\cite{pesah}.
However, these works only provide point-estimates.

Sequential approaches such as \textsc{snpe-a}~\cite{snpea}, \textsc{snpe-b}~\cite{snpeb} and \textsc{apt}/\textsc{snpe-c}~\cite{apt} iteratively adjust an approximate posterior parameterized as a mixture density network or a normalizing flow. Instead of learning the posterior directly, \textsc{snl}~\cite{snl} makes use of autoregressive flows to model an approximate likelihood. \textsc{aalr-mcmc} mirrors \textsc{snl} as the trained conditional density estimator is plugged into \textsc{mcmc} samplers to bypass the intractable marginal model. This allows \textsc{snl} to approximate the posterior numerically. Contrary to our approach, \textsc{snl} cannot directly provide estimates of the posterior posterior density function.

The usage of ratios is explored in several studies. \textsc{carl}~\cite{approxlr} models likelihood ratios for frequentist tests. As shown in Section~\ref{sec:improve_approx_lr}, \textsc{carl} does not produce accurate results in some cases. \textsc{lfire}~\cite{dutta2016likelihood} models a likelihood-to-evidence ratio by logistic regression and relies on the usage of summary statistics. Unlike us, they require samples from the marginal model and a specific (reference) likelihood, while we only require samples from the joint $p(\bx,\btheta)$. Therefore, \textsc{lfire} requires retraining for every evaluation of different $\btheta$.

Finally, an important concern of likelihood-free inference is minimizing the number of simulation calls. Active simulation
strategies such as \textsc{bolfi}~\cite{bolfi} and others~\cite{synlikelihood0,synlikelihood1} achieve this through Bayesian optimization. Emulator networks~\cite{emulatornetworks} exploit the uncertainty within an ensemble to guide simulations.
Recent works~\cite{brehmer2018mining,brehmer2018guide} significantly reduce the amount of required simulations, provided joint likelihood ratios and scores can be extracted from the simulator.
\begin{table*}[t!]
  \centering
  \begin{tabular}{llllll}
    \toprule
    Algorithm & Tractable problem & Detector calibration & Population model & \textsc{m/g/1} & \\
    \midrule
    \textsc{abc}          & $-6.668 \pm 0.000$            & $-2.180 \pm 0.000$         & \textsc{n/a}               & \textsc{n/a}           \\
    \textsc{snpe-a}                                    & $-6.141 \pm 1.227$            & $-1.775 \pm 1.775$         & ${\bf 7.024} \pm 0.515$    & $1.177 \pm 0.937$      \\
    \textsc{snpe-b}                                    & $-5.693 \pm 0.809$            & $-1.075 \pm 0.226$         & $-0.632 \pm 0.843$         & $1.105 \pm 0.384$      \\
    \textsc{apt}                                       & $-4.441 \pm 0.487$            & $-2.004 \pm 0.753$         & $6.366 \pm 0.432$          & $-2.741 \pm 3.356$     \\
    \textsc{snl}                                       & ${\bf -4.060} \pm 0.308$            & \textsc{n/a}         & \textsc{n/a}          & \textsc{n/a}     \\
    \midrule
    \textsc{aalr-mcmc} (ours)                           & $-4.126 \pm 0.004$      & ${\bf -1.005} \pm 0.074$    & $6.482 \pm 0.214$          & ${\bf 2.302} \pm 0.189$ \\
    \bottomrule
  \end{tabular}
  \caption{Posterior log probabilities $\log p(\btheta = \btheta^*\vert\bx = \bx_o)$ for generating parameters $\btheta^*$ and observation $\bx_o$. For \textsc{snpe-a}, \textsc{snpe-b} and \textsc{apt} we directly extracted the posterior log probability from the mixture of Gaussians. Since the proposed ratio estimator models the log likelihood-to-evidence ratio, we
    compute $\log p(\btheta = \btheta^*\vert\bx = \bx_o)$ as $\log r(\bx = \bx_o\vert\btheta = \btheta^*) + \log p(\btheta = \btheta^*)$. Assessing the quality of a method exclusively based on the observed log posterior probabilities is potentially {\bf misleading}, as the metric does not take the structure of the posterior into account.
As such, we provide this table for historic reasons to comply with previous studies such as~\citet{snpea},~\citet{snpeb} and \citet{apt}.}
  \label{table:results_log_probs}
\end{table*}
\section{Experiments}
\label{sec:experiments}
\subsection{Setup}
\label{sec:setup}
We compare \textsc{aalr-mcmc} against rejection \textsc{abc} and
established modern posterior approximation techniques such as \textsc{snl}, \textsc{snpe-a} \textsc{snpe-b} and \textsc{apt}. 
We allocate a \emph{simulation budget} of one million forward passes. Sequential approaches such as \textsc{snpe-a}, \textsc{snpe-b}, and \textsc{apt} spread this budget equally across 100 rounds.
These rounds focus the simulation budget to iteratively improve the approximation of a \emph{single} posterior.
For \textsc{snl}, due to the computational constraints of its inner \textsc{mcmc} sampling step, we limit the simulation budget to $100000$ forward passes spread equally over $100$ rounds. 
Unless stated otherwise, our evaluations assess the posterior estimate obtained in the final round.
Although the ratio estimator in \textsc{aalr-mcmc} is trained \emph{once} to model all posteriors (amortization), we only examine the posterior of interest $p(\btheta\vert\bx=\bx_o)$.
This choice puts our method at a disadvantage since the task of amortized inference is more complex compared to fitting of a single posterior. We stress that from a scientific point of view, accuracy of the approximation is preferred over simulation cost.
All experiments are repeated 25 times. \textsc{aalr-mcmc} makes use of the likelihood-free Metropolis-Hastings sampler. Implementation guidelines are discussed in Appendix~\ref{sec:recommended}. Experimental details, additional results and plots demonstrating several other aspects are discussed in Appendix~\ref{sec:experimental_details}.
Code is available at~\url{https://github.com/montefiore-ai/hypothesis}.

\subsubsection{Benchmark problems}
\paragraph{Tractable problem} Given a model parameter sample $\btheta\in\R^5$, the forward generative process is defined as:
\begin{align*}
  \bm{\mu}_\mathbf{\btheta} &= ({\btheta}_0, {\btheta}_1),\\
  s_1 &= {\btheta}^2_2,~~s_2 = {\btheta}^2_3,~~\rho = \tanh({\btheta}_4),\\
  \bm{\Sigma}_{\btheta} &=
  \begin{bmatrix}
    s_1^2 & \rho s_1 s_2 \\
    \rho s_1 s_2 & s_2^2 \\
  \end{bmatrix},
\end{align*}
\begin{equation*}
  \text{with}~\bx = ({\bx}_1,\ldots,{\bx}_4)~\text{where}~{\bx}_i\sim\mathcal{N}(\bm{\mu}_\mathbf{\btheta},\bm{\Sigma}_{\btheta})
\end{equation*}
The likelihood is $p(\bx\vert\btheta) = \prod_{i=1}^4\mathcal{N}(\bx_i\vert~\bm{\mu}_\mathbf{\theta},~\mathbf{\Sigma}_\mathbf{\theta})$, with a uniform prior $p(\btheta)$ between $[-3,3]$ for every $\btheta_i$. The resulting posterior is non-trivial due to squaring operations, which are responsible for the presence of multiple modes. An observation $\bx_o$ is generated by conditioning the forward model on $\btheta^* = (0.7,-2.9,-1.0,-0.9,0.6)$ as in~\citet{snl} and~\citet{apt}.

\paragraph{Detector calibration} We like to determine the offset $\btheta\in\R$ of a particle detector from the collision point given a detector response $\bx_o$.
Our particle detector emulates a $32 \times 32$ spherical uniform grid such that $\bx\in\R^{1024}$. Every detector pixel measures the momentum of particles passing through the detector material. The \texttt{pythia} simulator~\cite{pythia} generates electron-positron ($e^-e^+$) collisions and is configured according to the parameters derived by the Monash tune~\cite{monash}. The collision products and their momenta are processed by \texttt{pythiamill}~\cite{pythiamill} to compute the response of the detector by simulating the interaction of the collision products with the detector material. We consider a prior $p(\btheta)\triangleq\mathcal{U}(-30, 30)$ with $\bx_o$ generated at $\btheta^* = 0$.

\paragraph{Population model} The Lotka-Volterra model~\cite{lotka} describes the evolution of predator-prey populations. The population dynamics are driven by a set of differential equations with parameters $\btheta \in \R^4$. An observation describes the population counts of both groups over time. Simulations are typically compressed into a summary statistic $\bar{\bx} \in \R^9$~\cite{snl,apt}. We also follow this approach to remain consistent.
The prior $p(\btheta)\triangleq\mathcal{U}(-10,2)$ (log-scale) for every $\btheta_i$. We generate an observation from the narrow oscillating regime $\btheta^* = (-4.61, -0.69, 0, -4.61)$.

\paragraph{\textsc{m/g/1} queuing model} This model describes a queuing system of continuously arriving jobs at a single server and is described by a model parameter $\btheta\in\R^3$. The time it takes to process every job is uniformly distributed in the interval $\left[\btheta_1,\btheta_2\right]$. The arrival time between two consecutive jobs is exponentially distributed according to the rate $\btheta_3$. An observation $\bx$ are 5 equally spaced percentiles of interdeparture times, i.e., the 0th, 25th, 50th, 75th and 100th percentiles. An observation $\bx_o$ is generated by conditioning the forward model on $\btheta^* = (1.0,5.0,0.2)$ as in~\citet{snl}. We consider a uniform prior $p(\btheta) \triangleq \mathcal{U}(0,10) \times \mathcal{U}(0,10) \times \mathcal{U}(0,0.333)$.

\subsection{Results}
\label{sec:experiments_summary}
\begin{table}[b!]
  \centering
  \begin{tabular}{lll}
    \toprule
    Algorithm & \textsc{mmd} & \textsc{roc auc} \\
    \midrule
    \textsc{aalr-mcmc} (ours)        & ${\bf 0.05} \pm 0.005$  & ${\bf 0.58} \pm 0.0080$ \\
    \textsc{aalr-mcmc} (\textsc{lrt}) & $0.53 \pm 0.004$        & $0.99 \pm 0.0001$       \\
    \textsc{abc}      & $0.29 \pm 0.004$        & $0.98 \pm 0.0007$       \\
    \textsc{snpe-a}                   & $0.21 \pm 0.070$        & $0.93 \pm 0.0305$       \\
    \textsc{snpe-b}                   & $0.20 \pm 0.061$        & $0.91 \pm 0.0409$       \\
    \textsc{apt}                      & $0.17 \pm 0.036$        & $0.83 \pm 0.0145$       \\
    \textsc{snl}                      & $0.11 \pm 0.091$        & $0.63 \pm 0.0564$ \\
    \bottomrule
  \end{tabular}
  \caption{Results for the tractable benchmark. \textsc{aalr-mcmc} outperforms all other methods across in terms of accuracy and robustness (low variance). Numerical errors introduced by \textsc{mcmc} might have contributed to these results. The \textsc{mmd} scores are in agreement with~\citet{apt}.
    }
  \label{table:results_tractable}
\end{table}
\begin{figure*}[t!]
  \centering
  \begin{subfigure}{.162\linewidth}
     \centering
     \includegraphics[width=\linewidth]{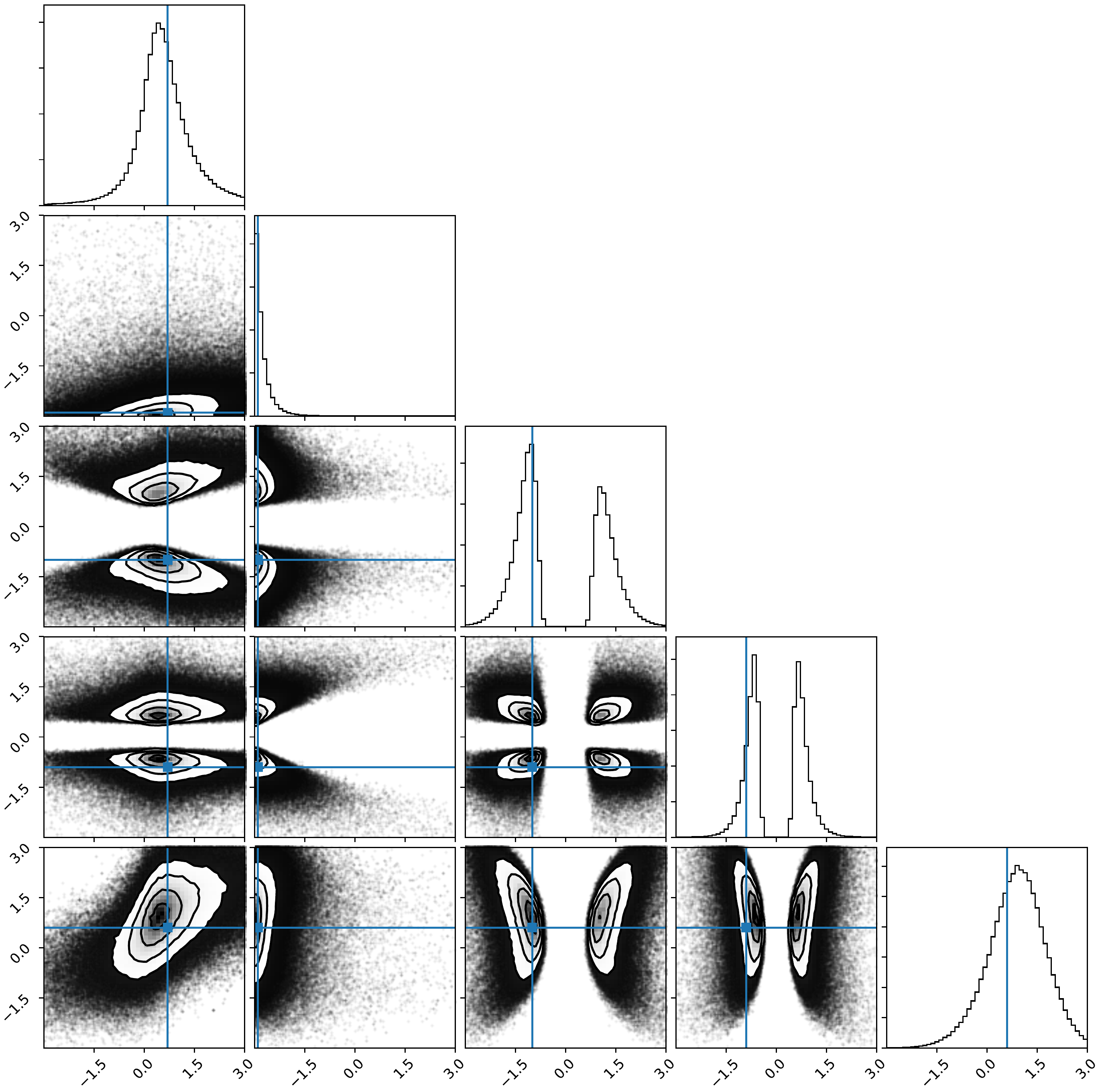}
     \caption{\textsc{mcmc} groundtruth}
  \end{subfigure}
  \begin{subfigure}{.162\linewidth}
     \centering
     \includegraphics[width=\linewidth]{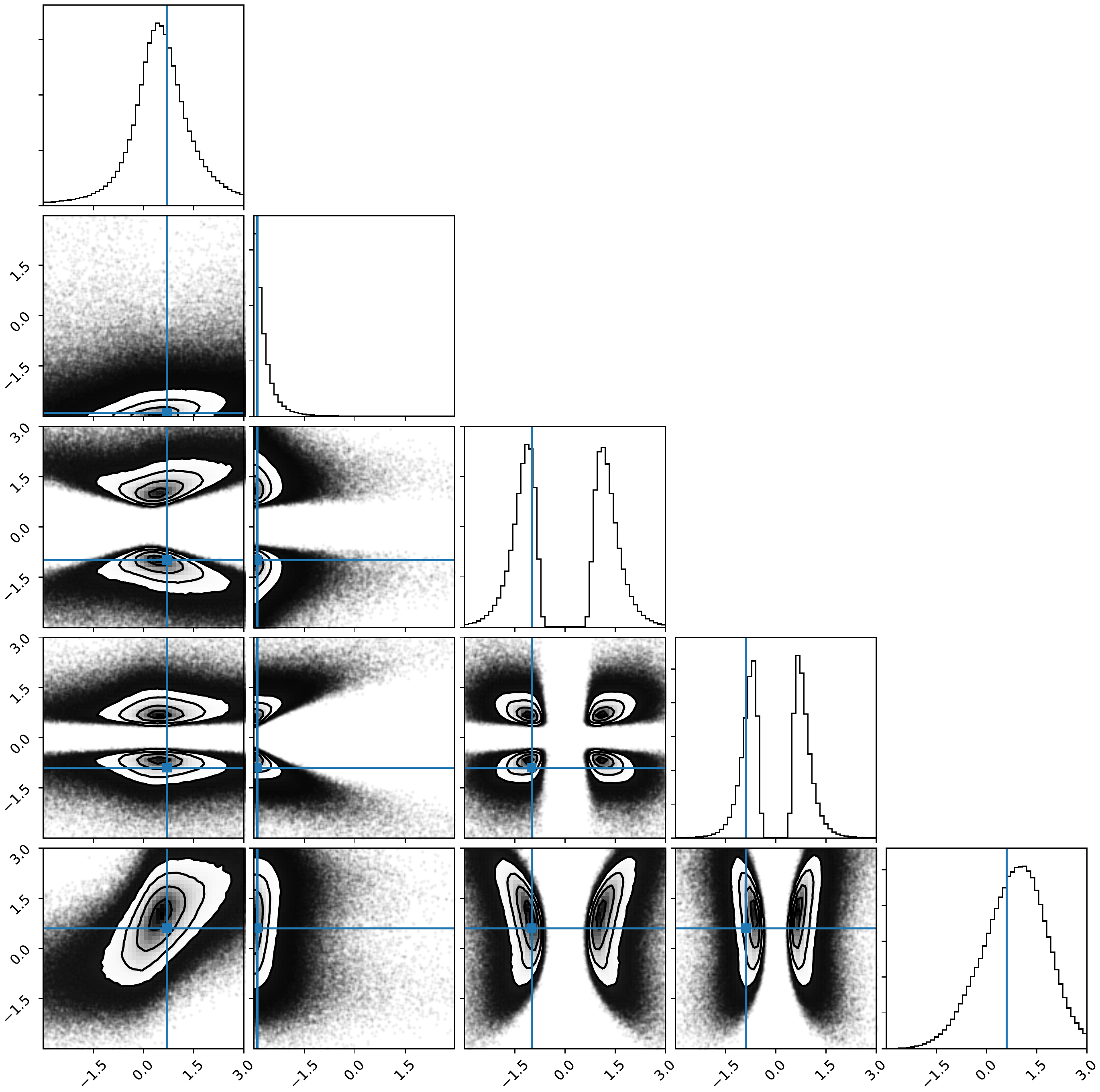}
     \caption{\textsc{aalr-mcmc}}
  \end{subfigure}
  \begin{subfigure}{.162\linewidth}
     \centering
     \includegraphics[width=\linewidth]{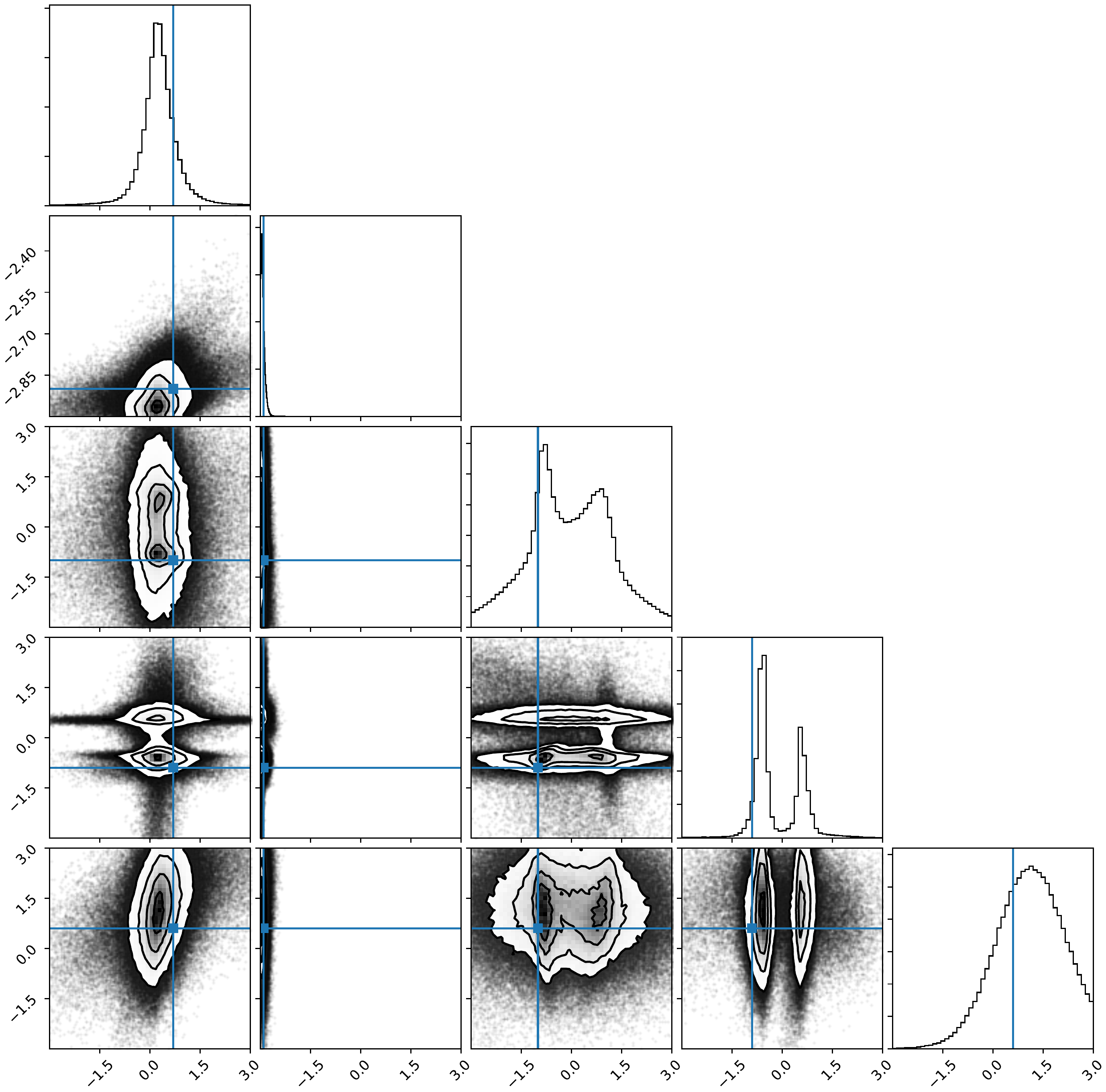}
     \caption{\textsc{snpe-a}}
  \end{subfigure}
  \begin{subfigure}{.162\linewidth}
     \centering
     \includegraphics[width=\linewidth]{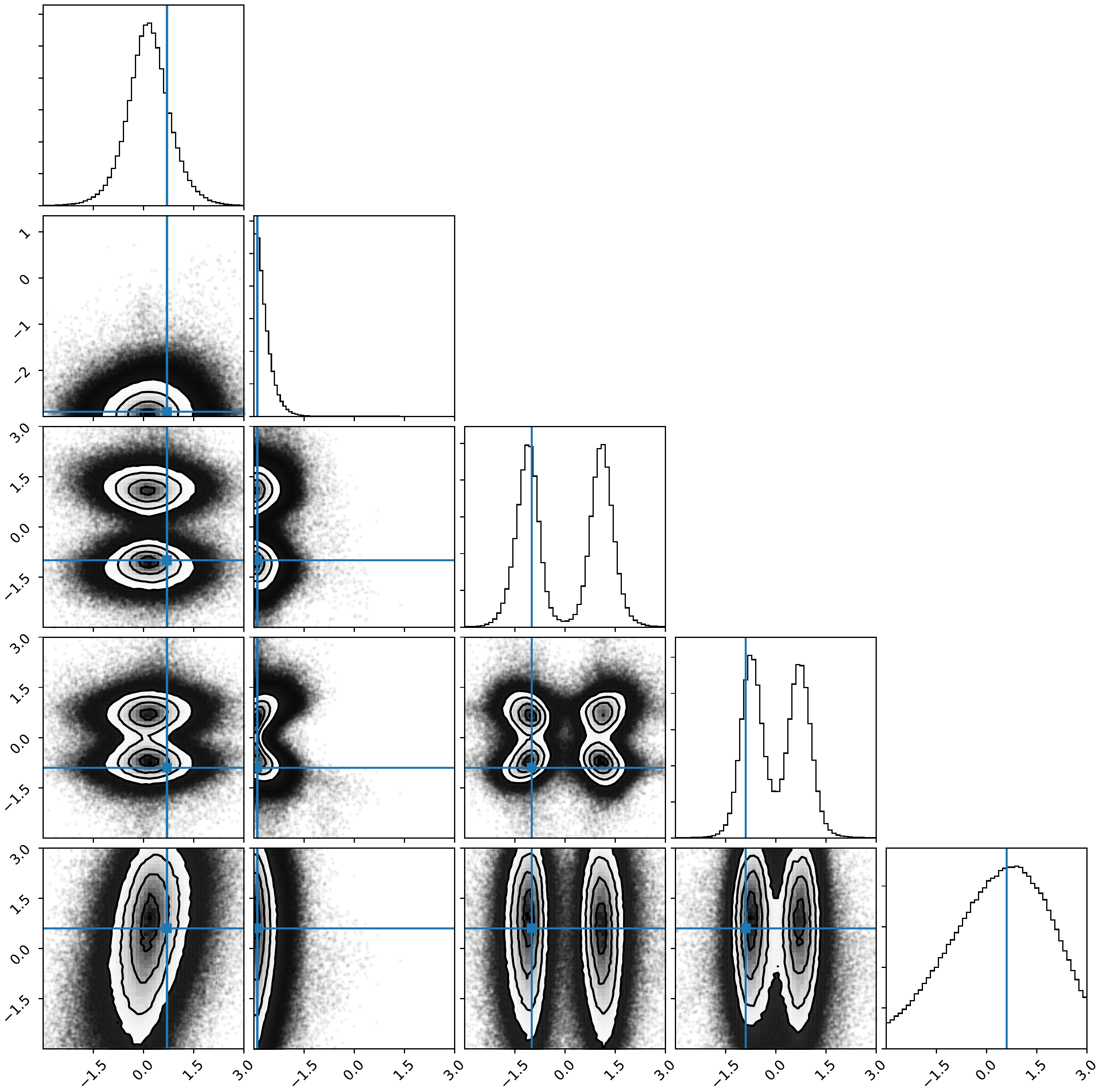}
     \caption{\textsc{snpe-b}}
  \end{subfigure}
  \begin{subfigure}{.162\linewidth}
     \centering
     \includegraphics[width=\linewidth]{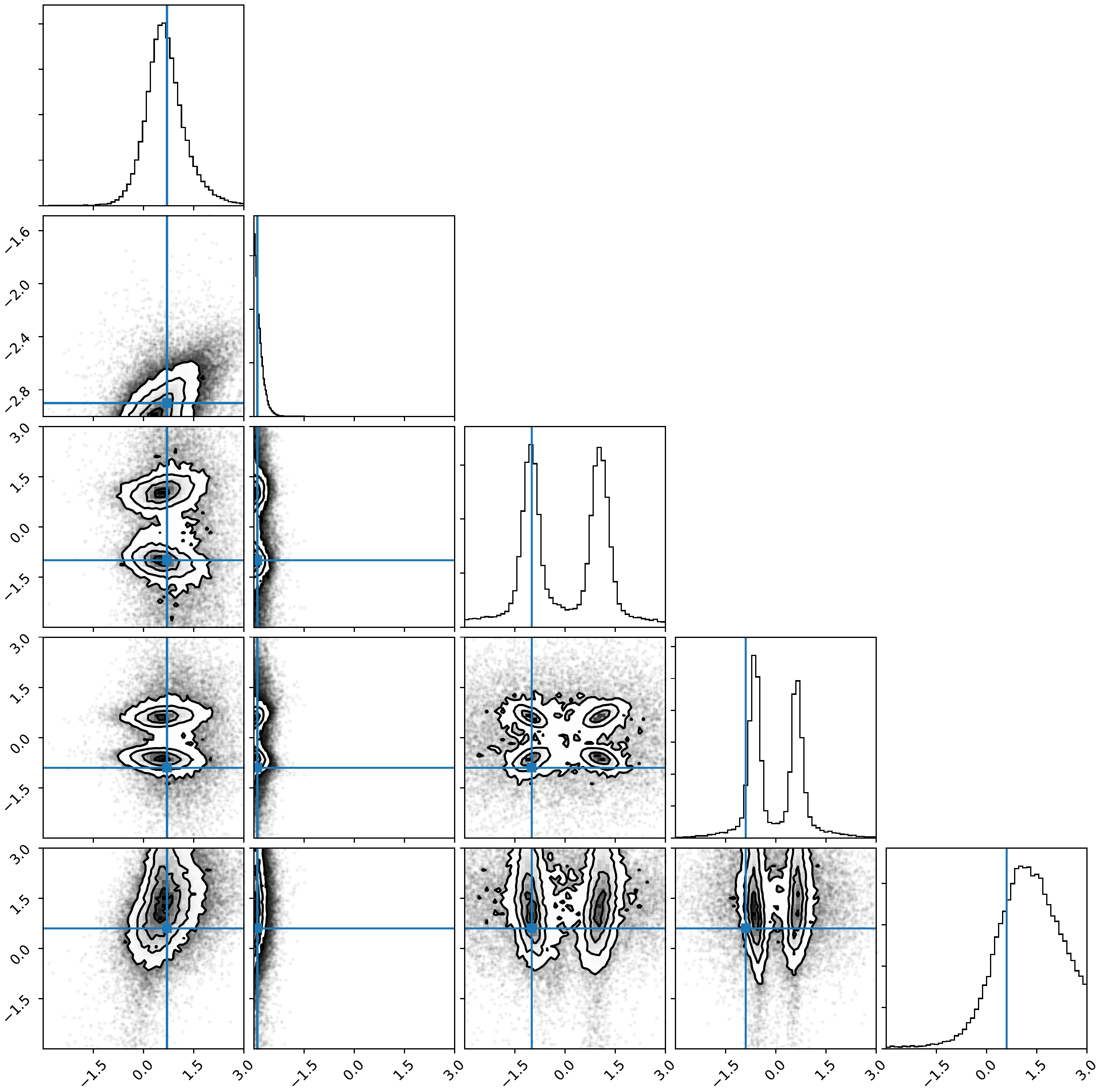}
     \caption{\textsc{apt}}
  \end{subfigure}
  \begin{subfigure}{.162\linewidth}
     \centering
     \includegraphics[width=\linewidth]{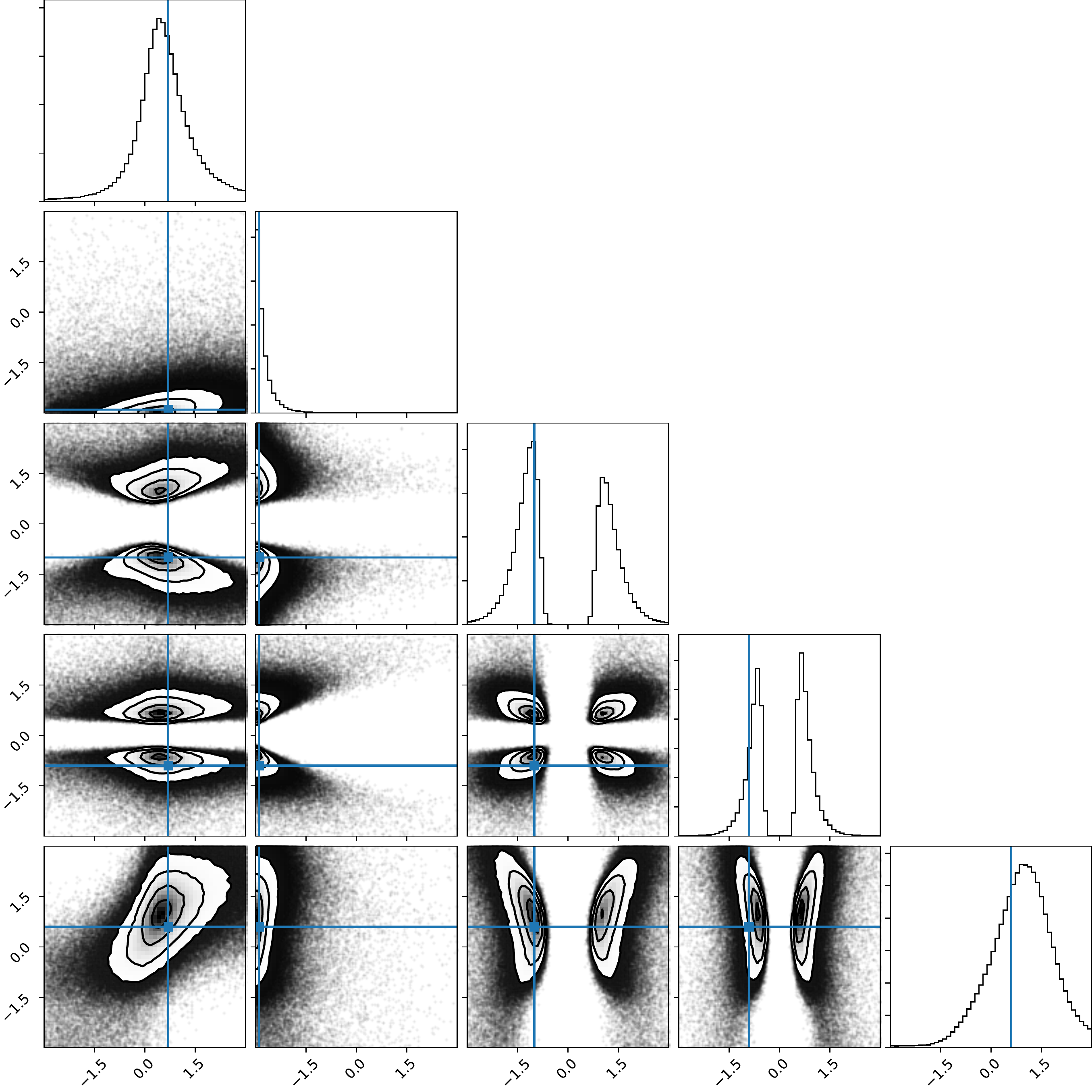}
     \caption{\textsc{snl}}
  \end{subfigure}
  \caption{Posteriors from the tractable benchmark. The experiments are repeated 25 times and the approximate posteriors are subsampled from those runs. An objective visual assessment can be made: \textsc{aalr-mcmc} shares the same structure with the \textsc{mcmc} truth, demonstrating its accuracy. Some runs of the other methods were not consistent, contributing to the variance observed in Table~\ref{table:results_tractable}.}
  \label{fig:tractable_all_posteriors}
\end{figure*}
Table~\ref{table:results_log_probs} shows the posterior log probabilities of the generating parameter $\btheta^*$ for an observation $\bx_o$. Additionally, the \textsc{roc} diagnostic  for our method  reports \textsc{auc} = 0.58 for the tractable problem, \textsc{auc} = 0.5 for the detector calibration and \textsc{m/g/1} benchmarks, and \textsc{auc} = 0.55 for the population evolution model. These results demonstrate that the proposed ratio estimator provides accurate ratio estimates. 

If we assess the quality of the methods exclusively based on the log posterior probabilities in Table~\ref{table:results_log_probs}, we could argue that all methods are close to each other in terms of approximation, with \textsc{aalr-mcmc} yielding the best results for the detector calibration and \textsc{m/g/1} and \textsc{snl} and \textsc{snpa-a} respectively producing the most accurate inference for the tractable problem and the population evolution model. However, this is potentially misleading, as the metric does not take the structure of the posterior into account.
Again, we stress that the ability of inference techniques to approximate the posterior accurately is critical in scientific applications which seek to, for instance, constrain the model parameter $\btheta$.
To demonstrate this point, we focus on the tractable problem and carry out two distinct quantitative analyses between the samples of the approximate posterior and the \textsc{mcmc} groundtruth. The first computes the Maximum Mean Discrepancy (\textsc{mmd})~\cite{mmd} while the latter trains a classifier to compute the \textsc{roc auc}.
Results are summarized in Table~\ref{table:results_tractable} while
Figure~\ref{fig:tractable_all_posteriors} shows the approximations and the groundtruth.
Both \textsc{aalr-mcmc} and \textsc{snl} accurately model the true posterior, but \textsc{snpa-a}, \textsc{snpa-b} and \textsc{apt} clearly fail to do so.
The observed discrepancy between the \textsc{lrt} and the proposed ratio estimator indicates that the improvements in Section~\ref{sec:improve_approx_lr} are \emph{critical}.



\begin{figure}[b!]
  \centering
  \vspace{-0.7cm}
  \includegraphics[width=\linewidth]{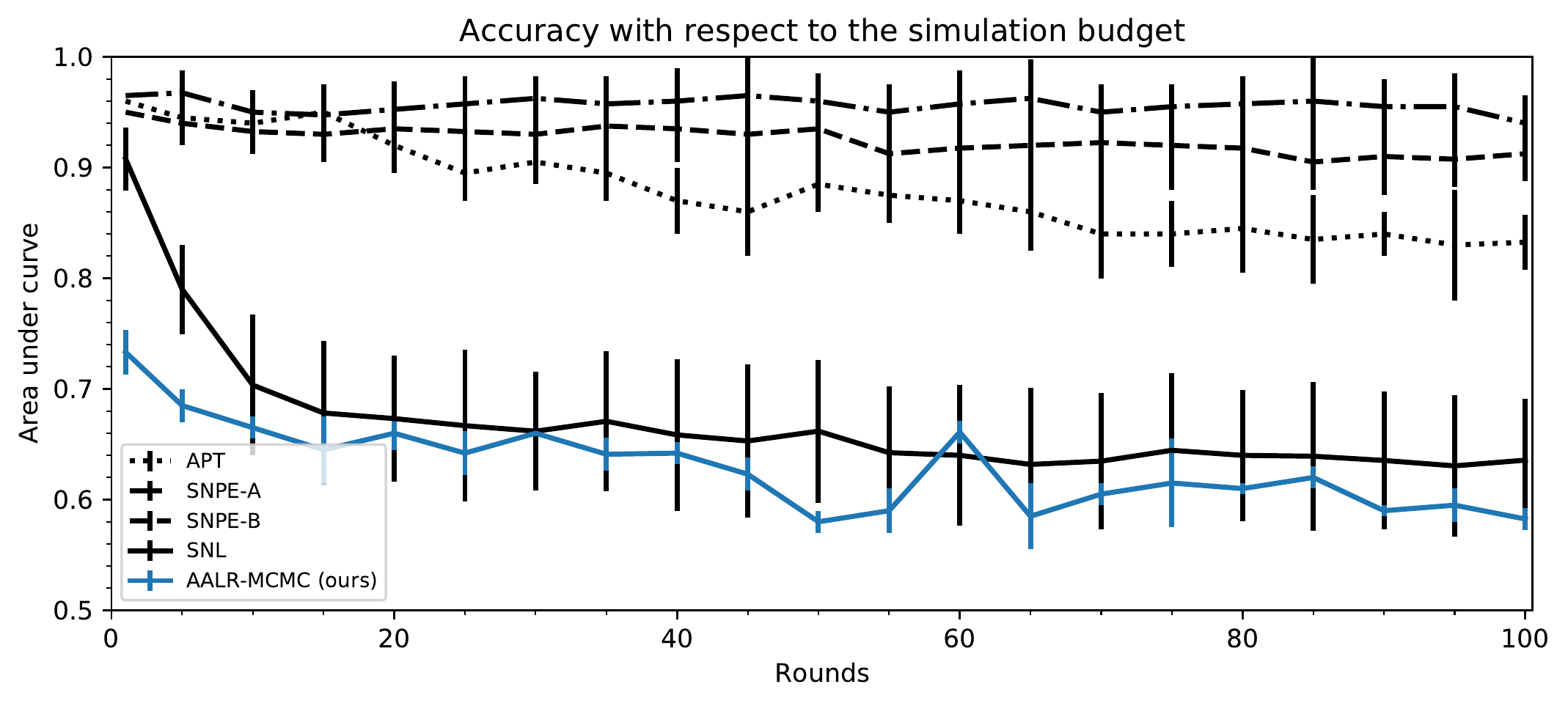}
  \caption{We evaluate the accuracy of the approximations with respect to different simulation budgets on the tractable benchmark. The accuracy is obtained by computing the \textsc{roc auc} between samples from the approximation and the \textsc{mcmc} groundtruth. Except for \textsc{snl} which yields comparable results, sequential approaches are not able to outperform \textsc{aalr-mcmc}.}
  \label{fig:simulation_efficiency}
\end{figure}

In addition to comparing the final approximations, we evaluate the accuracy of the approximations with respect to a given simulation budget. In doing so we challenge our method even further, as sequential approaches are specifically designed to be simulation efficient. We expect sequential approaches to obtain more accurate approximations with less simulations.
The results of this evaluation are shown in Figure~\ref{fig:simulation_efficiency}. With the exception of \textsc{snl} which produces results comparable to ours, we unexpectedly find that the sequential approaches were not able to outperform our method on this (toy) problem, even though \textsc{aalr-mcmc} and its ratio estimator tackle the harder task of amortized inference. This demonstrates the accuracy and robustness of our method.

\subsection{Demonstrations: strong gravitational lensing}
\label{sec:use_cases}
The following demonstrations will showcase several aspects of our method while considering the problem of strong gravitational lensing.
We use \texttt{autolens}~\cite{autolens} to simulate the telescope optics, imaging sensors and physics governing strong lensing. The simulation black-box encapsulates these components. The output of the simulation is a high-dimensional observation $\bx \in\R^{128\times128}$ with uninformative data dimensions. We use a ratio estimator based on \textsc{resnet-18}~\cite{resnet} parameterized by $\btheta$ in the fully connected trunk. Appendix~\ref{sec:lensing_appendix} discusses the setups and the simulation models in detail.

\subsubsection{Marginalization of nuisance parameters}
\label{sec:marginalization}
\begin{figure}[H]
  \centering
  \includegraphics[width=\linewidth]{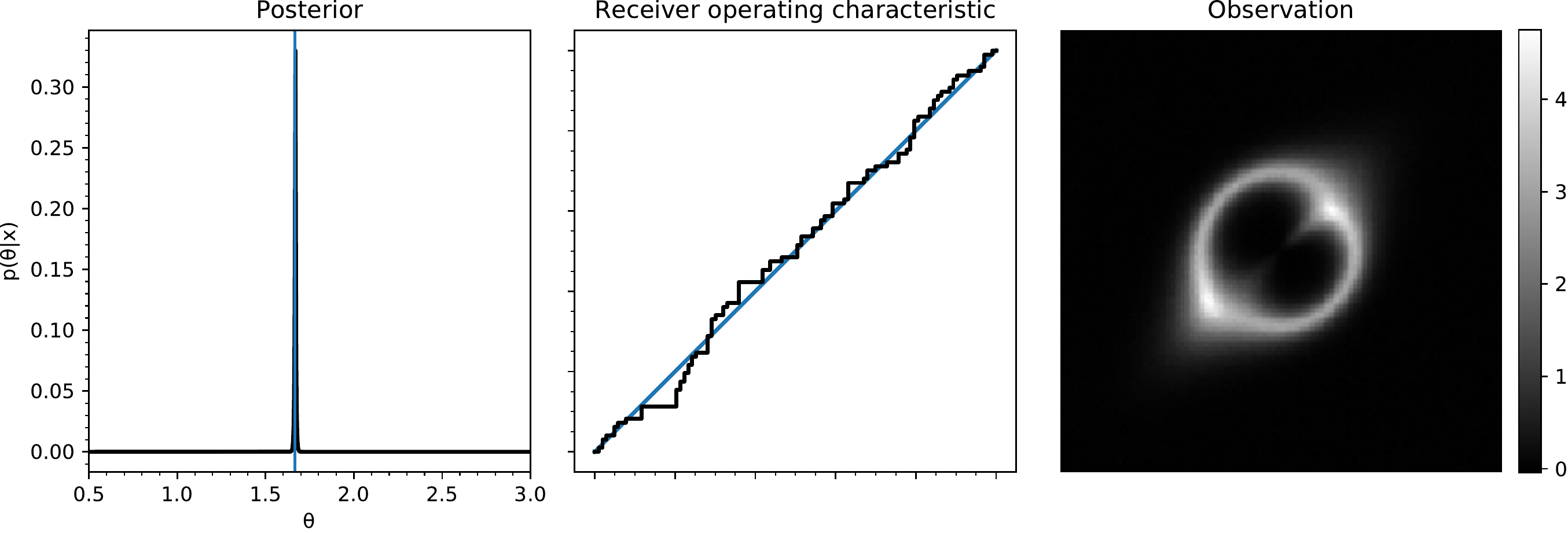}
  \caption{{\it (Left):} Approximation of the posterior. {\it (Middle):} Diagonal \textsc{roc} diagnostic, indicating a good approximation of the posterior. {\it (Right):} Observation associated with the posterior.}
  \label{fig:lensing_marginalization}
\end{figure}

Often scientists are aphetic about a posterior describing all model parameters. Rather, they are interested in a posterior in which nuisance parameters have been marginalized out. This is easily achieved within our framework by including all parameters (including nuisance parameters) to the simulation model, but only presenting the parameters of interest to the ratio estimator during training. The training procedure remains otherwise unchanged. This problem focuses on recovering the Einstein radius $\btheta\in\R$ of a gravitational lens. We are not interested in the parameters describing the source and foreground galaxy (15 parameters). Figure~\ref{fig:lensing_marginalization} depicts our posterior approximation, \textsc{roc} diagnostic and observation $\bx_o$ with $\btheta^* = 1.66$ and prior $p(\btheta)\triangleq\mathcal{U}(0.5,3.0)$.
\subsubsection{Amortization enables population studies}
\begin{figure}[H]
  \centering
  \includegraphics[width=\linewidth]{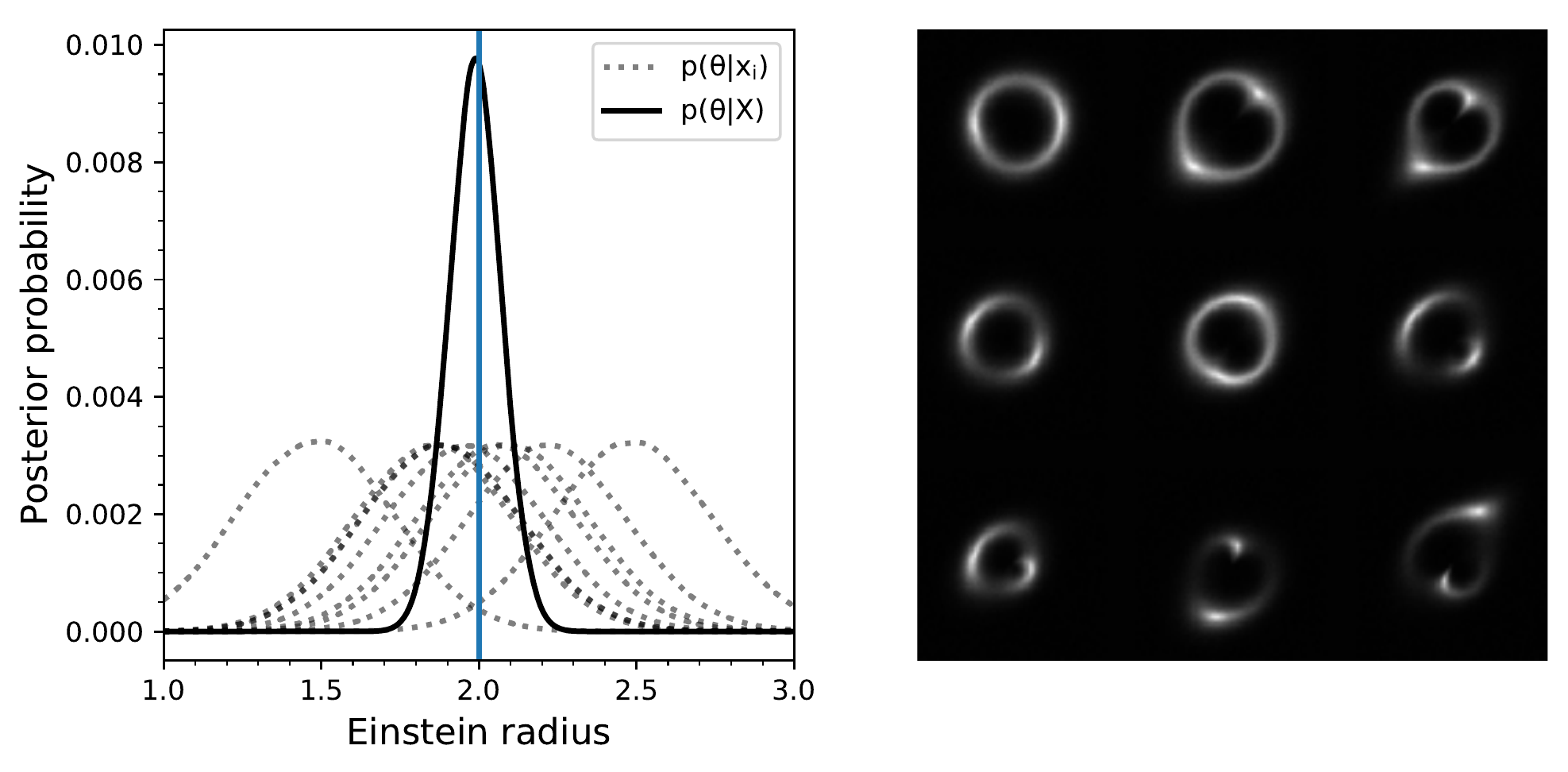}
  \caption{{\it (Left)}: The dotted lines represent the posteriors $\hat{p}(\btheta\vert\bx=\bx_i)$ for every independent and identically distributed observation $\bx_i$, while the solid line depicts the posterior $\hat{p}(\btheta\vert\mathcal{X})$. All posteriors are derived using the same pretrained ratio estimator. {\it (Right)}: Observations sampled from $p(\bx\vert\btheta = \btheta^*)$.}
  \label{fig:population_posterior}
\end{figure}

Consider a set of $n$ independent and identically distributed observations $\mathcal{X} = \{\bx_1,\ldots,\bx_n\}$. The amortization of the ratio estimator allows additional observations to be included in the computation of the posterior $p(\btheta\vert\mathcal{X})$ without requiring new simulations or retraining. This allows us to efficiently undertake population studies. Bayes' rule tells us
\begin{align}
  \begin{split}
  p(\btheta\vert\mathcal{X}) &= \frac{p(\btheta)\prod_{\bx\in\mathcal{X}}p(\bx\vert\btheta)}{\int p(\btheta)\prod_{\bx\in\mathcal{X}}p(\bx\vert\btheta) d\btheta}, \\
  &\approx \frac{p(\btheta)\prod_{\bx\in\mathcal{X}}\hat{r}(\bx\vert\btheta)}{\int p(\btheta)\prod_{\bx\in\mathcal{X}}\hat{r}(\bx\vert\btheta) d\btheta}.
\end{split}
\end{align}
The denominator can efficiently be approximated by Monte Carlo sampling using the ratio estimator $\hat{r}(\bx\vert\btheta)$.
However, with \textsc{mcmc} the denominator cancels out within the ratio between consecutive states $\btheta_t\rightarrow\btheta'$. Thereby obtaining
\begin{equation}
  \frac{\hat{p}({\btheta}'\vert\mathcal{X})}{\hat{p}({\btheta}_t\vert\mathcal{X})} =
  \frac{
      p(\btheta')\prod_{\bx\in\mathcal{X}}\hat{r}(\bx\vert\btheta')
  }{
      p(\btheta_t)\prod_{\bx\in\mathcal{X}}\hat{r}(\bx\vert\btheta_t)
  }.
\end{equation}
We consider the same simulation model as in Section~\ref{sec:marginalization}, with the exception that the Einstein radius used to simulate a gravitational lens is not $\btheta$, but instead drawn from $\mathcal{N}(\btheta, 0.25)$. We reduce the uncertainty about the generating parameter $\btheta^* = 2$ by modeling the posterior $\hat{p}(\btheta\vert\mathcal{X})$. This is demonstrated in Figure~\ref{fig:population_posterior}. All individual posteriors (dotted lines) are derived using the same pretrained ratio estimator. The posterior $\hat{p}(\btheta\vert\mathcal{X})$ is approximated using the formalism described above.

\subsubsection{Bayesian model selection}
\begin{figure}[H]
  \centering
  \includegraphics[width=\linewidth]{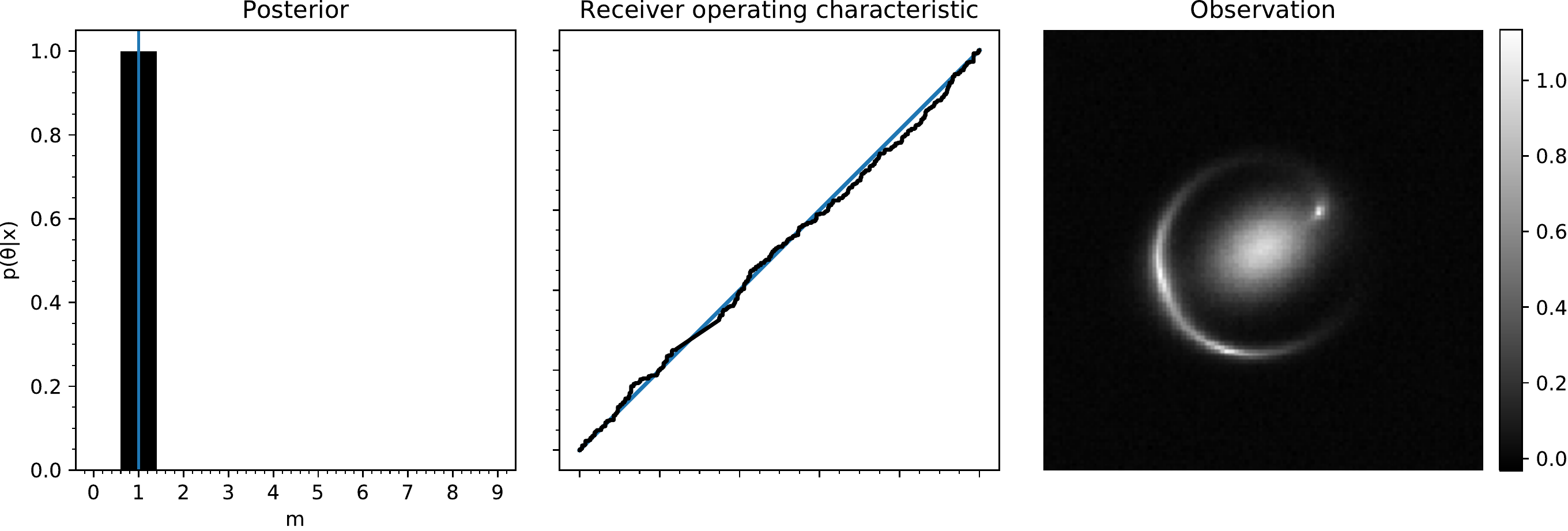}
  \includegraphics[width=\linewidth]{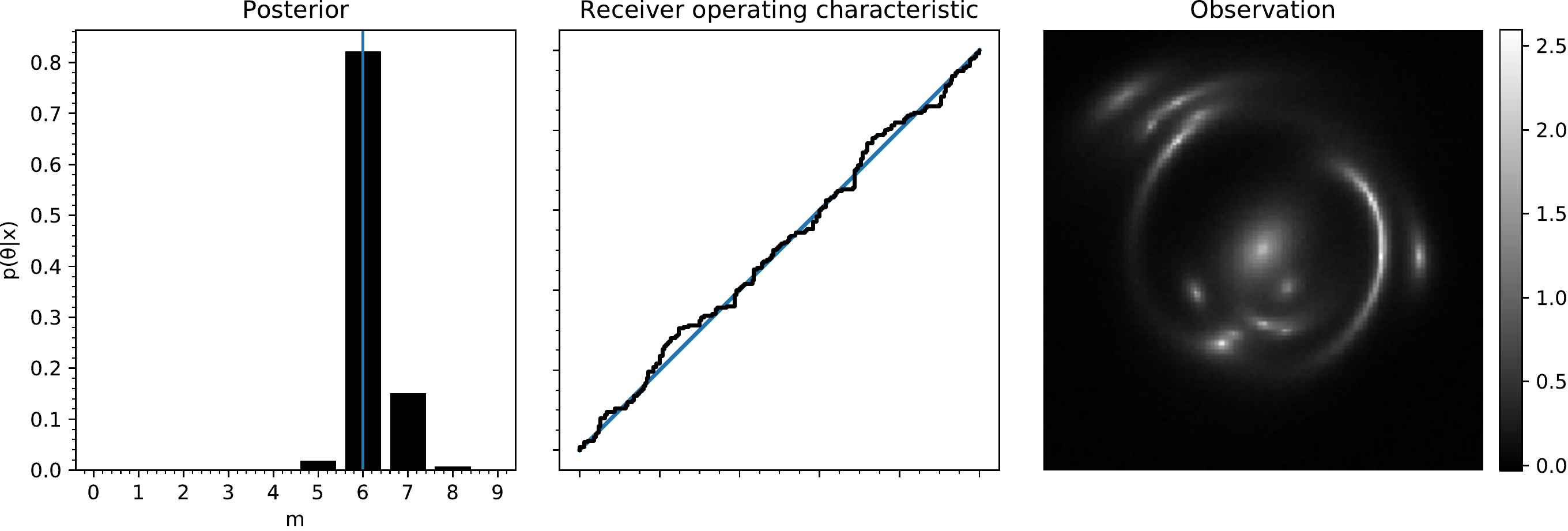}
  \caption{Posterior $\hat{p}(m\vert\bx)$ over the model space $\mathcal{M}$. Both diagnostics are diagonal. {\it (Top)}: Lensing system with a single source galaxy. {\it (Bottom):} Lensing system with 6 different source galaxies. The \textsc{map} of the posterior $\hat{p}(m\vert\bx)$ identifies the correct number of source galaxies, despite abundant lensing artifacts.}
  \label{fig:lensing_selection}
\end{figure}

 Until now we only considered posteriors with continuous model parameters. We turn to a setting in which scientists are interested in a discrete space of models. \emph{In essence casting classification as Bayesian model selection, allowing us the quantify the uncertainty among models (classes) with respect to an observation.}
We demonstrate the task of model selection by computing the posterior $\hat{p}(m\vert\bx)$ across a space of 10 models $\mathcal{M} = \left\{m_0,\ldots,m_{9}\right\}$. The index $i$ of a model $m_i$ corresponds to the number of source galaxies present in the lensing system. The categorical prior $p(m)$ is uniform. Figure~\ref{fig:lensing_selection} shows $\hat{p}(m\vert\bx)$ and the associated diagnostic for different observations. Both posteriors were computed using the same ratio estimator.

\subsection{Estimator capacity and sequential ratio estimation}

The amortization of our ratio estimator requires sufficient representational capacity to accurately approximate $r(\bx\vert\btheta)$, which of course directly depends on the complexity of the task at hand.
As we explore in Appendix~\ref{sec:capacity}, if the capacity of the ratio estimator is too low, then the quality of inference is impaired. 

However, increasing the capacity of a ratio estimator to match the complexity of the inference problem is not always a viable strategy, nor easy to determine beforehand. We observe that for a trained classifier $\bs(\bx,\btheta)$ with insufficient capacity (\textsc{auc} > 0.5) the posterior $\hat{p}(\btheta\vert\bx=\bx_o)$ is typically larger compared to the true posterior. 
Since the true decision function cannot be modeled, the loss of the classifier $\bs(\bx,\btheta)$ is indeed necessarily larger than the loss of the optimal classifier, which effectively means that the classifier $\bs(\bx,\btheta)$ should not be able to exclude sample-parameter pairs $(\bx,\btheta)$.
This is a desirable property because the generating parameters $\btheta^*$ should not be excluded either. 
From this observation, we can run a sequential ratio estimation procedure in which the posterior for $\bx=\bx_o$ is refined iteratively across a series of rounds. 
Starting with the initial prior $p_0(\btheta) := p(\btheta)$, we improve the posterior by setting as prior for the next round, $p_{t + 1}(\btheta)$, the posterior $\hat{p}_t(\btheta\vert\bx=\bx_o)$ obtained at the previous round. At each iteration, the  training procedure is repeated and eventually terminates based on the \textsc{roc} diagnostic (\textsc{auc} = 0.5).

To demonstrate this sequential ratio estimation procedure, let us assume the population model setting. Our ratio estimator is a low-capacity \textsc{mlp} with 3 layers and 50 hidden units. In every round $t$, 10,000 sample-parameter pairs are drawn from the joint $p(\bx,\btheta)$ with prior $p_t(\btheta)$ for training. The following \textsc{auc} scores were obtained: .99, .92, .54, and finally .50, terminating the algorithm.

Let us finally note that some time after the first version of this work, \citet{2020arXiv200203712D} identified that the sequential ratio estimation procedure outlined here is strongly related to $\textsc{apt/snpe-c}$, in the sense that both approaches can actually be viewed as instances of a more general and unified constrastive learning scheme.

\section{Summary and discussion}
\label{sec:conclusion}
This work introduces a novel approach for Bayesian inference.We achieve this by replacing the intractable evaluation of the likelihood ratio in \textsc{mcmc} with an amortized likelihood ratio estimator. We demonstrate that a straightforward application of the likelihood ratio trick to \textsc{mcmc} is insufficient. We solve this by modeling the likelihood-to-evidence ratio for arbitrary observations $\bx$ and model parameters $\btheta$. This implies that a pretrained ratio estimator can be used to infer the posterior density function of arbitrary observations. A theoretical argument demonstrates that the training procedure yields the optimal ratio estimator. The accuracy of an approximation can easily be verified by the proposed diagnostic. No summary statistics are required, as the technique directly learns mappings from observations and model parameters to likelihood-to-evidence ratios. Our framework allows for the usage of off-the-shelf neural architectures such as \textsc{resnet}~\cite{resnet}. Experiments highlight the accuracy and robustness of our method.

\paragraph{Simulation efficiency} We take the point of view that accuracy of the approximation is preferred over simulation cost. This is the case in many scientific disciplines which seek to reduce the uncertainty over a parameter of interest. Despite the experimental handicap, we have shown that existing simulation efficient approaches are not able to outperform our method in terms of accuracy with respect to a certain (and small) simulation budget.

\subsubsection*{Acknowledgments}
The authors would like to thank Antoine Wehenkel and Matthia Sabatelli for the insightful discussions and comments. Joeri Hermans would like to thank the National Fund for Scientific Research for his FRIA scholarship.
Gilles Louppe is recipient of the ULiège - NRB Chair on Big data and is thankful for the support of NRB.

\bibliography{main.bib}

\begin{thebibliography}{62}
\providecommand{\natexlab}[1]{#1}
\providecommand{\url}[1]{\texttt{#1}}
\expandafter\ifx\csname urlstyle\endcsname\relax
  \providecommand{\doi}[1]{doi: #1}\else
  \providecommand{\doi}{doi: \begingroup \urlstyle{rm}\Url}\fi

\bibitem[Azadi et~al.(2018)Azadi, Olsson, Darrell, Goodfellow, and
  Odena]{azadi2018discriminator}
Azadi, S., Olsson, C., Darrell, T., Goodfellow, I., and Odena, A.
\newblock Discriminator rejection sampling.
\newblock \emph{arXiv preprint arXiv:1810.06758}, 2018.

\bibitem[Baldi et~al.(2016)Baldi, Cranmer, Faucett, Sadowski, and
  Whiteson]{Baldi:2016fzo}
Baldi, P., Cranmer, K., Faucett, T., Sadowski, P., and Whiteson, D.
\newblock Parameterized neural networks for high-energy physics.
\newblock \emph{Eur. Phys. J. C}, 76\penalty0 (5):\penalty0 235, April 2016.
\newblock ISSN 1434-6044, 1434-6052.
\newblock URL \url{https://doi.org/10.1140/epjc/s10052-016-4099-4}.

\bibitem[Beaumont et~al.(2002)Beaumont, Zhang, and Balding]{abc1}
Beaumont, M.~A., Zhang, W., and Balding, D.~J.
\newblock Approximate {Bayesian} computation in population genetics.
\newblock \emph{Genetics}, 162\penalty0 (4):\penalty0 2025--2035, 2002.
\newblock URL \url{http://www.genetics.org/content/162/4/2025}.

\bibitem[Betancourt(2017)]{hmcsummary}
Betancourt, M.
\newblock A conceptual introduction to {Hamiltonian} {Monte} {Carlo}.
\newblock \emph{arXiv preprint arXiv:1701.02434}, 2017.
\newblock URL \url{https://arxiv.org/abs/1701.02434}.

\bibitem[Boelts et~al.(2019)Boelts, Lueckmann, Goncalves, Sprekeler, and
  Macke]{snpeb}
Boelts, J., Lueckmann, J.-M., Goncalves, P.~J., Sprekeler, H., and Macke, J.~H.
\newblock Comparing neural simulations by neural density estimation.
\newblock In \emph{2019 Conference on Cognitive Computational Neuroscience},
  pp.\  1289--1299. Cognitive Computational Neuroscience, 2019.
\newblock URL \url{https://doi.org/10.32470/ccn.2019.1291-0}.

\bibitem[Brehmer et~al.(2018)Brehmer, Cranmer, Louppe, and
  Pavez]{brehmer2018guide}
Brehmer, J., Cranmer, K., Louppe, G., and Pavez, J.
\newblock A guide to constraining effective field theories with machine
  learning.
\newblock \emph{Phys. Rev. D}, 98\penalty0 (5):\penalty0 052004, September
  2018.
\newblock ISSN 2470-0010, 2470-0029.
\newblock URL \url{https://doi.org/10.1103/physrevd.98.052004}.

\bibitem[Brehmer et~al.(2020)Brehmer, Louppe, Pavez, and
  Cranmer]{brehmer2018mining}
Brehmer, J., Louppe, G., Pavez, J., and Cranmer, K.
\newblock Mining gold from implicit models to improve likelihood-free
  inference.
\newblock \emph{Proc Natl Acad Sci USA}, 117\penalty0 (10):\penalty0
  5242--5249, February 2020.
\newblock ISSN 0027-8424, 1091-6490.
\newblock URL \url{https://doi.org/10.1073/pnas.1915980117}.

\bibitem[Clevert et~al.(2015)Clevert, Unterthiner, and Hochreiter]{elu}
Clevert, D.-A., Unterthiner, T., and Hochreiter, S.
\newblock Fast and accurate deep network learning by exponential linear units
  (elus).
\newblock \emph{arXiv preprint arXiv:1511.07289}, 2015.

\bibitem[Cranmer et~al.(2015)Cranmer, Pavez, and Louppe]{approxlr}
Cranmer, K., Pavez, J., and Louppe, G.
\newblock Approximating likelihood ratios with calibrated discriminative
  classifiers.
\newblock \emph{arXiv preprint arXiv:1506.02169}, 2015.
\newblock URL \url{https://arxiv.org/abs/1506.02169}.

\bibitem[Crosby(2020)]{pythiamill}
Crosby, O.
\newblock {PYTHIA,} pythia, April 2020.
\newblock URL \url{https://doi.org/10.1002/9783527809080.cataz13886}.

\bibitem[Dinev \& Gutmann(2018)Dinev and Gutmann]{dire}
Dinev, T. and Gutmann, M.~U.
\newblock Dynamic likelihood-free inference via ratio estimation (dire).
\newblock \emph{arXiv preprint arXiv:1810.09899}, 2018.

\bibitem[Duane et~al.(1987)Duane, Kennedy, Pendleton, and Roweth]{hmcoriginal}
Duane, S., Kennedy, A., Pendleton, B.~J., and Roweth, D.
\newblock Hybrid {Monte} {Carlo}.
\newblock \emph{Phys. Lett. B}, 195\penalty0 (2):\penalty0 216--222, September
  1987.
\newblock ISSN 0370-2693.
\newblock URL \url{https://doi.org/10.1016/0370-2693(87)91197-x}.

\bibitem[{Durkan} et~al.(2020){Durkan}, {Murray}, and
  {Papamakarios}]{2020arXiv200203712D}
{Durkan}, C., {Murray}, I., and {Papamakarios}, G.
\newblock {On Contrastive Learning for Likelihood-free Inference}.
\newblock \emph{arXiv e-prints}, art. arXiv:2002.03712, February 2020.

\bibitem[Dutta et~al.(2016)Dutta, Corander, Kaski, and
  Gutmann]{dutta2016likelihood}
Dutta, R., Corander, J., Kaski, S., and Gutmann, M.~U.
\newblock Likelihood-free inference by ratio estimation.
\newblock \emph{arXiv preprint arXiv:1611.10242}, 2016.

\bibitem[Fearnhead \& Prangle(2012)Fearnhead and
  Prangle]{fearnhead2012constructing}
Fearnhead, P. and Prangle, D.
\newblock Constructing summary statistics for approximate {Bayesian}
  computation: {Semi-automatic} approximate {Bayesian} computation.
\newblock \emph{Journal of the Royal Statistical Society: Series B (Statistical
  Methodology)}, 74\penalty0 (3):\penalty0 419--474, May 2012.
\newblock ISSN 1369-7412.
\newblock URL \url{https://doi.org/10.1111/j.1467-9868.2011.01010.x}.

\bibitem[Gershman \& Goodman(2014)Gershman and Goodman]{amortized}
Gershman, S. and Goodman, N.
\newblock Amortized inference in probabilistic reasoning.
\newblock In \emph{Proceedings of the Annual Meeting of the Cognitive Science
  Society}, volume~36, 2014.

\bibitem[Goodfellow et~al.(2014)Goodfellow, Pouget-Abadie, Mirza, Xu,
  Warde-Farley, Ozair, Courville, and Bengio]{goodfellow2014generative}
Goodfellow, I., Pouget-Abadie, J., Mirza, M., Xu, B., Warde-Farley, D., Ozair,
  S., Courville, A., and Bengio, Y.
\newblock Generative adversarial nets.
\newblock In \emph{Advances in neural information processing systems}, pp.\
  2672--2680, 2014.

\bibitem[Greenberg et~al.(2019)Greenberg, Nonnenmacher, and Macke]{apt}
Greenberg, D., Nonnenmacher, M., and Macke, J.
\newblock Automatic posterior transformation for likelihood-free inference.
\newblock In Chaudhuri, K. and Salakhutdinov, R. (eds.), \emph{Proceedings of
  the 36th International Conference on Machine Learning}, volume~97 of
  \emph{Proceedings of Machine Learning Research}, pp.\  2404--2414, Long
  Beach, California, USA, 2019. PMLR.
\newblock URL \url{http://proceedings.mlr.press/v97/greenberg19a.html}.

\bibitem[Gretton et~al.(2012)Gretton, Borgwardt, Rasch, Sch{\"o}lkopf, and
  Smola]{mmd}
Gretton, A., Borgwardt, K.~M., Rasch, M.~J., Sch{\"o}lkopf, B., and Smola, A.
\newblock A kernel two-sample test.
\newblock \emph{Journal of Machine Learning Research}, 13\penalty0
  (Mar):\penalty0 723--773, 2012.

\bibitem[Gutmann \& Corander(2016)Gutmann and Corander]{bolfi}
Gutmann, M.~U. and Corander, J.
\newblock {Bayesian} optimization for likelihood-free inference of
  simulator-based statistical models.
\newblock \emph{The Journal of Machine Learning Research}, 17\penalty0
  (1):\penalty0 4256--4302, 2016.
\newblock URL \url{http://jmlr.org/papers/v17/15-017.html}.

\bibitem[Gutmann et~al.(2017)Gutmann, Dutta, Kaski, and
  Corander]{classifierabc}
Gutmann, M.~U., Dutta, R., Kaski, S., and Corander, J.
\newblock Likelihood-free inference via classification.
\newblock \emph{Stat Comput}, 28\penalty0 (2):\penalty0 411--425, March 2017.
\newblock ISSN 0960-3174, 1573-1375.
\newblock URL \url{https://doi.org/10.1007/s11222-017-9738-6}.

\bibitem[Hastings(1970)]{hastings1970monte}
Hastings, W.
\newblock {Monte} {Carlo} sampling methods using {Markov} chains and their
  applications.
\newblock \emph{Biometrika}, 57\penalty0 (1):\penalty0 97--109, April 1970.
\newblock ISSN 1464-3510, 0006-3444.
\newblock URL \url{https://doi.org/10.1093/biomet/57.1.97}.

\bibitem[He et~al.(2016)He, Zhang, Ren, and Sun]{resnet}
He, K., Zhang, X., Ren, S., and Sun, J.
\newblock Deep residual learning for image recognition.
\newblock In \emph{2016 IEEE Conference on Computer Vision and Pattern
  Recognition (CVPR)}, pp.\  770--778. IEEE, June 2016.
\newblock URL \url{https://doi.org/10.1109/cvpr.2016.90}.

\bibitem[Hoffman et~al.(2013)Hoffman, Blei, Wang, and
  Paisley]{hoffman2013stochastic}
Hoffman, M.~D., Blei, D.~M., Wang, C., and Paisley, J.
\newblock Stochastic variational inference.
\newblock \emph{The Journal of Machine Learning Research}, 14\penalty0
  (1):\penalty0 1303--1347, 2013.

\bibitem[Ioffe \& Szegedy(2015)Ioffe and Szegedy]{batchnorm}
Ioffe, S. and Szegedy, C.
\newblock Batch normalization: {Accelerating} deep network training by reducing
  internal covariate shift.
\newblock \emph{arXiv preprint arXiv:1502.03167}, 2015.

\bibitem[J.~Neyman(1933)]{neymanpearson}
J.~Neyman, E.~P.
\newblock {IX.} on the problem of the most efficient tests of statistical
  hypotheses.
\newblock \emph{Phil. Trans. R. Soc. Lond. A}, 231\penalty0 (694-706):\penalty0
  289--337, February 1933.
\newblock ISSN 0264-3952, 2053-9258.
\newblock URL \url{https://doi.org/10.1098/rsta.1933.0009}.

\bibitem[Kingma \& Ba(2014)Kingma and Ba]{adam}
Kingma, D.~P. and Ba, J.
\newblock Adam: {A} method for stochastic optimization.
\newblock \emph{arXiv preprint arXiv:1412.6980}, 2014.

\bibitem[Klambauer et~al.(2017)Klambauer, Unterthiner, Mayr, and
  Hochreiter]{selu}
Klambauer, G., Unterthiner, T., Mayr, A., and Hochreiter, S.
\newblock Self-normalizing neural networks.
\newblock In \emph{Advances in Neural Information Processing Systems}, pp.\
  971--980, 2017.

\bibitem[Kormann et~al.(1994)Kormann, Schneider, and
  Bartelmann]{ellipticalisothermal}
Kormann, R., Schneider, P., and Bartelmann, M.
\newblock Isothermal elliptical gravitational lens models.
\newblock \emph{Astron. Astrophys.}, 284:\penalty0 285--299, 1994.

\bibitem[Lecun et~al.(1998)Lecun, Bottou, Bengio, and Haffner]{lenet}
Lecun, Y., Bottou, L., Bengio, Y., and Haffner, P.
\newblock Gradient-based learning applied to document recognition.
\newblock \emph{Proc. IEEE}, 86\penalty0 (11):\penalty0 2278--2324, 1998.
\newblock ISSN 0018-9219.
\newblock URL \url{https://doi.org/10.1109/5.726791}.

\bibitem[Lotka(1920)]{lotka}
Lotka, A.~J.
\newblock Analytical note on certain rhythmic relations in organic systems.
\newblock \emph{Proc Natl Acad Sci USA}, 6\penalty0 (7):\penalty0 410--415,
  June 1920.
\newblock ISSN 0027-8424, 1091-6490.
\newblock URL \url{https://doi.org/10.1073/pnas.6.7.410}.

\bibitem[Louppe et~al.(2017)Louppe, Hermans, and Cranmer]{avo}
Louppe, G., Hermans, J., and Cranmer, K.
\newblock Adversarial variational optimization of non-differentiable
  simulators.
\newblock \emph{arXiv preprint arXiv:1707.07113}, 2017.
\newblock URL \url{https://arxiv.org/abs/1707.07113}.

\bibitem[Lueckmann et~al.(2018)Lueckmann, Bassetto, Karaletsos, and
  Macke]{emulatornetworks}
Lueckmann, J.-M., Bassetto, G., Karaletsos, T., and Macke, J.~H.
\newblock Likelihood-free inference with emulator networks.
\newblock \emph{arXiv preprint arXiv:1805.09294}, 2018.
\newblock URL \url{https://arxiv.org/abs/1805.09294}.

\bibitem[MacKay(2003)]{mackay2003information}
MacKay, D.~J.
\newblock \emph{Information theory, inference and learning algorithms}.
\newblock Cambridge university press, 2003.

\bibitem[Marin et~al.(2011)Marin, Pudlo, Robert, and
  Ryder]{marin2012approximate}
Marin, J.-M., Pudlo, P., Robert, C.~P., and Ryder, R.~J.
\newblock Approximate {Bayesian} computational methods.
\newblock \emph{Stat Comput}, 22\penalty0 (6):\penalty0 1167--1180, October
  2011.
\newblock ISSN 0960-3174, 1573-1375.
\newblock URL \url{https://doi.org/10.1007/s11222-011-9288-2}.

\bibitem[Marjoram et~al.(2003)Marjoram, Molitor, Plagnol, and Tavare]{abcmcmc}
Marjoram, P., Molitor, J., Plagnol, V., and Tavare, S.
\newblock {Markov} chain {Monte} {Carlo} without likelihoods.
\newblock \emph{Proceedings of the National Academy of Sciences}, 100\penalty0
  (26):\penalty0 15324--15328, December 2003.
\newblock ISSN 0027-8424, 1091-6490.
\newblock URL \url{https://doi.org/10.1073/pnas.0306899100}.

\bibitem[Meeds \& Welling(2014)Meeds and Welling]{synlikelihood1}
Meeds, E. and Welling, M.
\newblock {GPS}-{ABC:} {Gaussian} process surrogate approximate {Bayesian}
  computation.
\newblock \emph{arXiv preprint arXiv:1401.2838}, 2014.
\newblock URL \url{https://arxiv.org/abs/1401.2838}.

\bibitem[Metropolis et~al.(1953)Metropolis, Rosenbluth, Rosenbluth, Teller, and
  Teller]{metropolis}
Metropolis, N., Rosenbluth, A.~W., Rosenbluth, M.~N., Teller, A.~H., and
  Teller, E.
\newblock Equation of state calculations by fast computing machines.
\newblock \emph{The Journal of Chemical Physics}, 21\penalty0 (6):\penalty0
  1087--1092, June 1953.
\newblock ISSN 0021-9606, 1089-7690.
\newblock URL \url{https://doi.org/10.1063/1.1699114}.

\bibitem[{Mohamed} \& {Lakshminarayanan}(2016){Mohamed} and
  {Lakshminarayanan}]{2016arXiv161003483M}
{Mohamed}, S. and {Lakshminarayanan}, B.
\newblock {Learning in Implicit Generative Models}.
\newblock \emph{ArXiv e-prints}, October 2016.

\bibitem[Neal(2011)]{hmc}
Neal, R.~M.
\newblock {MCMC} using {{H}amiltonian} dynamics.
\newblock \emph{Handbook of Markov Chain Monte Carlo}, 2\penalty0
  (11):\penalty0 2, 2011.
\newblock URL \url{https://arxiv.org/abs/1206.1901}.

\bibitem[Neal \& Hinton(1998)Neal and Hinton]{neal1998view}
Neal, R.~M. and Hinton, G.~E.
\newblock A view of the em algorithm that justifies incremental, sparse, and
  other variants.
\newblock In \emph{Learning in Graphical Models}, pp.\  355--368. Springer
  Netherlands, 1998.
\newblock URL \url{https://doi.org/10.1007/978-94-011-5014-9_12}.

\bibitem[Nightingale et~al.(2018)Nightingale, Dye, and Massey]{autolens}
Nightingale, J.~W., Dye, S., and Massey, R.~J.
\newblock {AutoLens:} {Automated} modeling of a strong lens's light, mass, and
  source.
\newblock \emph{Mon. Not. R. Astron. Soc.}, 478\penalty0 (4):\penalty0
  4738--4784, May 2018.
\newblock ISSN 0035-8711, 1365-2966.
\newblock URL \url{https://doi.org/10.1093/mnras/sty1264}.

\bibitem[Ong et~al.(2017)Ong, Nott, Tran, Sisson, and Drovandi]{synlikelihood0}
Ong, V.~M., Nott, D.~J., Tran, M.-N., Sisson, S.~A., and Drovandi, C.~C.
\newblock Variational {Bayes} with synthetic likelihood.
\newblock \emph{Stat Comput}, 28\penalty0 (4):\penalty0 971--988, August 2017.
\newblock ISSN 0960-3174, 1573-1375.
\newblock URL \url{https://doi.org/10.1007/s11222-017-9773-3}.

\bibitem[Papamakarios \& Murray(2016)Papamakarios and Murray]{snpea}
Papamakarios, G. and Murray, I.
\newblock Fast $\varepsilon $-free inference of simulation models with
  {Bayesian} conditional density estimation.
\newblock In \emph{Advances in Neural Information Processing Systems}, pp.\
  1028--1036, 2016.

\bibitem[Papamakarios \& Murray(2018)Papamakarios and Murray]{snl}
Papamakarios, G. and Murray, I.
\newblock Sequential neural likelihood: {Fast} likelihood-free inference with
  autoregressive flows.
\newblock \emph{arXiv preprint arXiv:1805.07226}, 2018.
\newblock URL \url{https://arxiv.org/abs/1805.07226}.

\bibitem[Paszke et~al.(2017)Paszke, Gross, Chintala, Chanan, Yang, DeVito, Lin,
  Desmaison, Antiga, and Lerer]{pytorch}
Paszke, A., Gross, S., Chintala, S., Chanan, G., Yang, E., DeVito, Z., Lin, Z.,
  Desmaison, A., Antiga, L., and Lerer, A.
\newblock Automatic differentiation in pytorch.
\newblock 2017.

\bibitem[Pesah et~al.(2018)Pesah, Wehenkel, and Louppe]{pesah}
Pesah, A., Wehenkel, A., and Louppe, G.
\newblock Recurrent machines for likelihood-free inference.
\newblock \emph{arXiv preprint arXiv:1811.12932}, 2018.

\bibitem[Pritchard et~al.(1999)Pritchard, Seielstad, Perez-Lezaun, and
  Feldman]{abc2}
Pritchard, J., Seielstad, M., Perez-Lezaun, A., and Feldman, M.
\newblock Population growth of human y chromosomes: {A} study of y chromosome
  microsatellites.
\newblock \emph{Mol. Biol. Evol.}, 16\penalty0 (12):\penalty0 1791--1798,
  December 1999.
\newblock ISSN 0737-4038, 1537-1719.
\newblock URL \url{https://doi.org/10.1093/oxfordjournals.molbev.a026091}.

\bibitem[Ritchie et~al.(2016)Ritchie, Horsfall, and Goodman]{amortizeddeep}
Ritchie, D., Horsfall, P., and Goodman, N.~D.
\newblock Deep amortized inference for probabilistic programs.
\newblock \emph{arXiv preprint arXiv:1610.05735}, 2016.

\bibitem[Salimans et~al.(2015)Salimans, Kingma, and Welling]{salimansmarkov}
Salimans, T., Kingma, D., and Welling, M.
\newblock {Markov} chain monte carlo and variational inference: {Bridging} the
  gap.
\newblock In \emph{International Conference on Machine Learning}, pp.\
  1218--1226, 2015.

\bibitem[Sjöstrand et~al.(2008)Sjöstrand, Mrenna, and Skands]{pythia}
Sjöstrand, T., Mrenna, S., and Skands, P.
\newblock A brief introduction to {PYTHIA} 8.1.
\newblock \emph{Comput. Phys. Commun.}, 178\penalty0 (11):\penalty0 852--867,
  June 2008.
\newblock ISSN 0010-4655.
\newblock URL \url{https://doi.org/10.1016/j.cpc.2008.01.036}.

\bibitem[Skands et~al.(2014)Skands, Carrazza, and Rojo]{monash}
Skands, P., Carrazza, S., and Rojo, J.
\newblock Tuning {PYTHIA} 8.1: {The} monash 2013 tune.
\newblock \emph{Eur. Phys. J. C}, 74\penalty0 (8):\penalty0 3024, August 2014.
\newblock ISSN 1434-6044, 1434-6052.
\newblock URL \url{https://doi.org/10.1140/epjc/s10052-014-3024-y}.

\bibitem[Sutton et~al.(2000)Sutton, McAllester, Singh, and
  Mansour]{policygradient}
Sutton, R.~S., McAllester, D.~A., Singh, S.~P., and Mansour, Y.
\newblock Policy gradient methods for reinforcement learning with function
  approximation.
\newblock In \emph{Advances in neural information processing systems}, pp.\
  1057--1063, 2000.

\bibitem[Talts et~al.(2018)Talts, Betancourt, Simpson, Vehtari, and
  Gelman]{sbc}
Talts, S., Betancourt, M., Simpson, D., Vehtari, A., and Gelman, A.
\newblock Validating {Bayesian} inference algorithms with simulation-based
  calibration.
\newblock \emph{arXiv preprint arXiv:1804.06788}, 2018.

\bibitem[Tavar{\'e} et~al.(1997)Tavar{\'e}, Balding, Griffiths, and
  Donnelly]{abc0}
Tavar{\'e}, S., Balding, D.~J., Griffiths, R.~C., and Donnelly, P.
\newblock Inferring coalescence times from {DNA} sequence data.
\newblock \emph{Genetics}, 145\penalty0 (2):\penalty0 505--518, 1997.
\newblock URL
  \url{http://www.genetics.org/content/genetics/145/2/505.full.pdf}.

\bibitem[Toni et~al.(2008)Toni, Welch, Strelkowa, Ipsen, and Stumpf]{smcabc}
Toni, T., Welch, D., Strelkowa, N., Ipsen, A., and Stumpf, M.~P.
\newblock Approximate {Bayesian} computation scheme for parameter inference and
  model selection in dynamical systems.
\newblock \emph{J. R. Soc. Interface.}, 6\penalty0 (31):\penalty0 187--202,
  July 2008.
\newblock ISSN 1742-5689, 1742-5662.
\newblock URL \url{https://doi.org/10.1098/rsif.2008.0172}.

\bibitem[Tran et~al.(2017)Tran, Ranganath, and Blei]{tran2017hierarchical}
Tran, D., Ranganath, R., and Blei, D.
\newblock Hierarchical implicit models and likelihood-free variational
  inference.
\newblock In \emph{Advances in Neural Information Processing Systems}, pp.\
  5523--5533, 2017.

\bibitem[Turner et~al.(2018)Turner, Hung, Saatci, and
  Yosinski]{turner2018metropolis}
Turner, R., Hung, J., Saatci, Y., and Yosinski, J.
\newblock Metropolis-{Hastings} generative adversarial networks.
\newblock \emph{arXiv preprint arXiv:1811.11357}, 2018.

\bibitem[Uehara et~al.(2016)Uehara, Sato, Suzuki, Nakayama, and
  Matsuo]{uehara2016generative}
Uehara, M., Sato, I., Suzuki, M., Nakayama, K., and Matsuo, Y.
\newblock Generative adversarial nets from a density ratio estimation
  perspective.
\newblock \emph{arXiv preprint arXiv:1610.02920}, 2016.

\bibitem[Wegmann et~al.(2009)Wegmann, Leuenberger, and Excoffier]{abcmcmc2}
Wegmann, D., Leuenberger, C., and Excoffier, L.
\newblock Efficient approximate {Bayesian} computation coupled with {Markov}
  chain {Monte} {Carlo} without likelihood.
\newblock \emph{Genetics}, 182\penalty0 (4):\penalty0 1207--1218, June 2009.
\newblock ISSN 0016-6731, 1943-2631.
\newblock URL \url{https://doi.org/10.1534/genetics.109.102509}.

\bibitem[Williams(1992)]{reinforce}
Williams, R.~J.
\newblock Simple statistical gradient-following algorithms for connectionist
  reinforcement learning.
\newblock \emph{Mach Learn}, 8\penalty0 (3-4):\penalty0 229--256, May 1992.
\newblock ISSN 0885-6125, 1573-0565.
\newblock URL \url{https://doi.org/10.1007/bf00992696}.

\bibitem[Wong et~al.(2018)Wong, Jiang, Wu, and Zheng]{jiang2017learning}
Wong, W., Jiang, B., Wu, T.-y., and Zheng, C.
\newblock Learning summary statistic for approximate {Bayesian} computation via
  deep neural network.
\newblock \emph{STAT SINICA}, pp.\  1595--1618, 2018.
\newblock ISSN 1017-0405.
\newblock URL \url{https://doi.org/10.5705/ss.202015.0340}.

\end{thebibliography}
\bibliographystyle{icml2020}


\appendix
\onecolumn

\section{Likelihood-free Markov chain Monte Carlo samplers}
\label{sec:algorithms}

\begin{algorithm}
  \scriptsize
  \caption{Likelihood-free Metropolis-Hastings}
  \label{algo:metropolis_hastings}
  \begin{tabular}{ l l }
    {\it Inputs:} & Initial parameter $\btheta_0$ \\
                  & Prior $p(\btheta)$ \\
                  & Transition distribution $q(\btheta)$ \\
                  & Trained ratio estimator $\bs(\bx, \btheta)$ \\
                  & Observation $\bx$ \\
    {\it Outputs:} & Markov chain $\btheta_{0:T}$ \\
    {\it Hyperparameters:} & Steps $T$ \\
  \end{tabular}

  \begin{algorithmic}[1]
    \State{$t \gets 0$}
    \State{$\btheta_t \gets \btheta_0$}
    \For{$t < T$}
    \State $\btheta' \sim q(\btheta \vert \btheta_t)$
    \State $\displaystyle\lambda \gets (\log\hat{r}(\bx\vert\btheta{'}) + \log p(\btheta{'})) - (\log\hat{r}(\bx\vert{\btheta}_t) + \log p({\btheta}_t))$
    \State $\displaystyle\rho \gets \min(\exp(\lambda) \frac{q({\btheta}_t \vert {\btheta}')}{q({\btheta}' \vert {\btheta}_t)},~1)$
    \State $\btheta_{t + 1} \gets \begin{cases}
    \btheta' & \text{with probability } \rho \\
    \btheta_t & \text{with probability } 1 - \rho
    \end{cases}$
    \State $t \gets t + 1$
    \EndFor
    \State \Return{$\btheta_{0:T}$}
  \end{algorithmic}
\end{algorithm}
\begin{algorithm}
  \scriptsize
  \caption{Likelihood-free Hamiltonian Monte Carlo}
  \label{algo:hmc}
  \begin{tabular}{ l l }
    {\it Inputs:} & Initial parameter $\btheta_0$ \\
                  & Prior $p(\btheta)$ \\
                  & Momentum distribution $q(\bM)$ \\
                  & Trained ratio estimator $\bs(\bx, \btheta)$ \\
                  & Observation $\bx$ \\
    {\it Outputs:} & Markov chain $\btheta_{0:T}$ \\
    {\it Hyperparameters:} & Steps $T$. \\
                           & Leapfrog-integration steps $l$ and stepsize $\eta$. \\
  \end{tabular}
  \\
  \begin{algorithmic}[1]
    \State{$t \gets 0$}
    \State{$\btheta_t \gets \btheta_0$}
    \For{$t < T$}
    \State $\bM_t \sim q(\bM)$
    \State $k \gets 0$
    \State ${\bM}_k \gets \bM_t$
    \State ${\btheta}_k \gets {\btheta}_t$
    \For{$k < l$}
    \State $\displaystyle{\bM}_k \gets {\bM}_k + \frac{\eta}{2}\frac{\nabla_{\btheta}~\hat{r}(\bx\vert\btheta_k)}{\hat{r}(\bx\vert{\btheta}_k)}$
    \State $\displaystyle{\btheta}_k \gets {\btheta}_k + \eta{\bM}_k$
    \State $\displaystyle{\bM}_k \gets {\bM}_k + \frac{\eta}{2}\frac{\nabla_{\btheta}~\hat{r}(\bx\vert{\btheta}_k)}{\hat{r}(\bx\vert{\btheta}_k)}$
    \State $k \gets k + 1$
    \EndFor
    \State $\lambda \gets (\log\hat{r}(\bx\vert{\btheta}_k) + \log p({\btheta}_k)) - (\log\hat{r}(\bx\vert{\btheta}_t) + \log p({\btheta}_t)) + K({\bM}_k) - K({\bM}_t)$
    \State $\rho \gets \min(\exp(\lambda), 1)$
    \State $\btheta_{t + 1} \gets \begin{cases}
    {\btheta}_k & \text{with probability } \rho \\
    \btheta_t & \text{with probability } 1 - \rho
    \end{cases}$
    \State $t \gets t + 1$
    \EndFor
    \State \Return{$\btheta_{0:T}$}
  \end{algorithmic}
\end{algorithm}

\newpage
\onecolumn
\section{Correctness of Algorithm~\ref{algo:optimization}}
\label{sec:proof}

The core of our contribution rests on the proper estimation of the likelihood-to-evidence ratio.
In this section, we show that  the minimization of the binary cross-entropy (\textsc{bce}) loss of a classifier tasked to distinguish between dependent input pairs $(\bx,\btheta) \sim p(\bx,\btheta)$ and independent input pairs $(\bx,\btheta) \sim p(\bx)p(\btheta)$ results in an optimal classifier.

Using calculus of variations and reproducing the structure of Algorithm~\ref{algo:optimization}, we define the loss functional
\begin{align}
  \begin{split}
    L[\bs(\bx,\btheta)] &= \int d\btheta \int d\bx \int d\bm{\theta}' p(\btheta) p(\bx\vert\btheta) p(\bm{\theta}') \Big[ -\log \bs(\bx,\btheta) - \log(1-(\bs(\bx,\bm{\theta}')) \Big] \\
    &= \int d\btheta \int d\bx \underbrace{p(\btheta) p(\bx\vert\btheta) \Big [ -\log \bs(\bx,\btheta) \Big] + p(\btheta) p(\bx) \Big[ - \log(1-\bs(\bx,\btheta)) \Big]}_{F(\bs)}.
    \end{split}
\end{align}
This loss functional is minimized for a function $\bsopt(\bx,\btheta)$ such that
\begin{align}
    0 = \frac{\delta F}{\delta \bs}\Biggr|_{\bsopt} = p(\btheta)p(\bx\vert\btheta) \Big[ -\frac{1}{\bsopt(\bx,\btheta)} \Big] +  p(\btheta)p(\bx) \Big[ \frac{1}{1-\bsopt(\bx,\btheta)} \Big].
\end{align}
As long as $p(\btheta)>0$, this is equivalent to
\begin{align}
    p(\bx\vert\btheta)  \frac{1}{\bsopt(\bx,\btheta)} = p(\bx) \frac{1}{1-\bsopt(\bx,\btheta)},
\end{align}
and finally
\begin{align}
    \bsopt(\bx,\btheta) = \frac{p(\bx\vert\btheta)}{p(\bx\vert\btheta) + p(\bx)}.
\end{align}\hfill\ensuremath{\square}

\newpage

\section{Recommended strategy for applications}
\label{sec:recommended}

This section discusses several recommended strategies to successfully apply our technique to (scientific) applications. We show several code listings, with a focus on a Pytorch~\cite{pytorch} implementation. A reference implementation can be found at \url{https://github.com/montefiore-ai/hypothesis}. As shown in Figure~\ref{fig:architecture}, we directly output the log ratio before applying the sigmoidal projection to improve numerical stability when sampling from the posterior using \textsc{mcmc}. The output of the decision function $\bs(\bx,\btheta)$ is also given. This architecture forms the basis for accurate posterior inference. Figure~\ref{lst:ratio_estimator} shows the base ratio estimator implementation. For completeness, the variable name \texttt{inputs} relates to the model parameters $\btheta$ while \texttt{outputs} relates to observations $\bx$. This particular naming scheme is chosen to depict their relation with respect to the simulation model.

\begin{figure*}[h]
  \centering
  \usetikzlibrary{arrows}
  \def\layersep{1cm}
  \begin{scaletikzpicturetowidth}{12cm}
  \begin{tikzpicture}[shorten >= 1pt, ->, node distance=\layersep,scale=\tikzscale]
  \tikzstyle{neuron} = [circle, minimum size=0.25cm, draw=black!20, line width=0.3mm, fill=white]

  \draw[rounded corners] (-1,-0.5) rectangle (4,-5.5);
  \path[->, shorten >= 0pt] (-2,-4) edge (-1,-4);
  \node[left] at (-2,-4) {$\btheta$};
  \path[->, shorten >= 0pt] (-2,-2) edge (-1,-2);
  \node[left] at (-2,-2) {$\bx$};
  \path[-, shorten >= 0pt] (3.5, -3) edge (3.5, -4);
  \node[right] at (5.8,-4) (S) {$\bs(\bx, \btheta)$};
  \path[->, shorten >= 0pt] (3.5,-4) edge (S);
  \node[right] at (4.2,-2) (R) {$\hat{r}(\bx\vert\btheta) = \frac{\bs(\bx, \btheta)}{1 - \bs(\bx, \btheta)}$};
  \path[->, shorten >= 0pt] (S) edge (R);
  \foreach \name / \y in {1,...,3}
      \node[neuron] (f-I-\name) at (-0.5,-1-\y) {};
  \foreach \name / \y in {1,...,5}
      \node[neuron] (f-H1-\name) at (-0.5cm+\layersep,-\y cm) {};
  \foreach \name / \y in {1,...,5}
      \node[neuron] (f-H2-\name) at (-0.5cm+3*\layersep,-\y cm) {};
  \node[neuron] (f-O) at (-0.5cm+4*\layersep,-3cm) {};
  \foreach \source in {1,...,3}
      \foreach \dest in {1,...,5}
          \path[-, black] (f-I-\source) edge (f-H1-\dest);
  \foreach \source in {1,...,5}
      \path[-, black] (f-H2-\source) edge (f-O);
  \node[black] at (1.5,-3) {...};

  \begin{scope}[shift={(14,0)}]
  \draw[rounded corners] (-1,-0.5) rectangle (8.3,-5.5);
  \path[->, shorten >= 0pt] (-2,-4) edge (-1,-4);
  \node[left] at (-2,-4) {$\btheta$};
  \path[->, shorten >= 0pt] (-2,-2) edge (-1,-2);
  \node[left] at (-2,-2) {$\bx$};
  \path[-, shorten >= 0pt] (3.5, -3) edge (3.5, -2);
  \node[right] at (4,-2) (R) {$\log\hat{r}(\bx\vert\btheta)$};
  \node[right] at (3.54,-4) (S) {$\sigma(\log\hat{r}(\bx\vert\btheta))$};
  \node[right] at (8.8, -2) (R_out) {$\log\hat{r}(\bx\vert\btheta)$};
  \node[right] at (9.1, -4) (S_out) {$\bs(\bx,\btheta)$};
  \path[->, shorten >= 0pt] (3.5,-2) edge (R);
  \path[->, shorten >= 0pt] (R) edge (S);
  \path[->, shorten >= 0pt] (R) edge (R_out);
  \path[->, shorten >= 0pt] (S) edge (S_out);
  \foreach \name / \y in {1,...,3}
      \node[neuron] (f-I-\name) at (-0.5,-1-\y) {};
  \foreach \name / \y in {1,...,5}
      \node[neuron] (f-H1-\name) at (-0.5cm+\layersep,-\y cm) {};
  \foreach \name / \y in {1,...,5}
      \node[neuron] (f-H2-\name) at (-0.5cm+3*\layersep,-\y cm) {};
  \node[neuron] (f-O) at (-0.5cm+4*\layersep,-3cm) {};
  \foreach \source in {1,...,3}
      \foreach \dest in {1,...,5}
          \path[-, black] (f-I-\source) edge (f-H1-\dest);
  \foreach \source in {1,...,5}
      \path[-, black] (f-H2-\source) edge (f-O);
  \node[black] at (1.5,-3) {...};
  \end{scope}
  \end{tikzpicture}
  \end{scaletikzpicturetowidth}
   \caption{Two approaches to extract the approximate ratio $\hat{r}(\bx\vert\btheta)$ from a parameterized classifier. {\it (Left):} The vanilla architecture which is susceptible to numerical errors and loss of information as $\hat{r}(\bx\vert\btheta)$ is computed by transforming the sigmoidal projection $\sigma$. This issue arises if the classifier is (almost) able to perfectly discriminate between samples from $p(\bx\vert\btheta)$ and $p(\bx\vert\bthetaref)$. {\it (Right):} The modified architecture directly outputs $\log\hat{r}(\bx\vert\btheta)$ before applying the sigmoidal projection.}
   \label{fig:architecture}
\end{figure*}
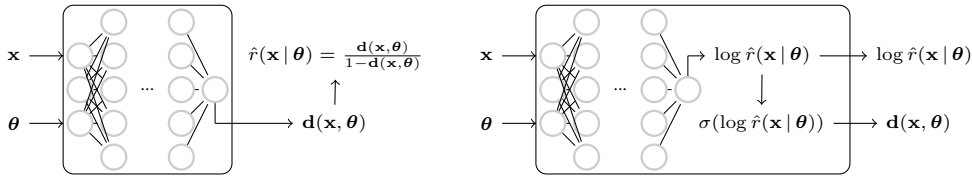

\begin{figure}[h]
  \begin{lstlisting}[language=python]
class RatioEstimator(torch.nn.Module):
    def __init__(self):
        super(RatioEstimator, self).__init__()
        self.network = (...)  # Define your neural network

    def forward(self, inputs, outputs):
         # Process the inputs (model parameters)
         (...)
         
         # Process the outputs (observations)
         (...)
         
         log_ratio = self.network(inputs, outputs)
         classifier_output = log_ratio.sigmoid()

         return classifier_output, log_ratio\end{lstlisting}
  \caption{Base ratio estimator.}
  \label{lst:ratio_estimator}
\end{figure}

\subsection{Dataset generation and training}

Algorithm~\ref{algo:optimization} actively samples from the prior and the simulation model in the optimization loop. This is not efficient in practice. A dataset consisting of samples from the joint can be generated offline before training the ratio estimators. Note that no class labels are assigned to specific samples of the dataset. In our training algorithm, the independence of $\bx$ and $\btheta$ can be guaranteed by sampling two batches from the dataset, and simply switch the $\btheta$ tensors in the computation of each individual loss. Alternatively, the mini-batch containing $\btheta$ or $\bx$ can be randomly shuffled. As a result, the implementation of the optimization loop does not depend on the prior.
This produces a mathematically equivalent procedure to Algorithm~\ref{algo:optimization}.  Figure~\ref{lst:optimization} shows a \texttt{pytorch} implementation of the proposed optimization loop for an even number of batches.

The optimal discriminator is the one which minimizes the training criterion described in Algorithm~\ref{algo:optimization}. Therefore, the accuracy of the approximation can be improved by using techniques such as learning rate scheduling, or by increasing the batch size to reduce the variance of the gradient. From an architectural perspective, we found that the \textsc{elu}~\cite{elu} and \textsc{selu}~\cite{selu} activations work well in general. However, \textsc{relu}s typically required significantly less parameters (weights) to accurately approximate sharp posteriors. We hypothesize that this behavior is attributable to the sparsity induced by \textsc{relu} activations. We did not perform a study on the required number of simulations to properly approximate $r(\bx\vert\btheta)$. This aspect is left for future work.

\begin{figure}[h]
  \begin{lstlisting}[language=python]
loader = DataLoader(dataset, batch_size=batch_size)
num_iterations = len(loader) // 2
loader = iter(loader)
ratio_estimator.train()

for batch_index in range(num_iterations):
    # Load the data and move to the device.
    a_inputs, a_outputs = next(loader)
    a_inputs = a_inputs.to(device, non_blocking=True)
    a_outputs = a_outputs.to(device, non_blocking=True)
    b_inputs, b_outputs = next(loader)
    b_inputs = b_inputs.to(device, non_blocking=True)
    b_outputs = b_outputs.to(device, non_blocking=True)
    
    # Apply a forward pass with the ratio estimator.
    y_dep_a, _ = ratio_estimator(a_inputs, a_outputs)
    y_idep_a, _ = ratio_estimator(a_inputs, b_outputs)
    y_dep_b, _ = ratio_estimator(b_inputs, b_outputs)
    y_idep_b, _ = ratio_estimator(b_inputs, a_outputs)
    
    # Loss and backward.
    loss_a = criterion(y_dep_a, ones) +
             criterion(y_idep_a, zeros)
    loss_b = criterion(y_dep_b, ones) +
             criterion(y_idep_b, zeros)
    loss = loss_a + loss_b
    optimizer.zero_grad()
    loss.backward()
    optimizer.step()\end{lstlisting}
  \caption{Proposed optimization loop.}
  \label{lst:optimization}
\end{figure}

\subsection{Validation and inference}
In general, we recommend to train multiple (if the computational budget allows) ratio estimators. Besides the improvements that ensembling typically brings, the resulting estimators can be used to determine the variance of the approximation. From our empirical evaluations, large variances in the approximation of the likelihood-to-evidence ratio indicate that the capacity of the ratio estimator might be insufficient (see Appendix~\ref{sec:capacity}). As mentioned in Section~\ref{sec:roc}, the accuracy of the approximation can be verified in a more principled way by means of a \textsc{roc} curve and its \textsc{auc}. Contrary to the experimental section of the main manuscript, real applications do not have access to the generating parameters $\btheta^*$. There are currently two approaches to test the accuracy of the ratio estimator using the \textsc{roc} diagnostic: \begin{enumerate*}[label=(\roman*)]
\item test the ratio estimator for all distinct modes of the posterior and
  \item test the ratio estimator for a set of random samples $\btheta\sim p(\btheta)$
\end{enumerate*}. While the first approach specifically tests the solution, the latter validates the behavior of the ratio estimator across the prior $p(\btheta)$. Alternatively, Simulation Based Calibration~\cite{sbc} is a frequentist test for a Bayesian computation, but it cannot verify the accuracy of a single approximate posterior. After validating the ratio estimator, \textsc{mcmc} can be used to draw samples from the posterior. If the dimensionality of the problem permits, estimates of the \textsc{pdf} can be obtained directly.

\newpage
\section{Experimental details and additional results}
\label{sec:experimental_details}

\subsection{Overview of hyperparameters and model architectures}
\label{sec:summary_hyperparameters}
\begin{table}[h]
  \centering
  \begin{tabular}{llllll}
    \toprule
    Hyperparameter & Tractable problem & Detector calibration & Lensing & Lotka-Volterra & \textsc{m/g/1} \\
    \midrule
    Activation function & \textsc{selu} & \textsc{selu} & \textsc{relu}      & \textsc{relu} & \textsc{relu} \\
    \textsc{amsgrad}    & Yes           & Yes           & Yes                & Yes           & Yes           \\
    Architecture        & \textsc{mlp}  & \textsc{mlp}  & \textsc{resnet-18} & \textsc{mlp}  & \textsc{mlp}  \\
    Batch normalization & No            & No            & Yes                & No            & No            \\
    Batch size          & 256           & 256           & 256                & 1024          & 256           \\
    Criterion           & \textsc{bce}  & \textsc{bce}  & \textsc{bce}       & \textsc{bce}  & \textsc{bce}  \\
    Dropout             & No            & No            & No                 & No            & No            \\
    Epochs              & 250           & 250           & 100                & 1000          & 1000          \\
    Learning rate       & 0.001         & 0.0001        & 0.001              & 0.00005       & 0.0001        \\
    Learning rate scheduling & No & No & No & Yes & Yes \\
    Optimizer           & \textsc{adam} & \textsc{adam} & \textsc{adam}      & \textsc{adam} & \textsc{adam} \\
    Weight decay        & 0.0           & 0.0           & 0.0                & 0.0           & 0.0           \\
    \bottomrule
  \end{tabular}
  \caption{Hyperparameters associated with the training procedure of our ratio estimator.}
  \label{table:hyperparameters}
\end{table}
\subsection{Tractable problem}
\label{sec:tractable_appendix}
\begin{figure}[H]
  \centering
  \includegraphics[width=\linewidth]{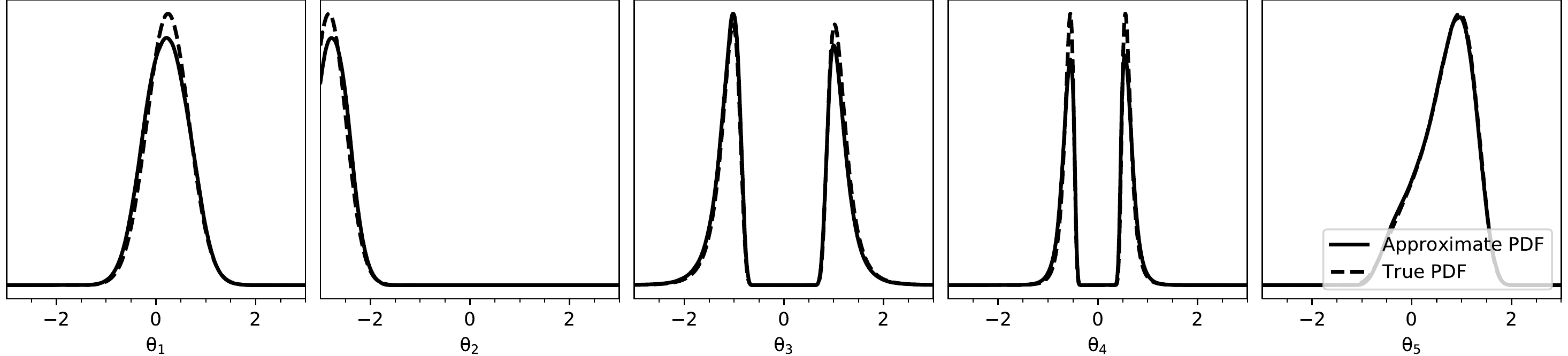}
  \caption{\textsc{pdf} of true posterior marginals $p(\btheta_i\vert\bx)$ and the corresponding approximations extracted from our ratio estimator. This can be computed for arbitrary model parameters $\btheta$ and observations $\bx$ by simply computing $p(\btheta)\hat{r}(\bx\vert\btheta)$.}
  \label{fig:e0_true_vs_ours_pdf}
\end{figure}
\subsubsection{Regularization and posterior approximation}
\label{sec:regularization}
\begin{figure}[h]
  \centering
  \begin{subfigure}{.75\linewidth}
    \centering
    \includegraphics[width=\linewidth]{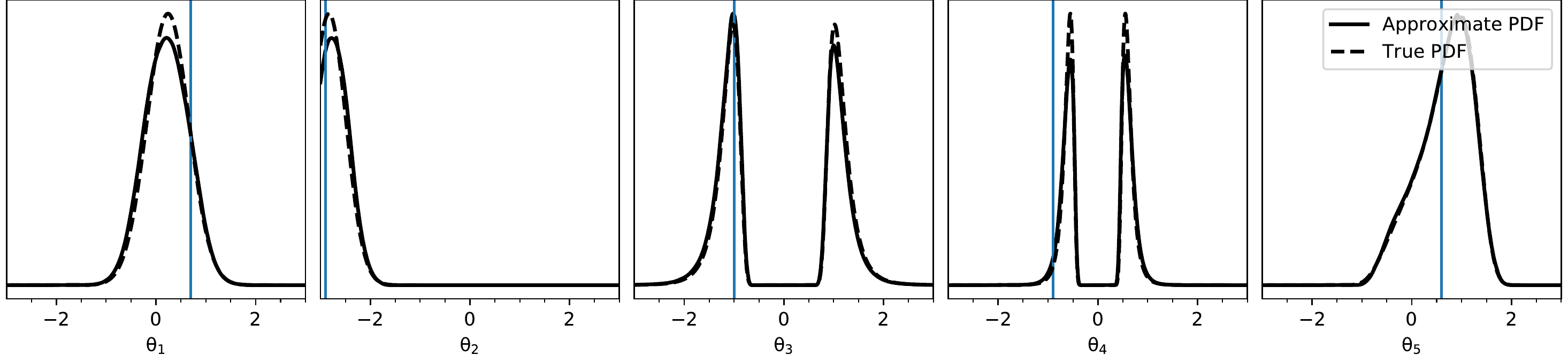}
    \caption{Weight decay = 0.0}
    \vspace{0.25cm}
  \end{subfigure}
  \begin{subfigure}{.75\linewidth}
    \centering
    \includegraphics[width=\linewidth]{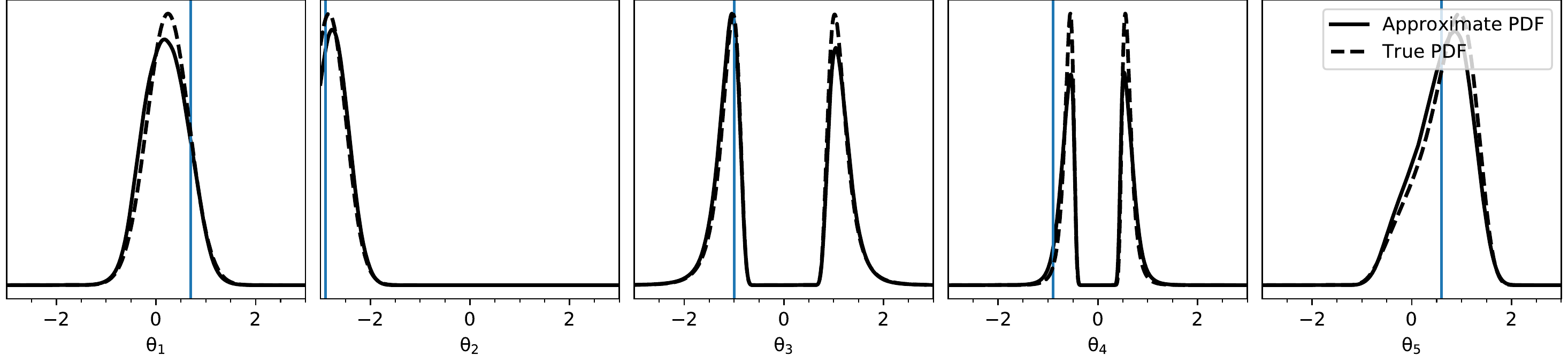}
    \caption{Weight decay = 0.00001}
    \vspace{0.25cm}
  \end{subfigure}
  \begin{subfigure}{.75\linewidth}
    \centering
    \includegraphics[width=\linewidth]{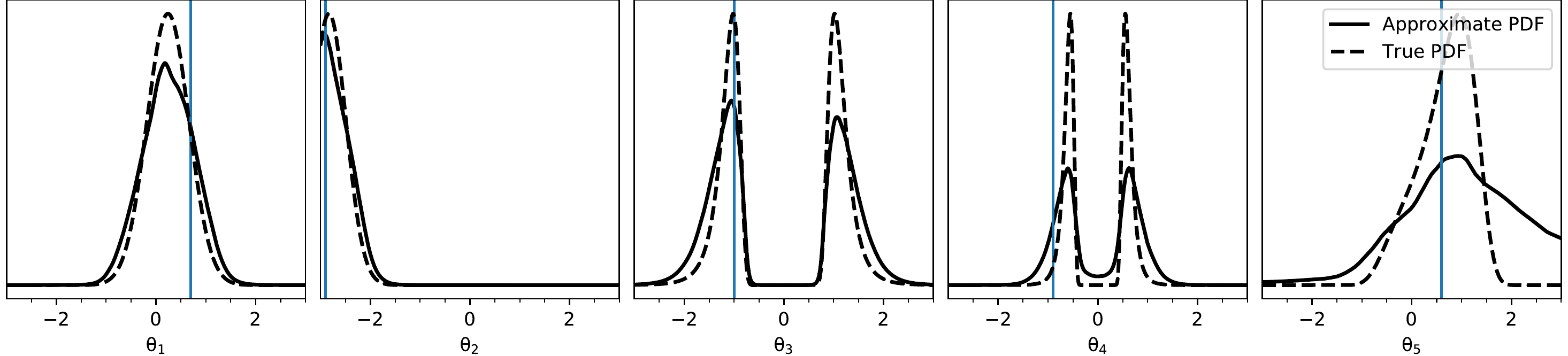}
    \caption{Weight decay = 0.01}
  \end{subfigure}
  \caption{The degree of regularization corresponds to a proportionally larger variance of the approximation compared to the truth. A slight bias in the training dataset might explain the consistently larger peak in the third posterior marginal.}
  \label{fig:regularization}
\end{figure}

\begin{figure}[H]
  \centering
  \begin{subfigure}{.32\linewidth}
    \centering
    \includegraphics[width=\linewidth]{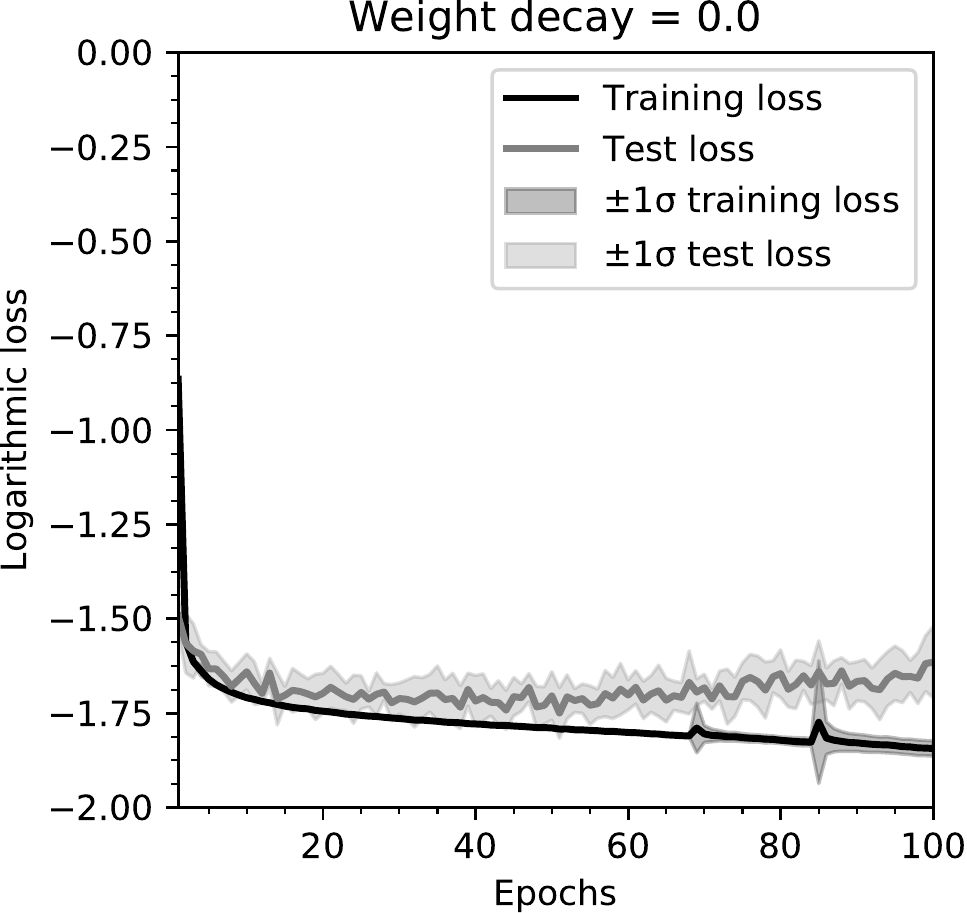}
    \caption{Weight decay = 0.0}
  \end{subfigure}
  \begin{subfigure}{.32\linewidth}
    \centering
    \includegraphics[width=\linewidth]{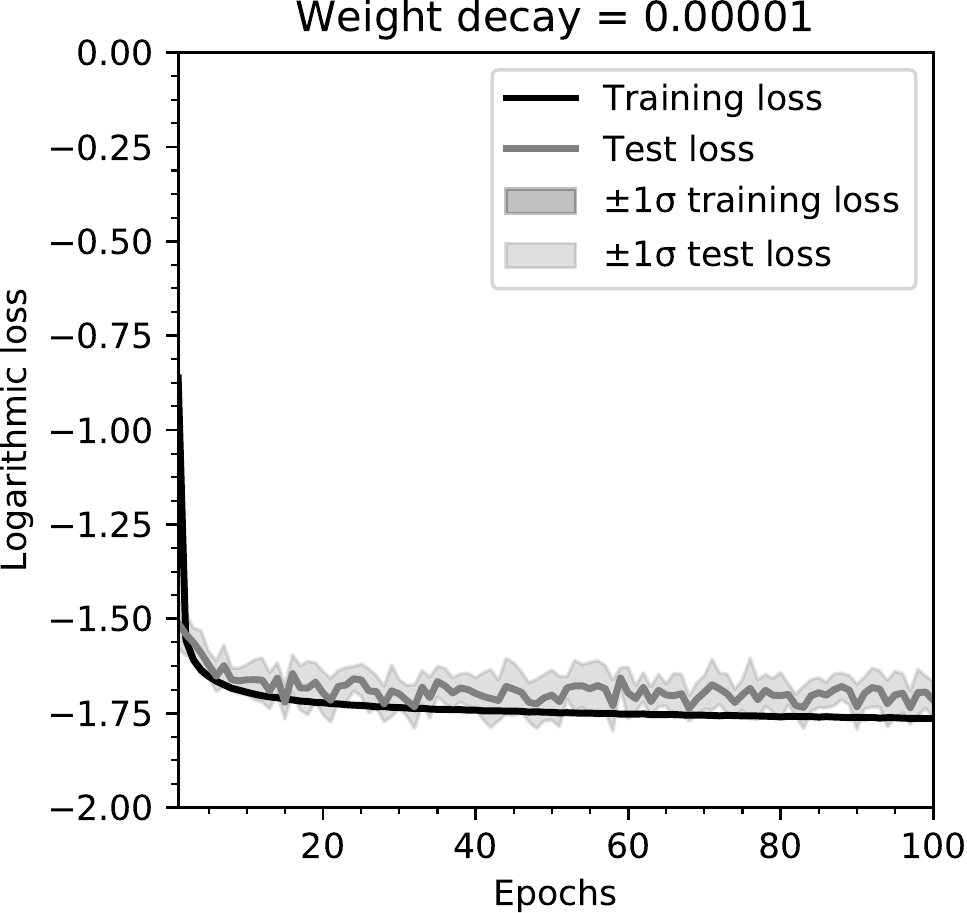}
    \caption{Weight decay = 0.00001}
  \end{subfigure}
  \begin{subfigure}{.32\linewidth}
    \centering
    \includegraphics[width=\linewidth]{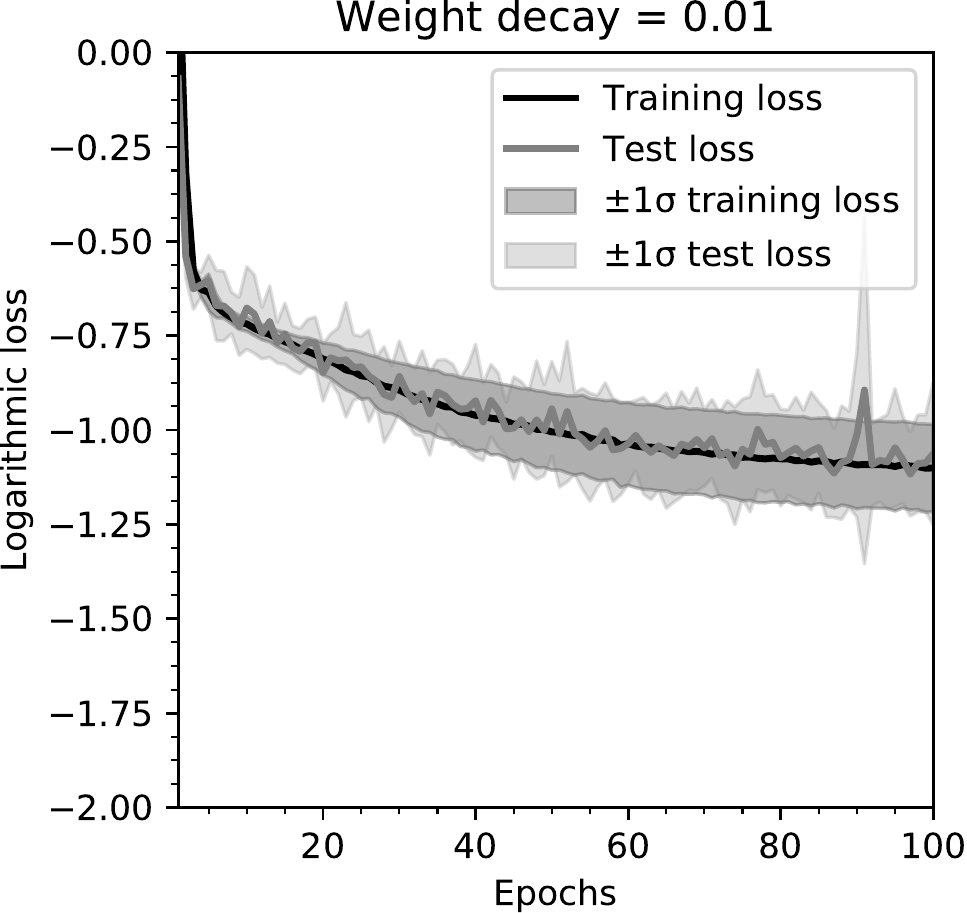}
    \caption{Weight decay = 0.01}
  \end{subfigure}
  \caption{Loss plots of the ratio estimators in Figure~\ref{fig:regularization}. We empirically find that a larger loss corresponds with a larger variance of the approximation with respect to the truth. Similar behavior is observed for ratio estimators with insufficient capacity, as studied in Appendix~\ref{sec:capacity}.}
  \label{fig:regularization_losses}
\end{figure}

This section studies the effects of regularization on the posterior approximation. We empirically find that the degree of regularization is proportional to an increase in variance of the approximation with respect to the true posterior. This translates into a proportionally larger (test) loss. Similar behavior can be observed in ratio estimators with insufficient capacity (Appendix~\ref{sec:capacity}). Figure~\ref{fig:regularization} demonstrates the effect of regularization on the approximate posterior with respect to the true posterior. The training and test losses are shown in Figure~\ref{fig:regularization_losses}.

\subsection{Detector calibration}
\label{sec:hep_results}
\begin{figure}[H]
  \centering
  \includegraphics[width=\linewidth]{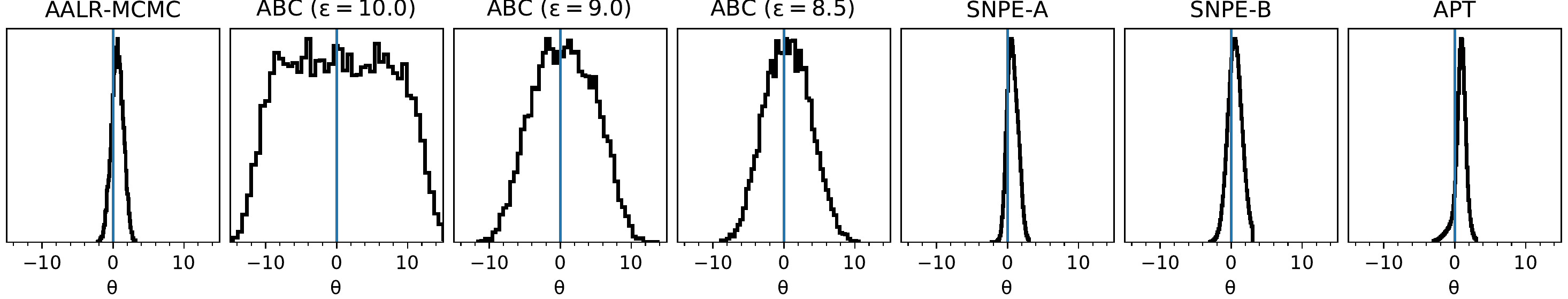}
  \caption{Approximate posteriors for the detector calibration benchmark. The posteriors are subsampled from several experimental runs.}
  \label{hep:other_posteriors}
\end{figure}
\begin{figure}[H]
  \centering
  \includegraphics[width=.65\linewidth]{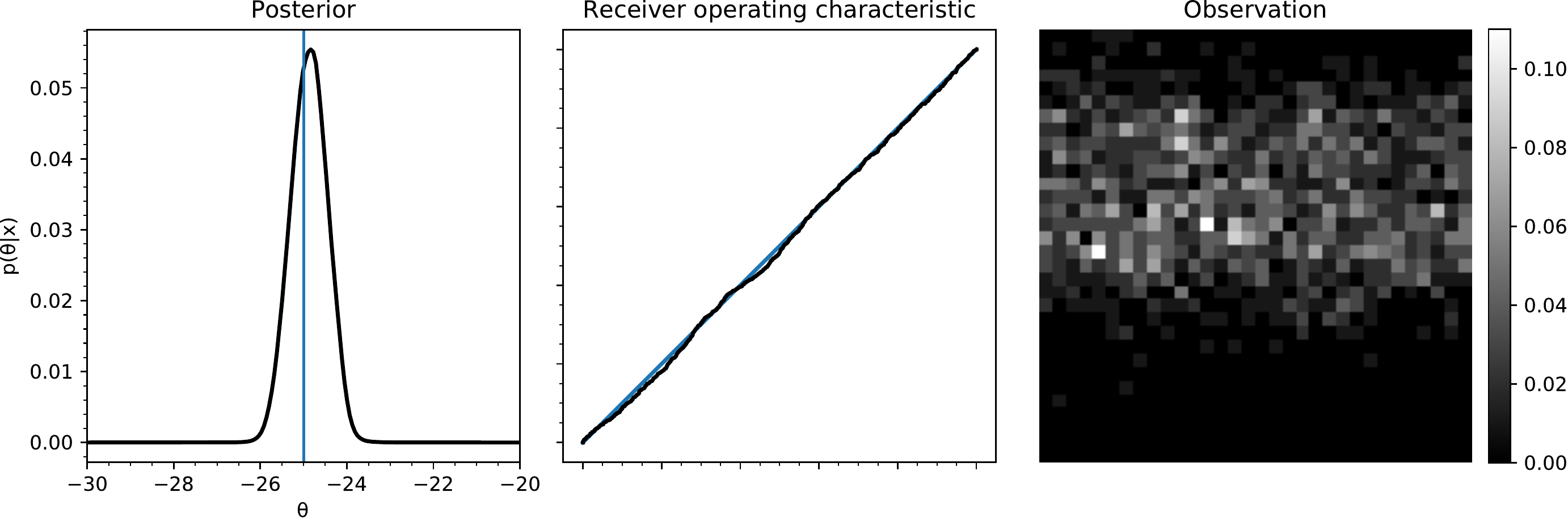}
  \includegraphics[width=.65\linewidth]{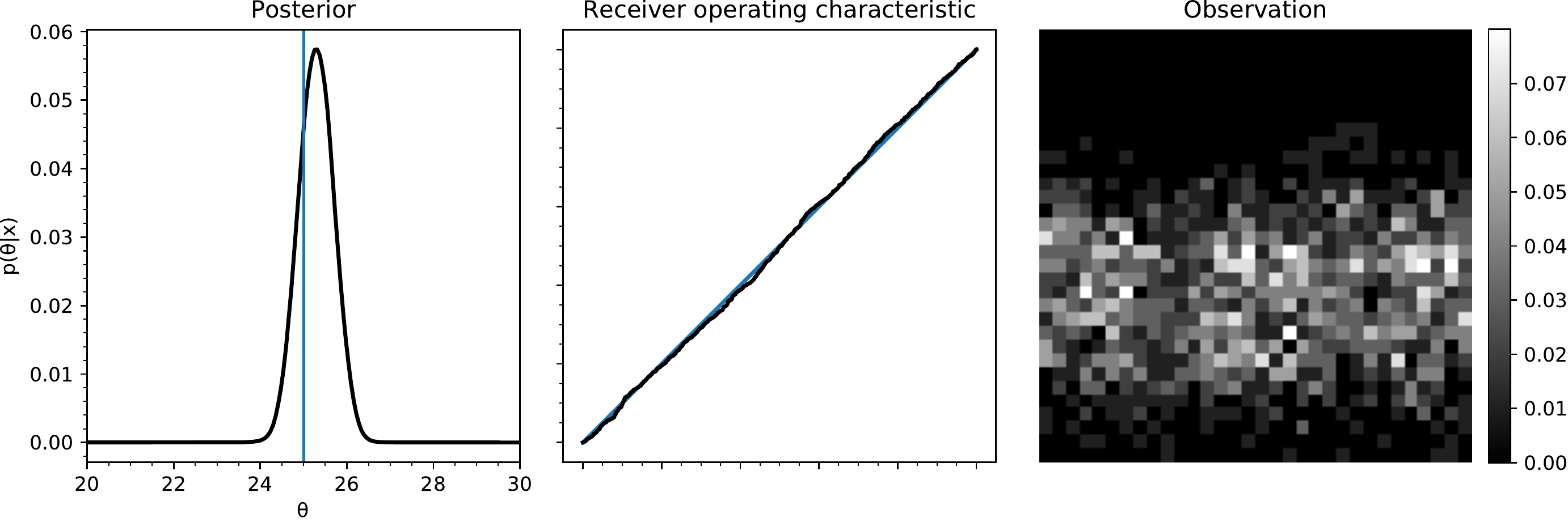}
  \caption{{\it (Left:)} Posteriors are obtained using the same ratio estimator. {\it (Middle):} Diagonal \textsc{roc} diagnostic, demonstrating the ability of the proposed method to model posteriors for arbitrary observations. {\it (Right):} Observations $\bx_o$.}
  \label{fig:pythia_amortized}
  \vspace{-0cm}
\end{figure}

\subsection{Lotka-Volterra}
\label{sec:lv_results}
\begin{figure}[H]
  \centering
  \begin{subfigure}{.24\linewidth}
     \centering
     \includegraphics[width=\linewidth]{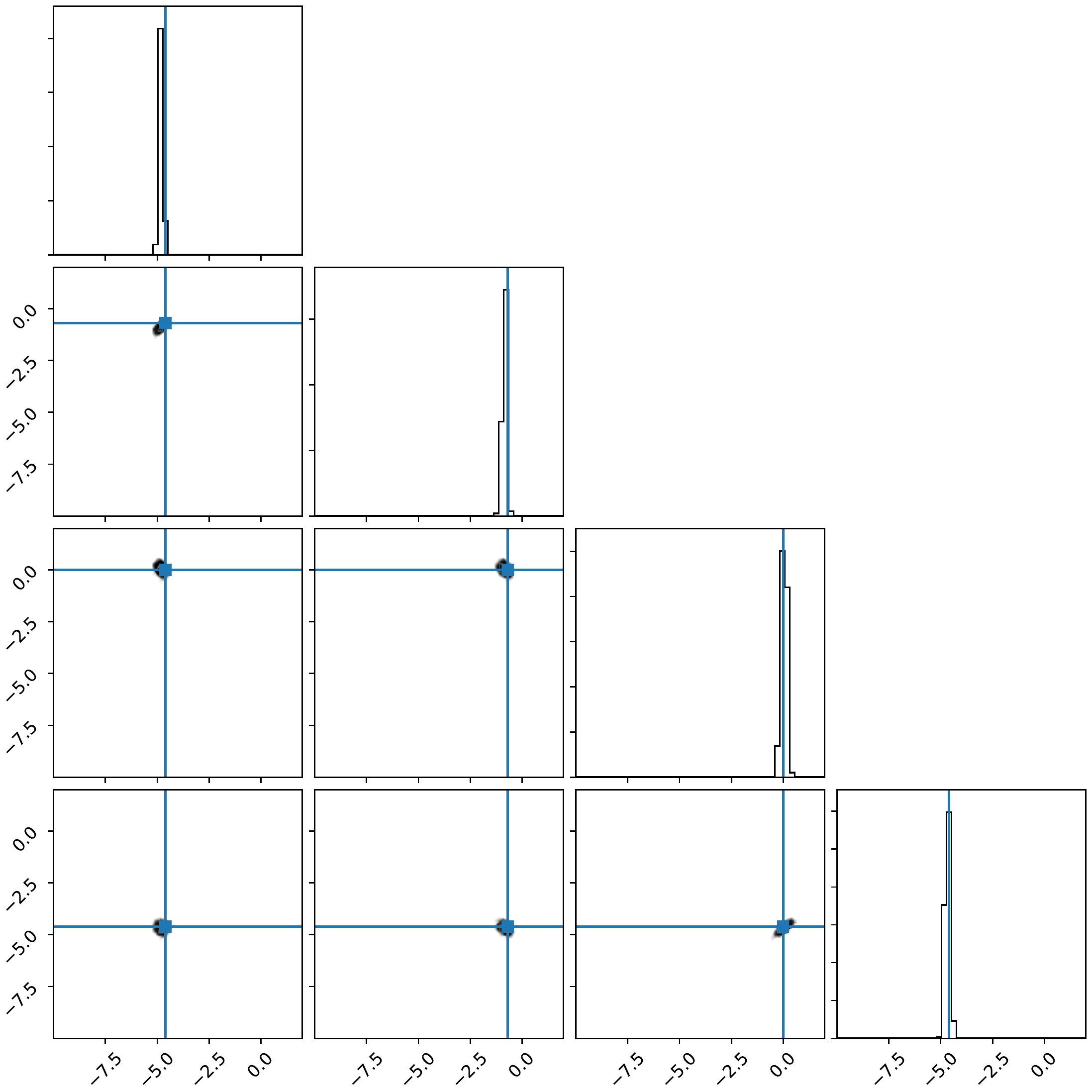}
     \caption{\textsc{aalr-mcmc}}
  \end{subfigure}
  \begin{subfigure}{.24\linewidth}
     \centering
     \includegraphics[width=\linewidth]{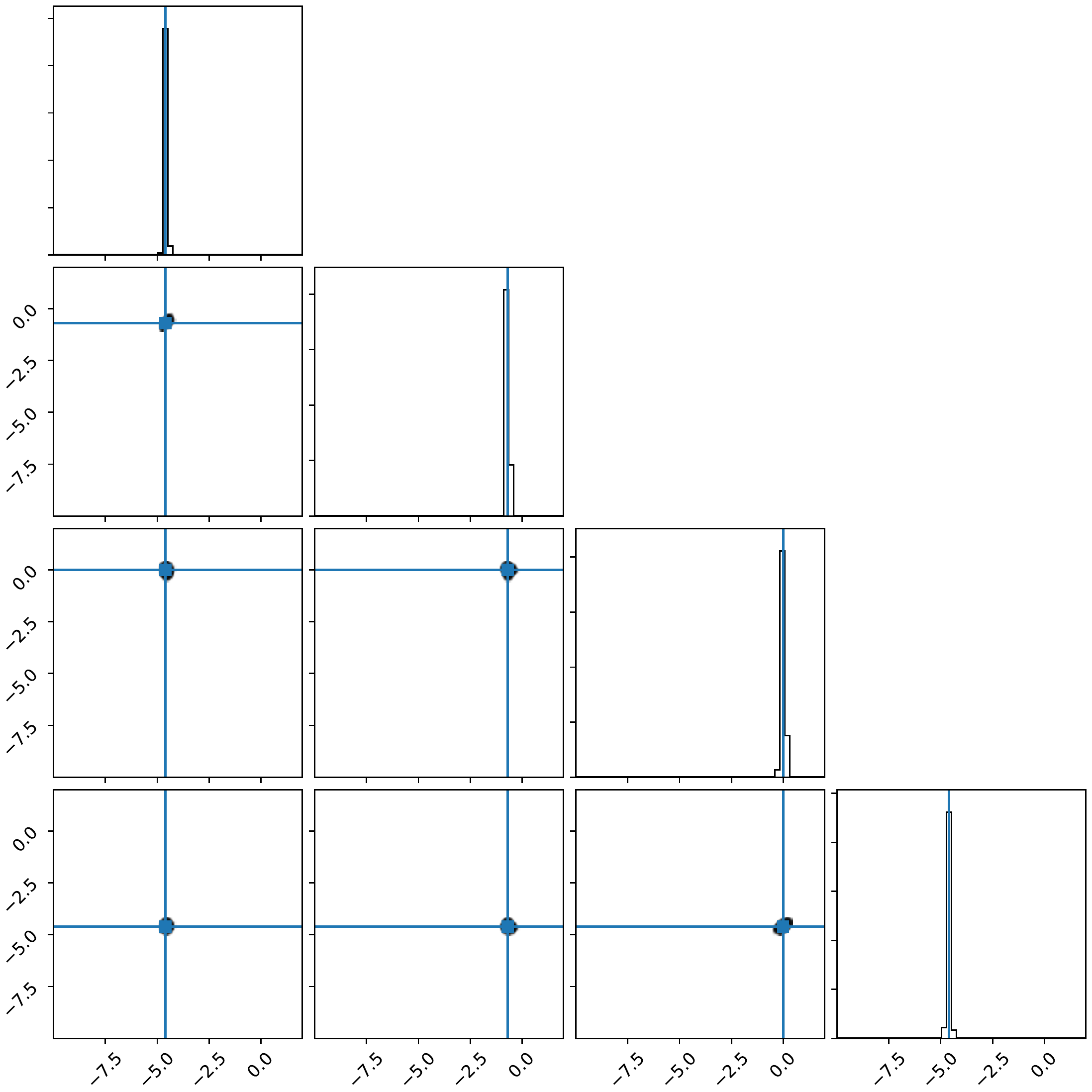}
     \caption{\textsc{snpe-a}}
  \end{subfigure}
  \begin{subfigure}{.24\linewidth}
     \centering
     \includegraphics[width=\linewidth]{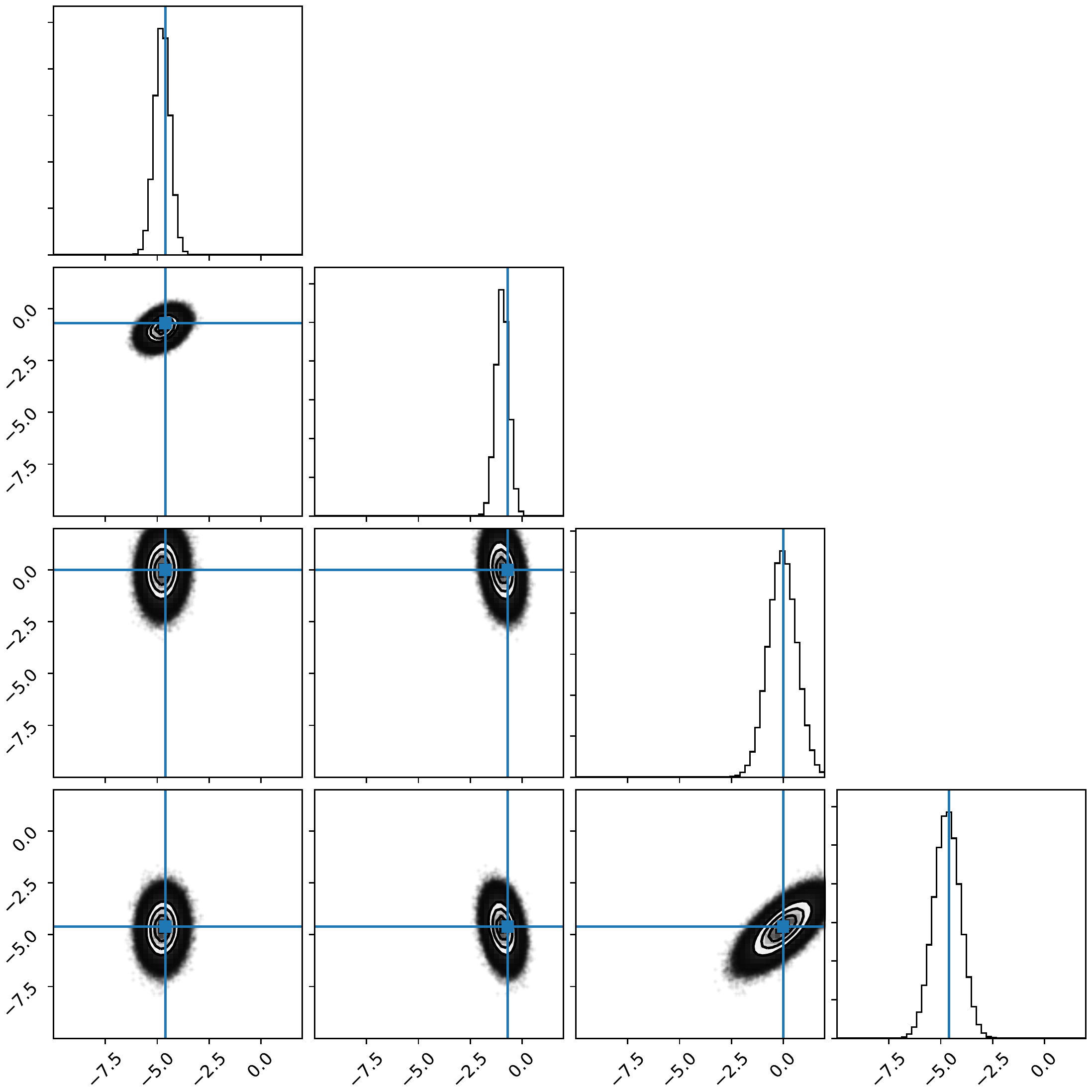}
     \caption{\textsc{snpe-b}}
  \end{subfigure}
  \begin{subfigure}{.24\linewidth}
     \centering
     \includegraphics[width=\linewidth]{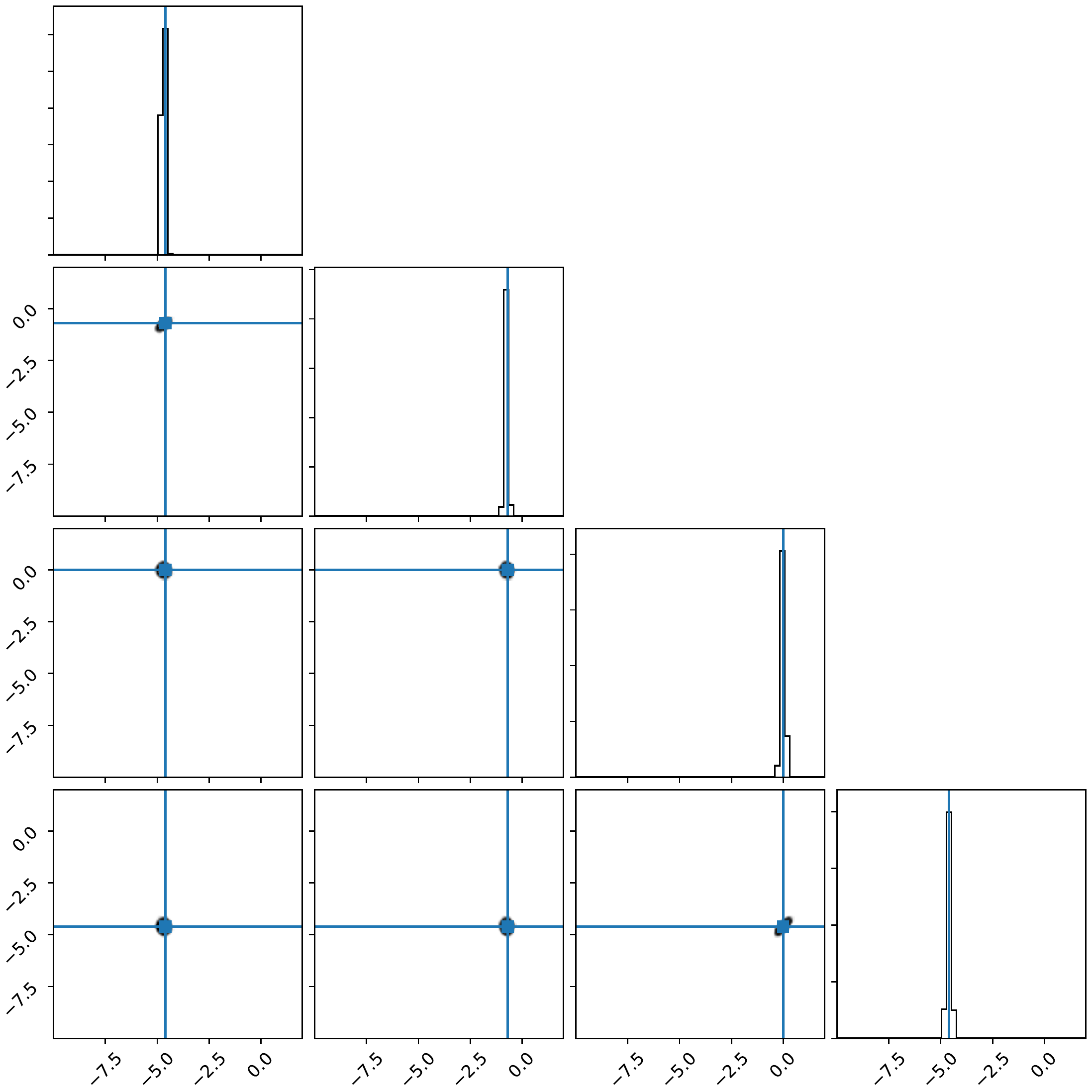}
     \caption{\textsc{apt}}
  \end{subfigure}
  \caption{Posterior approximations for the Lokta-Volterra problem. \textsc{aalr-mcmc}, \textsc{snpe-a} and \textsc{apt} are in agreement, while the \textsc{snpe-b} approximation is significantly broader.}
  \label{fig:lv_posteriors}
\end{figure}
\begin{figure}[H]
  \centering
  \includegraphics[width=.75\linewidth]{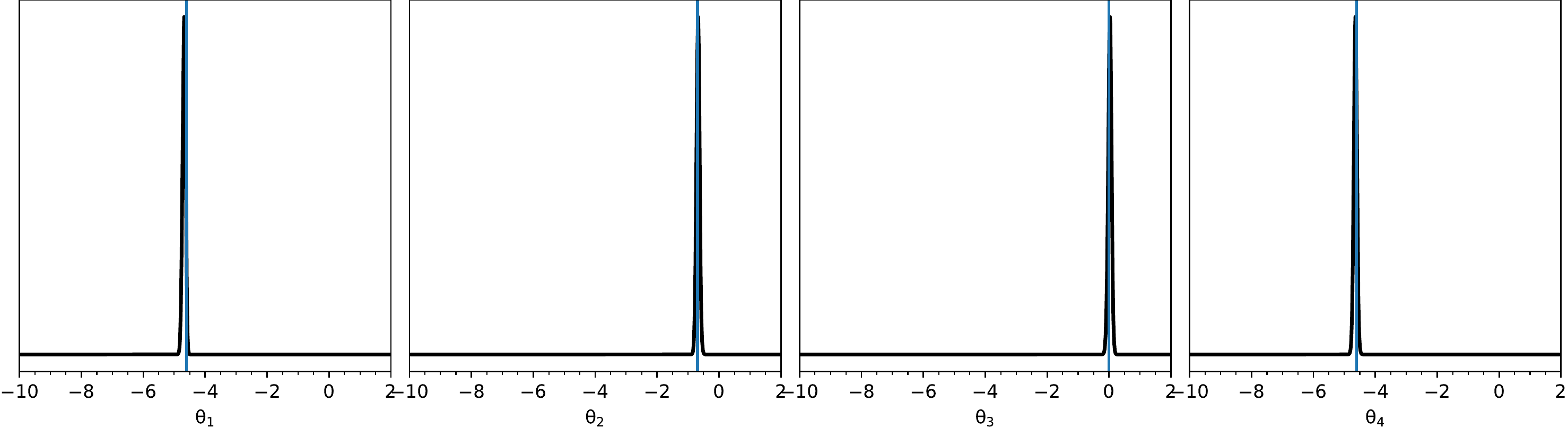}
  \caption{Posterior marginals of our approximation for the Lotka-Volterra problem.}
  \label{fig:lv_marginals}
\end{figure}

\subsection{\textsc{m/g/1} queuing model}
\label{sec:mg1}

\begin{figure}[H]
  \centering
  \begin{subfigure}{.24\linewidth}
     \centering
     \includegraphics[width=\linewidth]{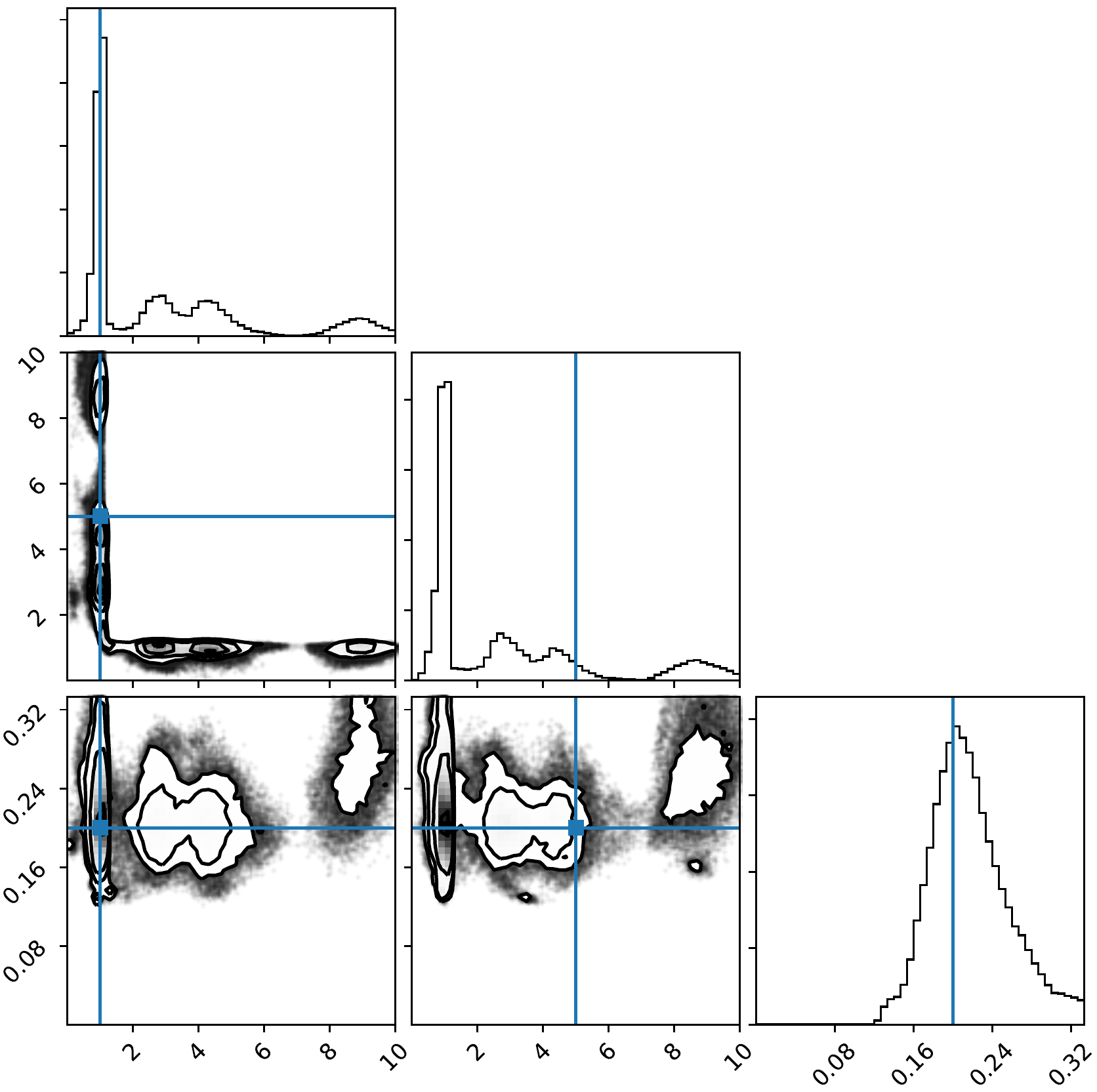}
     \caption{\textsc{aalr-mcmc}}
  \end{subfigure}
  \begin{subfigure}{.24\linewidth}
     \centering
     \includegraphics[width=\linewidth]{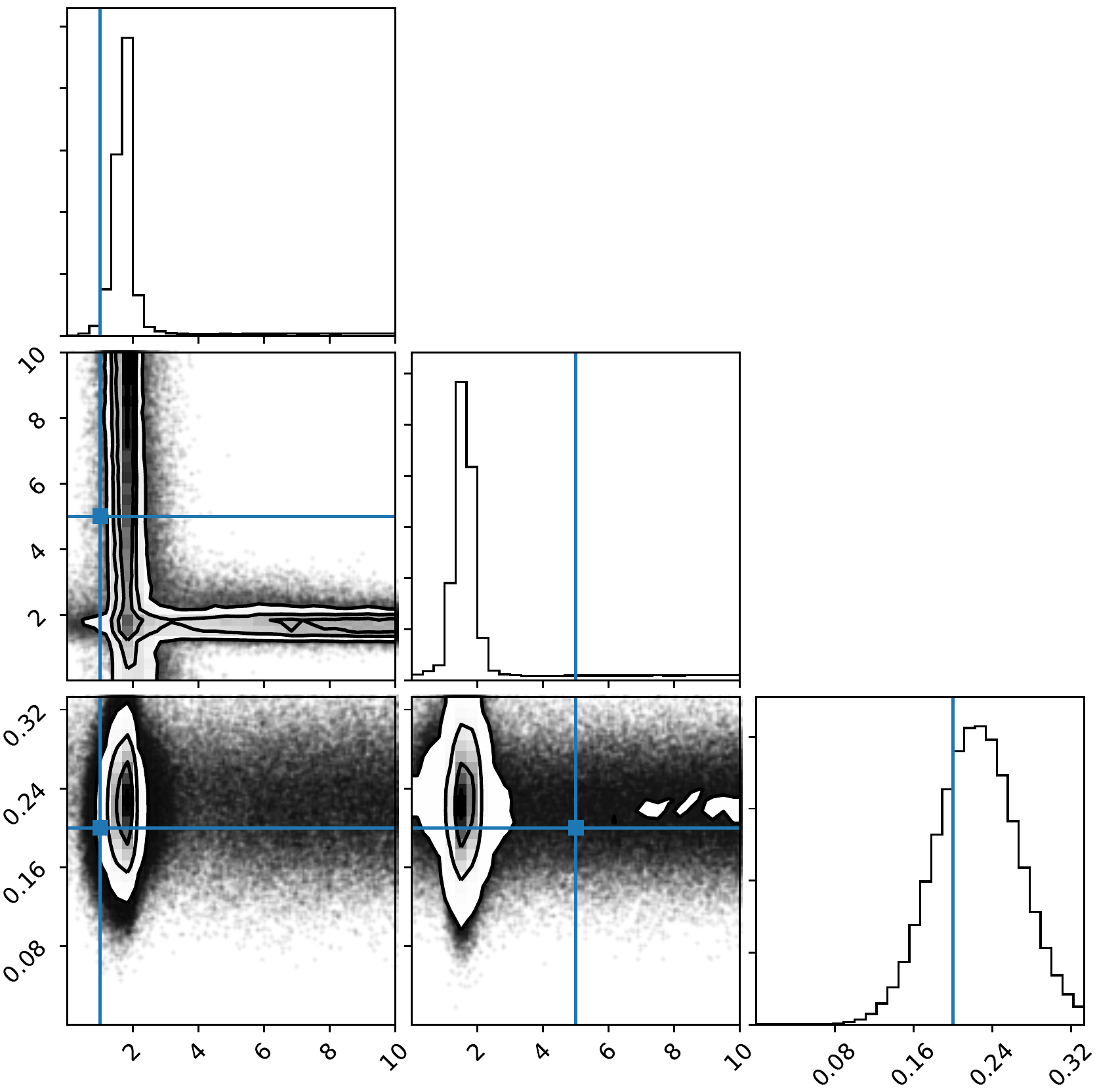}
     \caption{\textsc{snpe-a}}
  \end{subfigure}
  \begin{subfigure}{.24\linewidth}
     \centering
     \includegraphics[width=\linewidth]{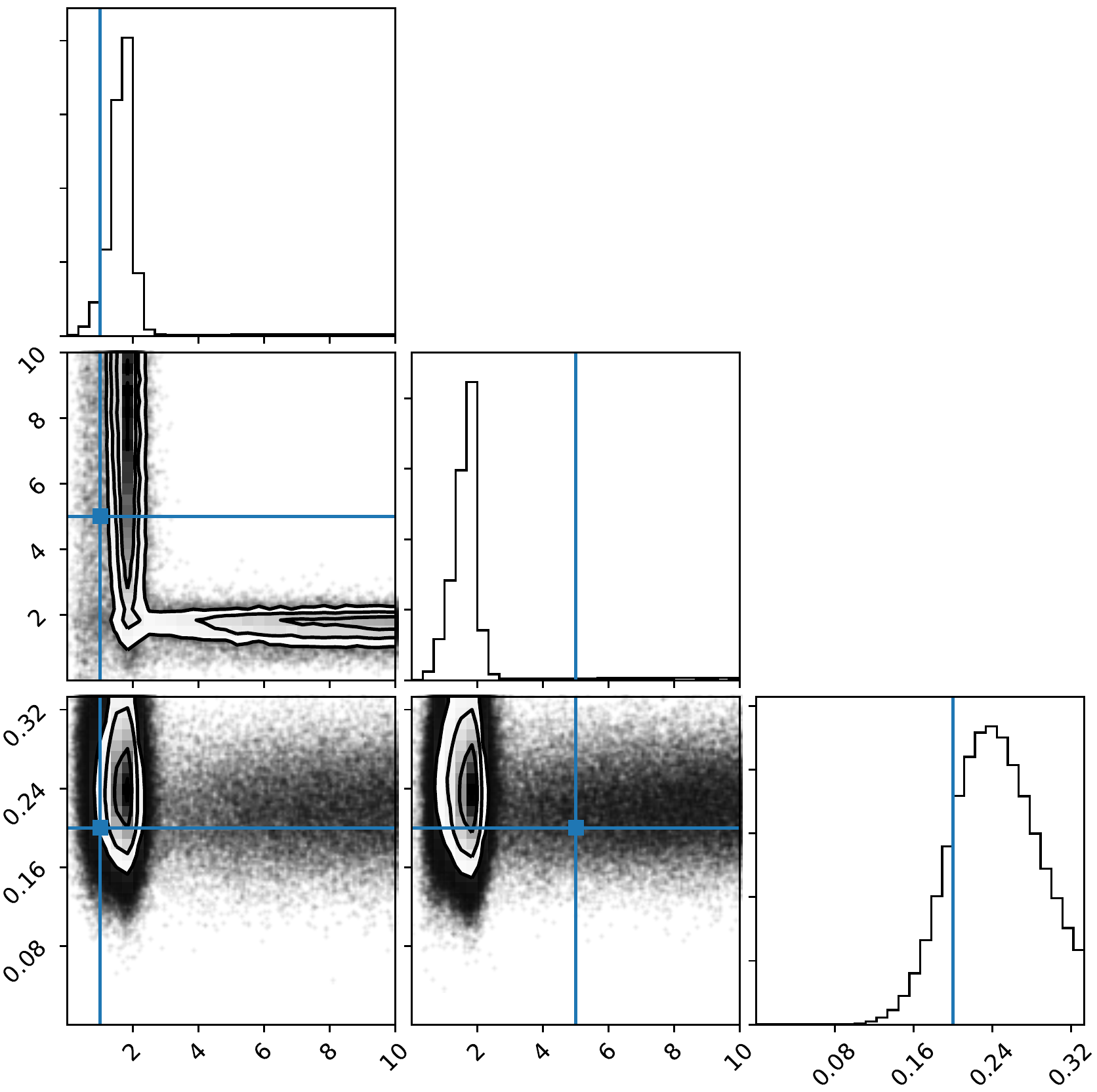}
     \caption{\textsc{snpe-b}}
  \end{subfigure}
  \begin{subfigure}{.24\linewidth}
     \centering
     \includegraphics[width=\linewidth]{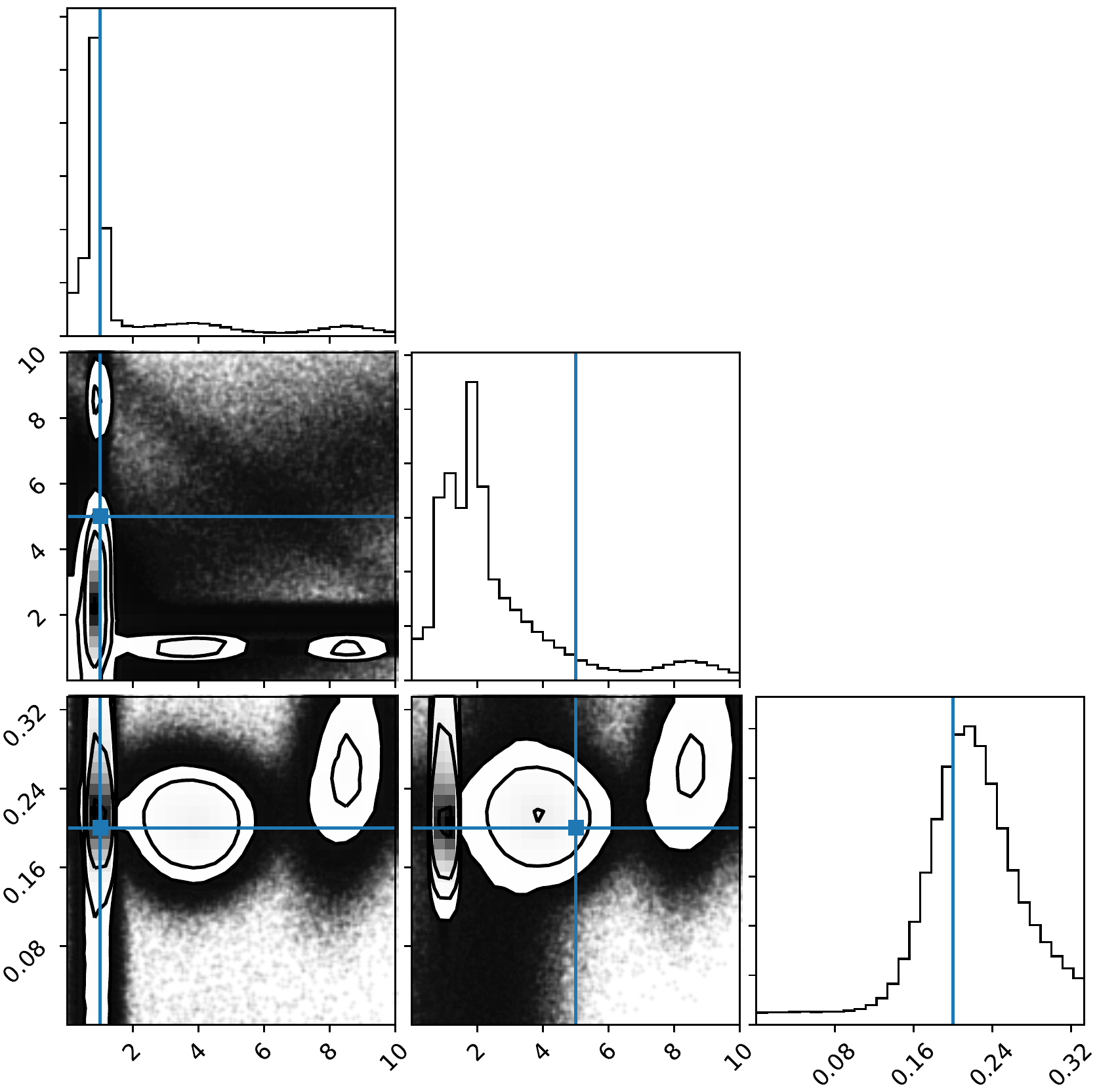}
     \caption{\textsc{apt}}
  \end{subfigure}
  \caption{Posteriors from the \textsc{m/g/1} benchmark. The experiments are repeated 10 times and the approximate posteriors are subsampled from those runs. Despite the high variance of \textsc{apt}, it shares the most structure with \textsc{aalr-mcmc}.}
  \label{fig:mg1_posteriors}
\end{figure}

\subsection{Scientific use case: strong gravitational lensing}
\label{sec:lensing_appendix}
The simulation model consists of 4 main components. The first involves the telescope optics. We model the \textsc{psf} (point spread function) as a Gaussian with standard deviation 0.5 in a $3 \times 3$ pixel kernel. The \textsc{ccd} sensor is set to an exposure time of 1000 seconds, background sky level = 0.1 and \textsc{ccd} noise is added. The mass distribution of the foreground galaxy is modeled as an elliptical isothermal~\cite{ellipticalisothermal} at redshift $z = 0.5$ with axis ratio = 0.99, a random orientation-angle and an Einstein radius sampled from the prior. We do not model galaxy foreground light for the marginalization problem. For the Bayesian model selection problem, we model the foreground light of the lensing galaxy as an elliptical sersic with a random orientation angle and a sersic index sampled from $\mathcal{U}(.5, 1.5)$. For every source galaxy, we only model the light profile and their relative positions with respect to the lens. Source galaxies have an assumed redshift of $z = 2$. We assume the Plack15 cosmology. Table~\ref{table:source_galaxy_parameters} describes the parameters and respective distributions we sampled from to generate a light profile for a single source galaxy.
\begin{table}[H]
  \centering
  \begin{tabular}{ll}
    \toprule
    Parameter & Distribution \\
    \midrule
    Axis ratio & $\mathcal{U}(0.1,0.9)$  \\
    Effective radius & $\mathcal{U}(0.1,0.4)$ \\
    Intensity (flux) & $\mathcal{U}(0.1,0.5)$ \\
    Location $x$ & $\mathcal{U}(-1.0,1.0)$ \\
    Location $y$ & $\mathcal{U}(-1.0,1.0)$ \\
    Axis orientation & $\mathcal{U}(0, 360)$ \\
    Sersic index & $\mathcal{U}(0.5, 3.0)$ \\
    \bottomrule
  \end{tabular}
  \caption{A complete description of the parameters describing the light profile is described in the \texttt{autolens} documentation.}
  \label{table:source_galaxy_parameters}
\end{table}


\newpage
\section{Capacity of the ratio estimator and its effect on the estimated posterior}
\label{sec:capacity}

\begin{figure}[H]
  \centering
  \hspace{-0.6cm}
  \includegraphics[width=\linewidth]{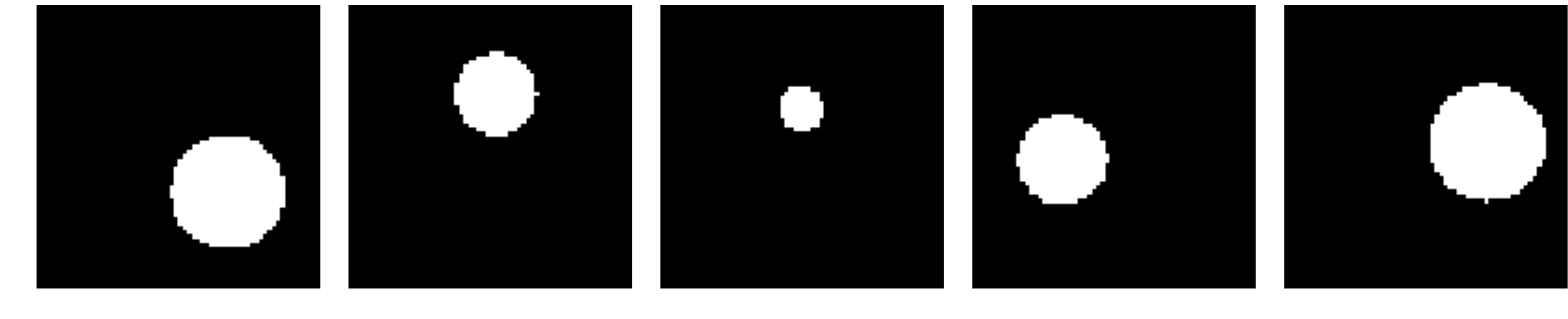}
  \caption{Random subsamples from the marginal model $p(\bx)$ for the circle problem.}
  \label{fig:circle_samples}
\end{figure}

\begin{figure}[h!]
  \centering
  \includegraphics[width=.7\linewidth]{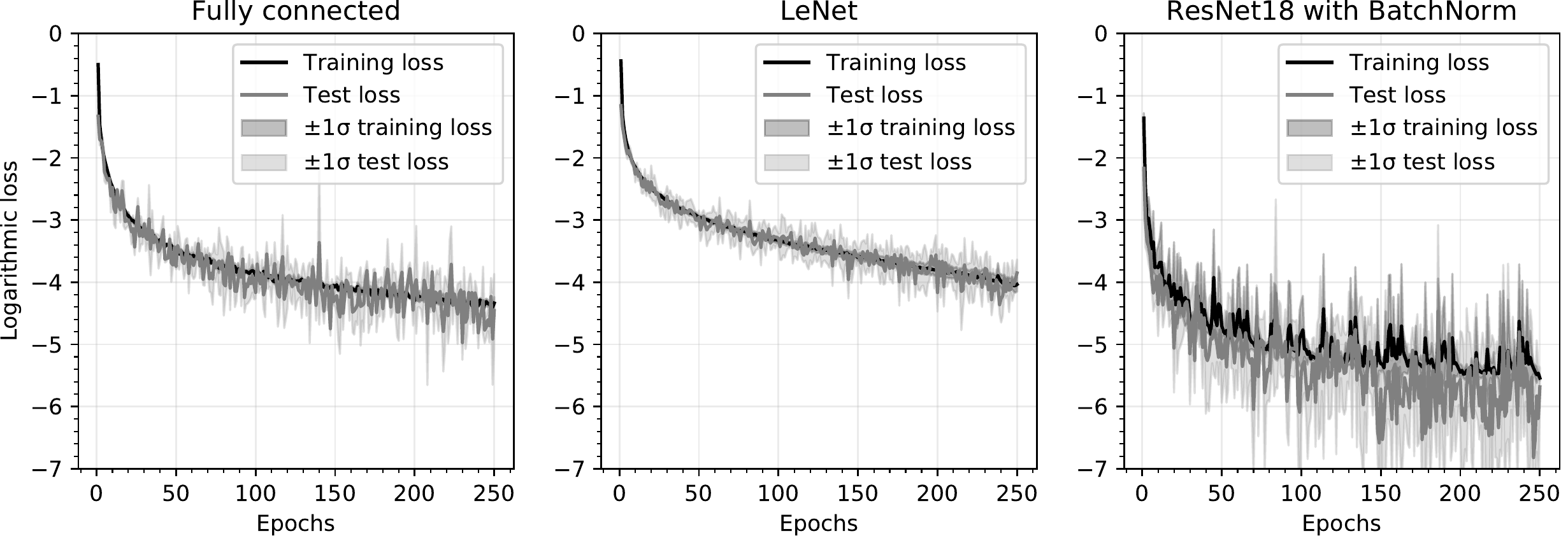}
  \caption{Loss plots of ratio estimators with different architectures. Mean training and test loss is reported, including the respective standard deviations. The plots indicate that the \textsc{resnet-18} ratio estimator should preform best. {\it (Left)}: The fully connected architecture. {\it (Middle)}: The \textsc{lenet} architecture. {\it (Right)}: The \textsc{resnet-18} architecture. }
  \label{fig:circle_losses}
\end{figure}

We investigate the approximation error in relation to the capacity of a ratio estimator. We consider a simple simulation model which accepts a model parameter $\btheta\triangleq(x,y,r)$ and produces an image with a resolution of $64\times 64$ containing a circle at position $x,y$ with radius $r$. Given the deterministic nature of the simulation model, we expect the posterior to be tight surrounding the generating parameters. The radius parameter $r$ is multimodal due to squaring operation, which implies there should be a peak at $-r$ as well. We consider a uniform prior in the range $[-1,1]$ for the parameters $x$ and $y$. The radius has a uniform prior between $[-.5,.5]$. Random samples from the simulation model under $p(\btheta)$ are shown in Figure~\ref{fig:circle_samples}.
We evaluate the following architectures: \begin{enumerate*}[label=(\roman*)]
  \item a fully connected architecture with 3 hidden layers and 128 units each (assumed low capacity),
  \item \textsc{lenet}~\cite{lenet} (assumed mid-range capacity),
  \item and \textsc{resnet-18}~\cite{resnet} (assumed high-capacity).\end{enumerate*}
All models are trained according to the procedure described in Appendix~\ref{sec:recommended}. We train the ratio estimators using a batch-size of 256 samples and the \textsc{adam}~\cite{adam} optimizer. As in other experiments, the neural networks use the \textsc{selu}~\cite{selu} activation function. The networks are trained for 250 epochs. No regularization or data normalization techniques are applied, with the exception of batch normalization~\cite{batchnorm} for the \textsc{resnet-18} architecture. For every architecture, we train 5 models. Figure~\ref{fig:circle_losses} shows the mean loss curves and their standard deviations.
We did not explore other hyperparameters or invest in additional training iterations. The loss plots indicate that the ratio estimator based on \textsc{resnet-18} should perform best.


\begin{figure}[h!]
  \centering
  \begin{subfigure}{\linewidth}
    \centering
    \includegraphics[width=.49\linewidth]{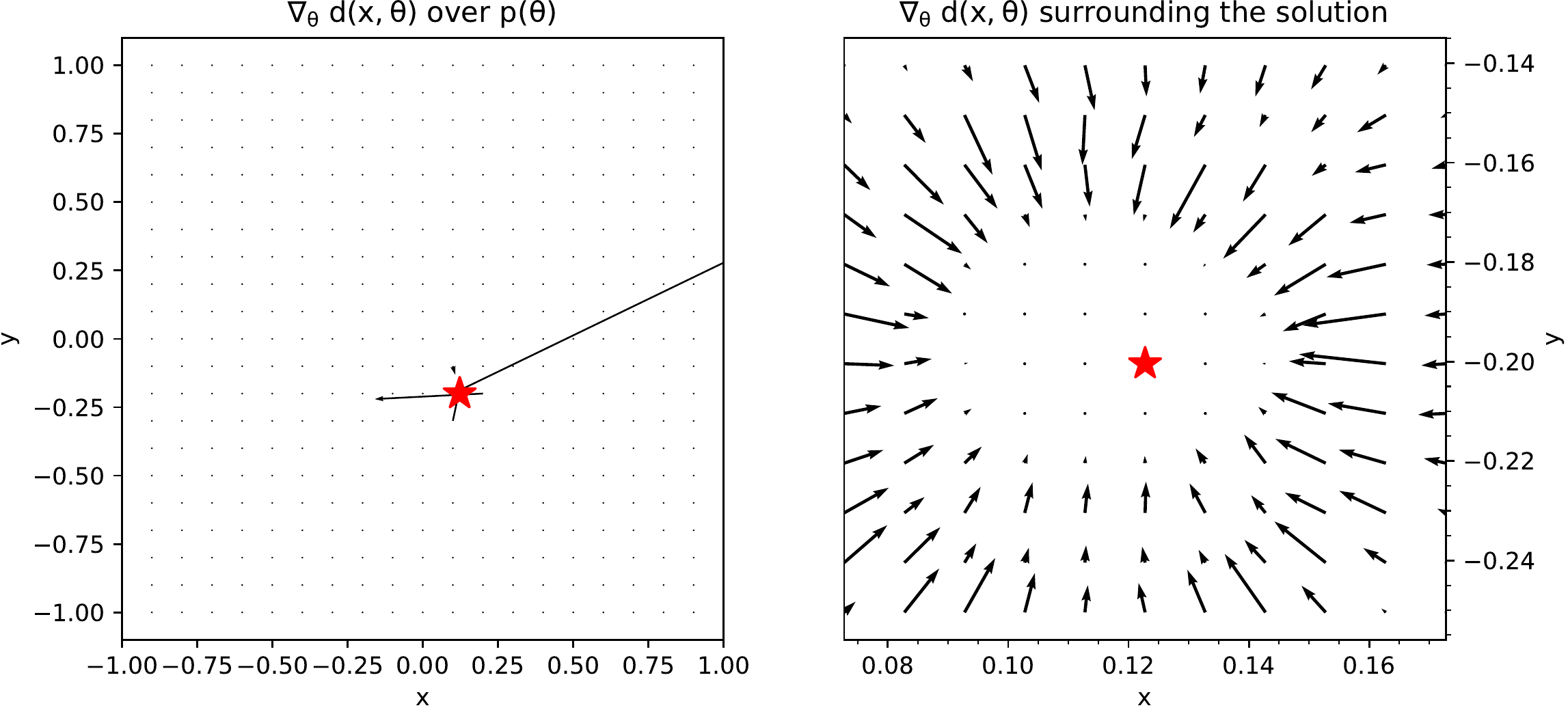}
    \includegraphics[width=.49\linewidth]{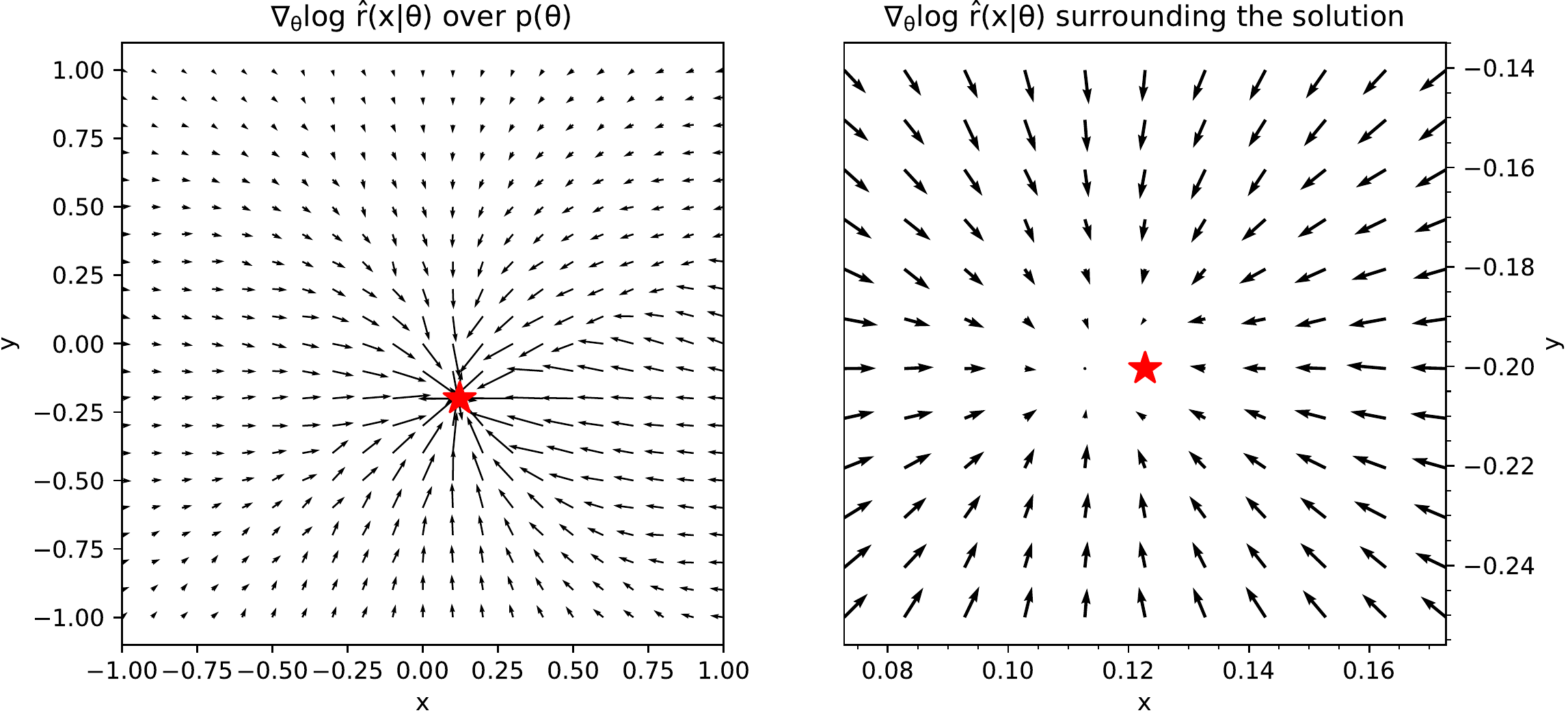}
    \caption{Fully connected ratio estimator}
    \vspace{.2cm}
  \end{subfigure}
  \begin{subfigure}{\linewidth}
    \centering
    \includegraphics[width=.49\linewidth]{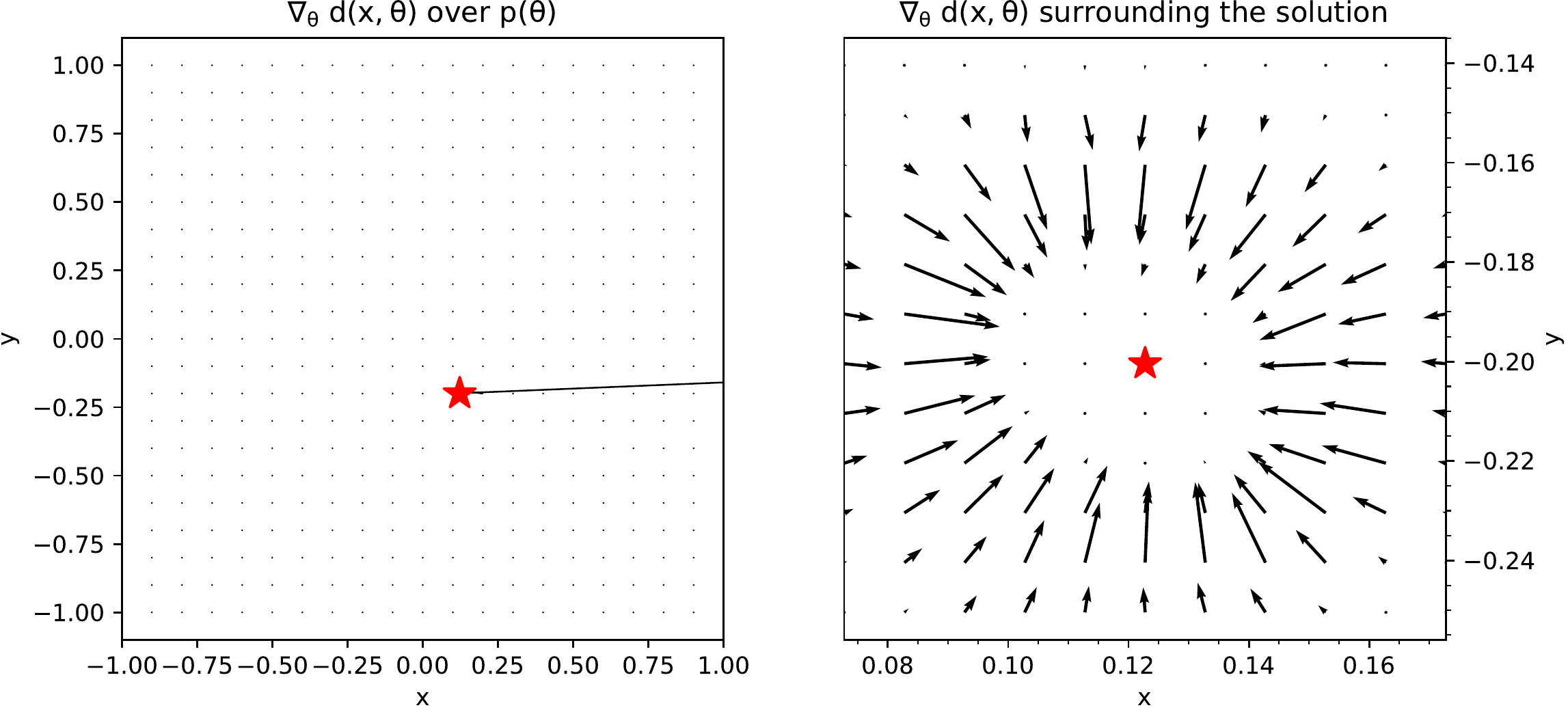}
    \includegraphics[width=.49\linewidth]{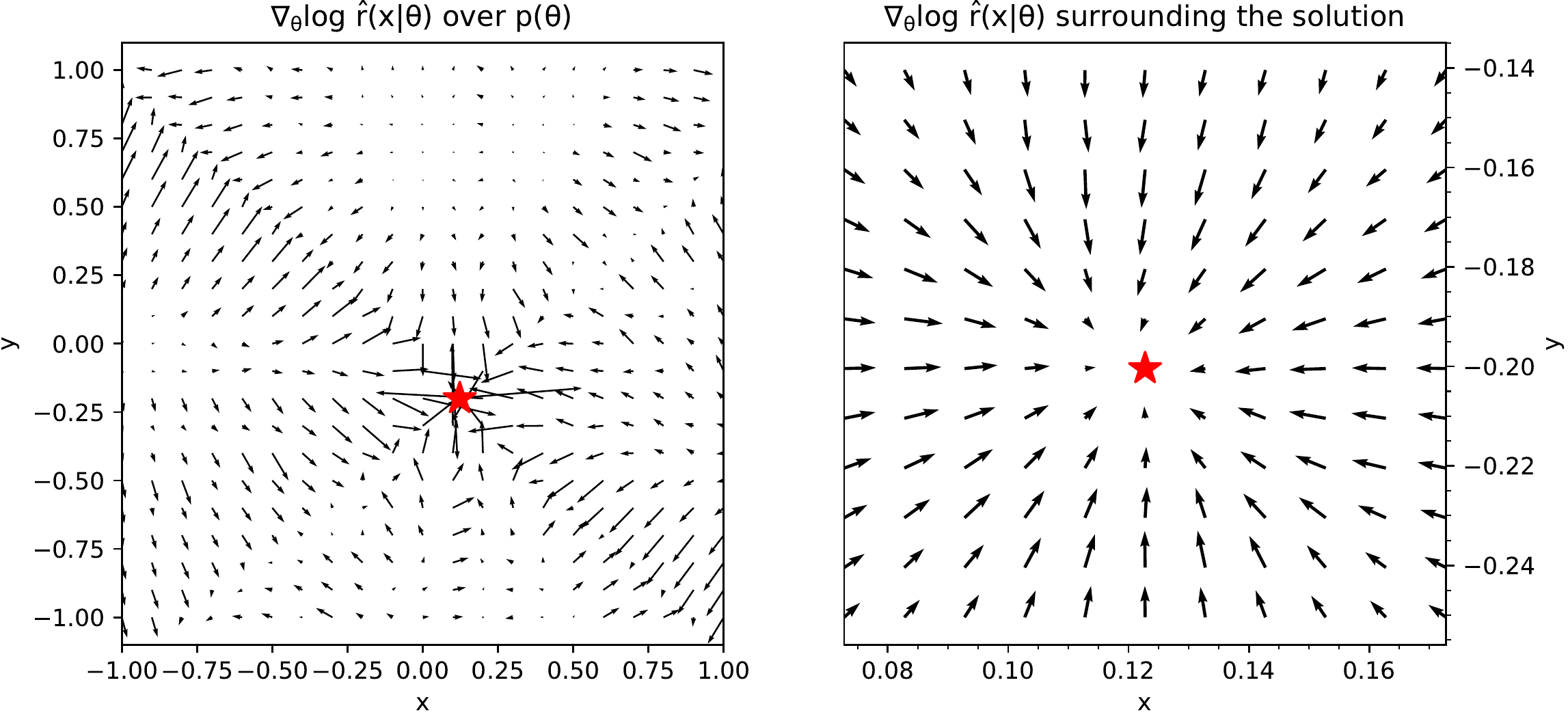}
    \caption{\textsc{lenet} ratio estimator}
    \vspace{.2cm}
  \end{subfigure}
  \begin{subfigure}{\linewidth}
    \centering
    \includegraphics[width=.49\linewidth]{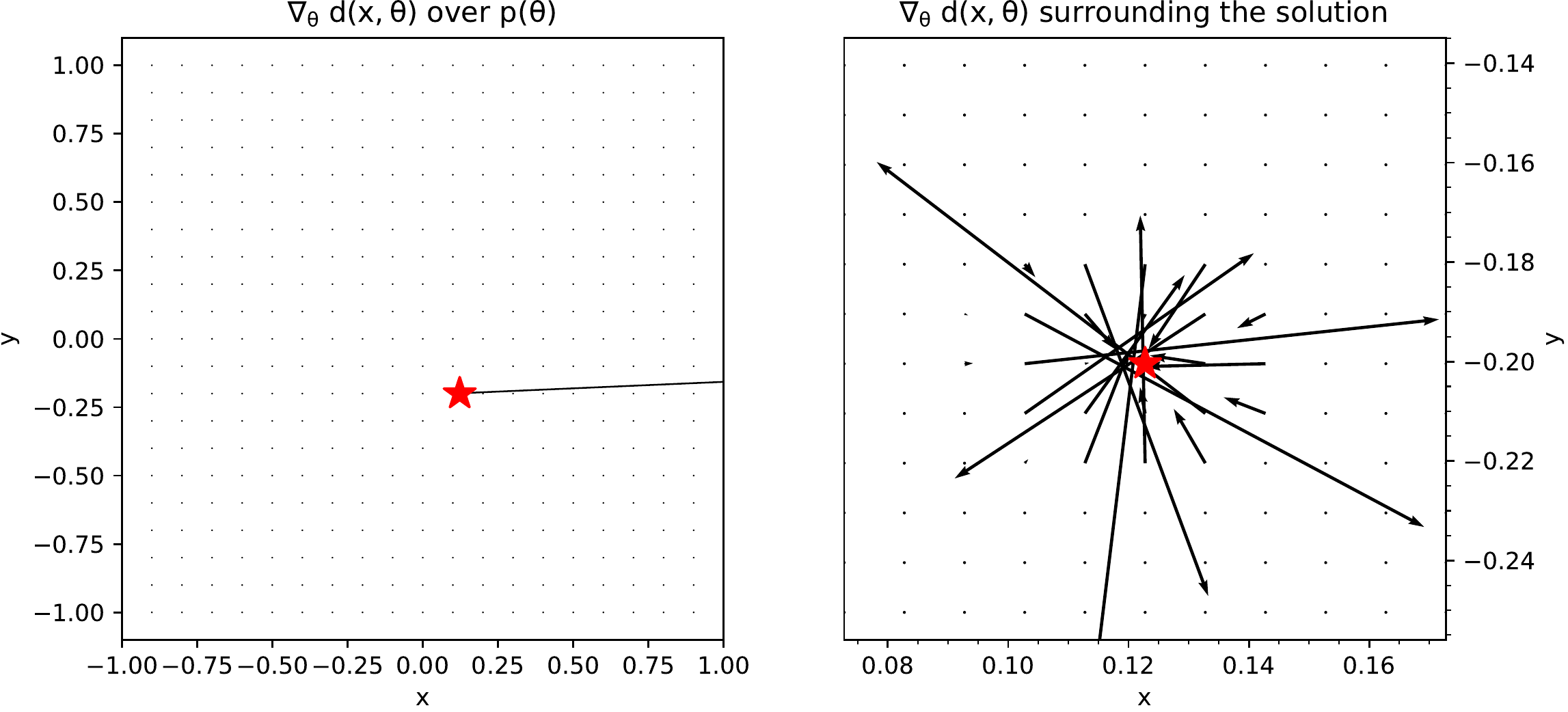}
    \includegraphics[width=.49\linewidth]{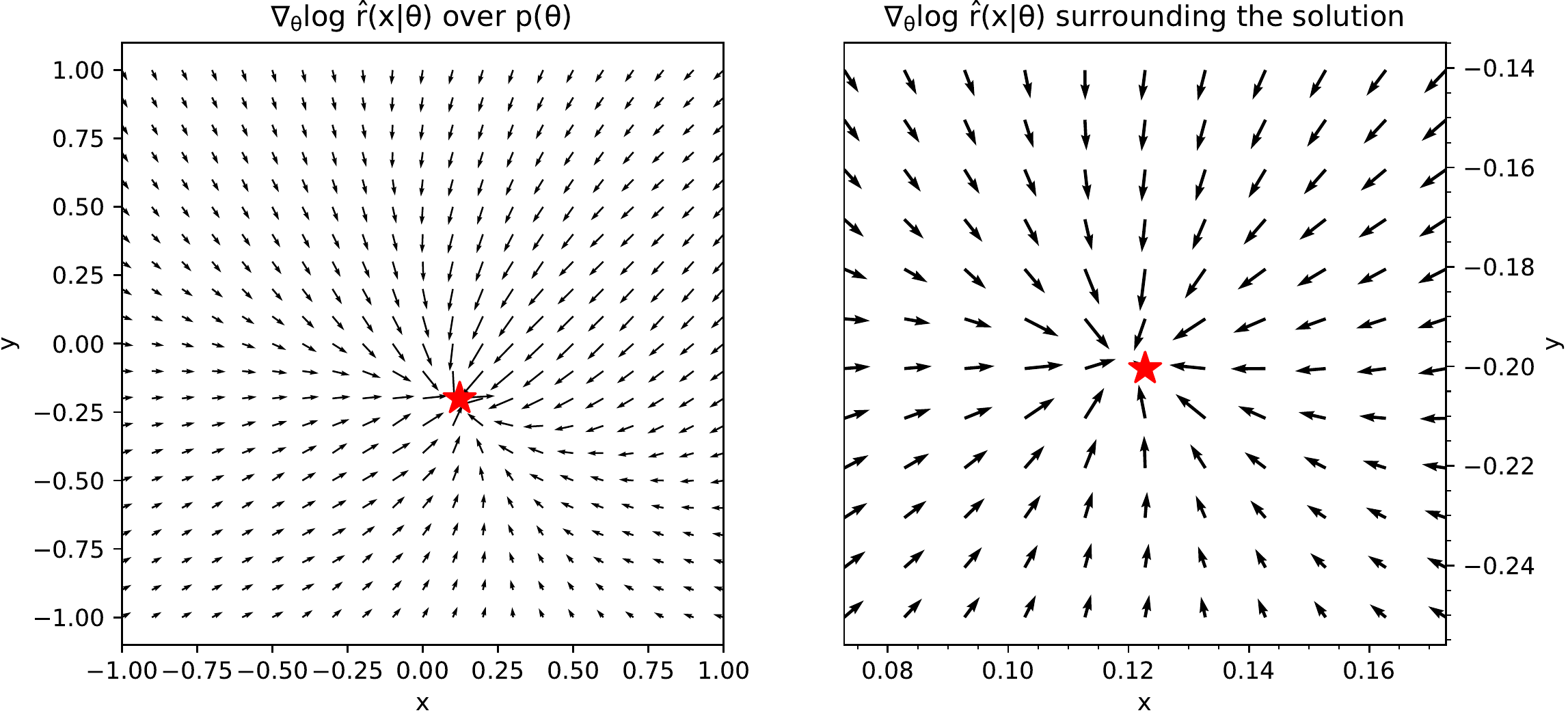}
    \caption{\textsc{resnet-18} ratio estimator}
  \end{subfigure}
  \caption{Vector (gradient) fields for the $x$ and $y$ parameters. The red star indicates the true solution. The parameter $r$ remains fixed during the computation of these fields. The left-hand side shows the fields when backpropagating through the classifier output $\bs(\bx,\btheta)$. The effect of the sigmoidal operation is clear, as the gradient saturates when the classifier $\bs(\bx,\btheta)$ is (almost) able to perfectly discriminate samples. This supports the innovations presented in Section~\ref{sec:improve_approx_lr}, as $\nabla_{\btheta}~\log\hat{r}(\bx\vert\btheta)$ does not show this behavior.}
  \label{fig:circle_field_maps}
\end{figure}

\begin{figure}
  \centering
  \begin{subfigure}{.8\linewidth}
    \centering
    \includegraphics[height=.3\linewidth]{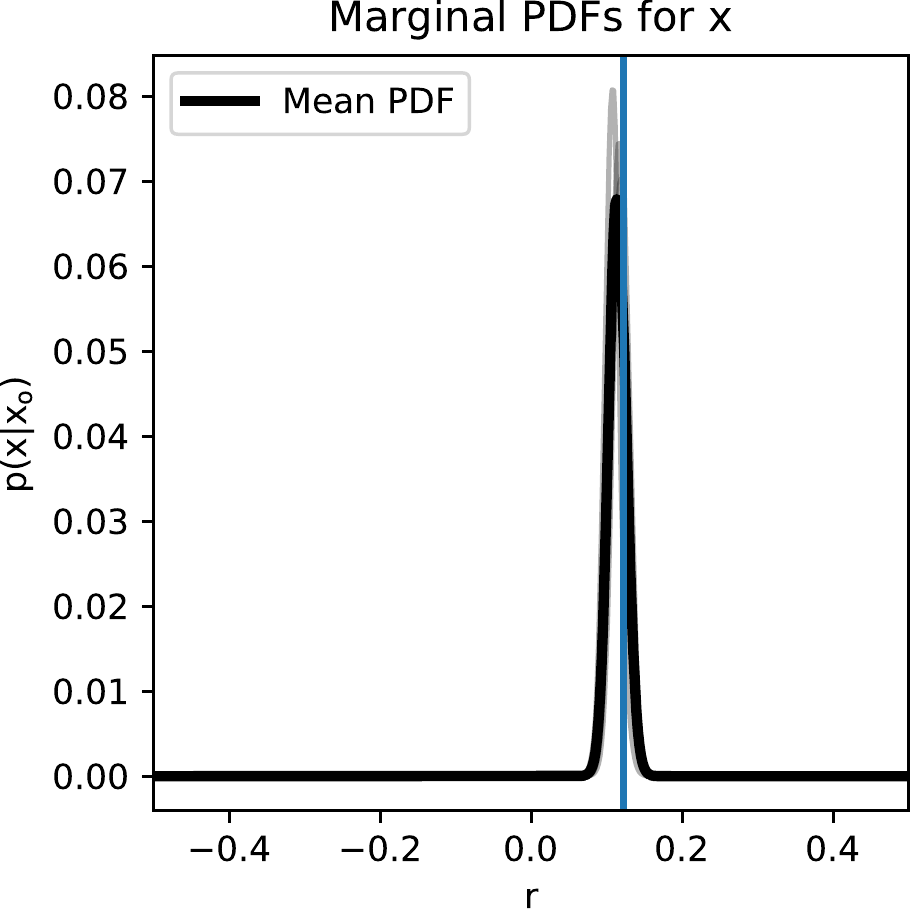}
    \includegraphics[height=.3\linewidth]{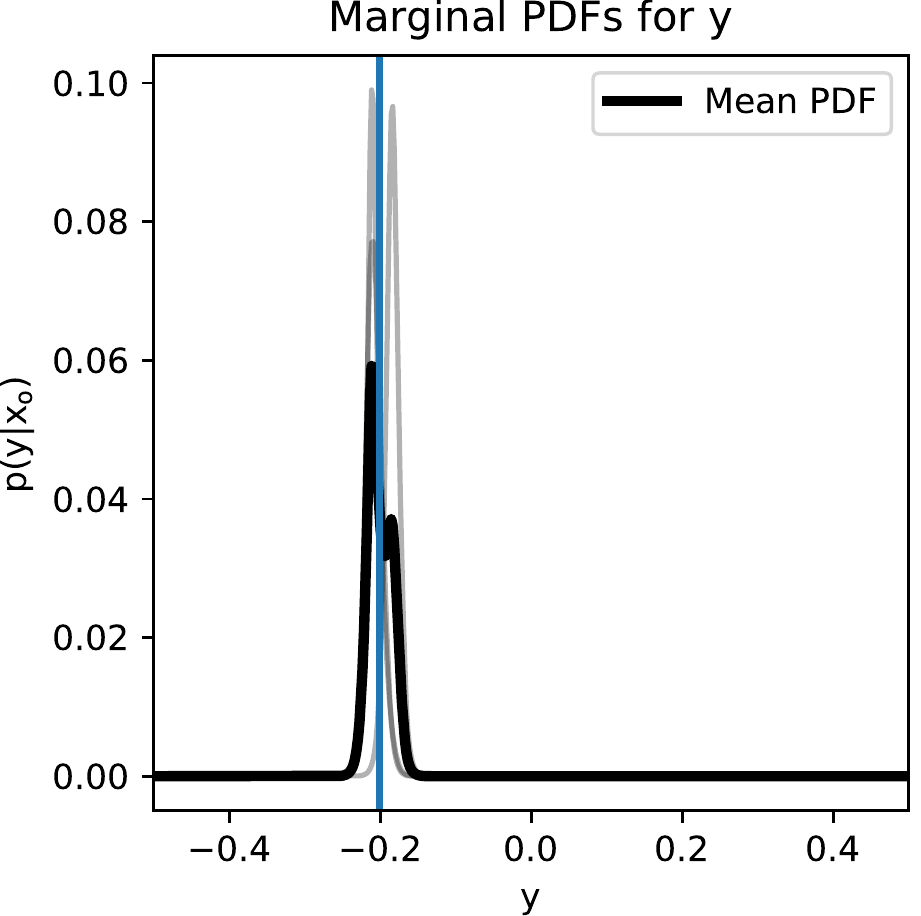}
    \includegraphics[height=.3\linewidth]{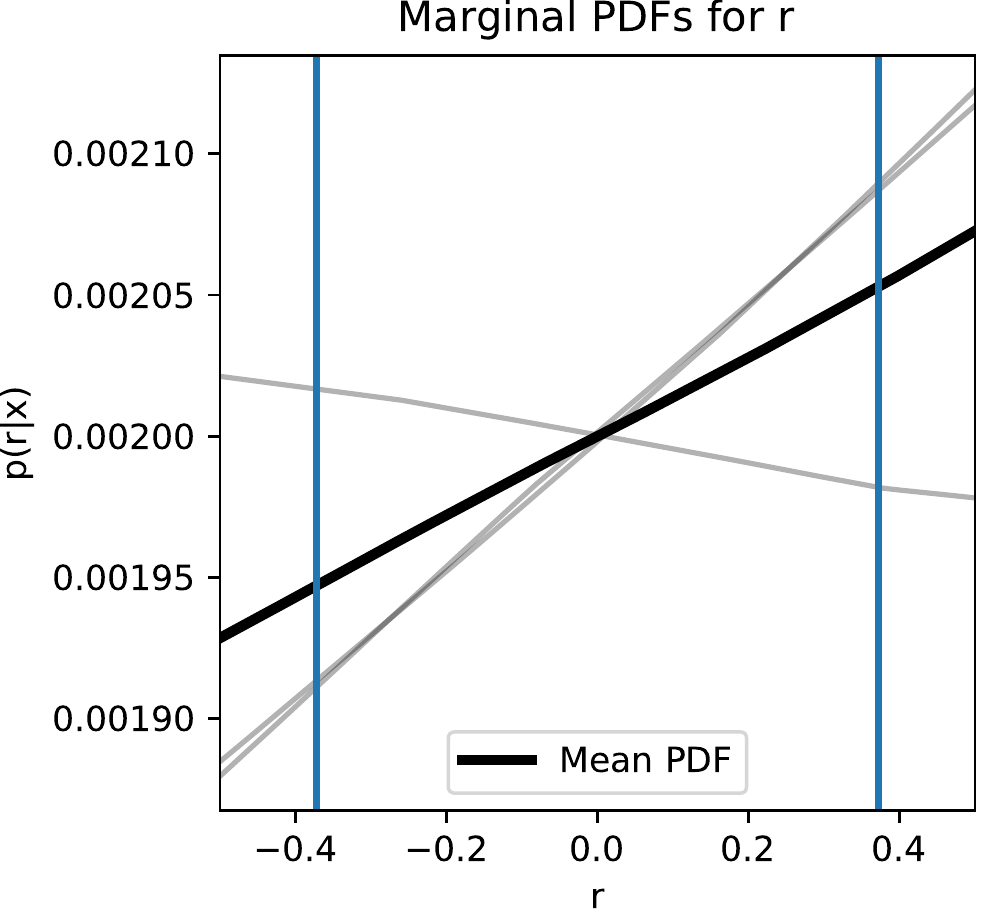}
    \caption{Fully connected ratio estimator}
    \vspace{.2cm}
  \end{subfigure}
  \begin{subfigure}{.8\linewidth}
    \centering
    \includegraphics[height=.3\linewidth]{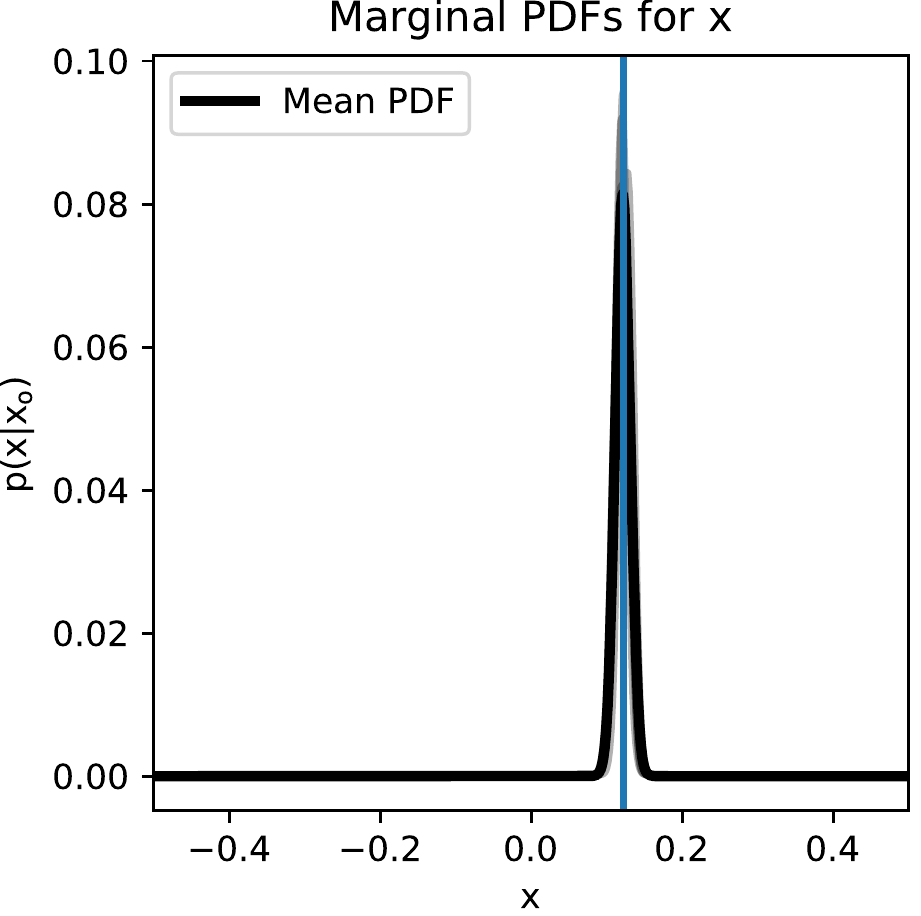}
    \includegraphics[height=.3\linewidth]{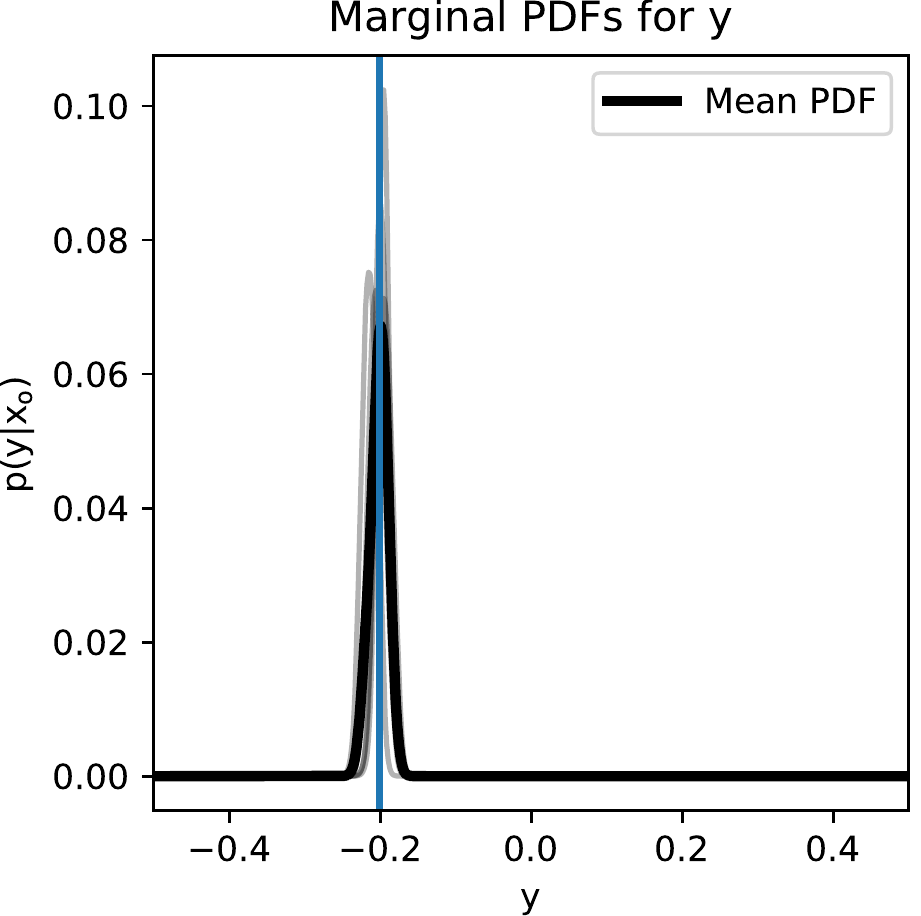}
    \includegraphics[height=.3\linewidth]{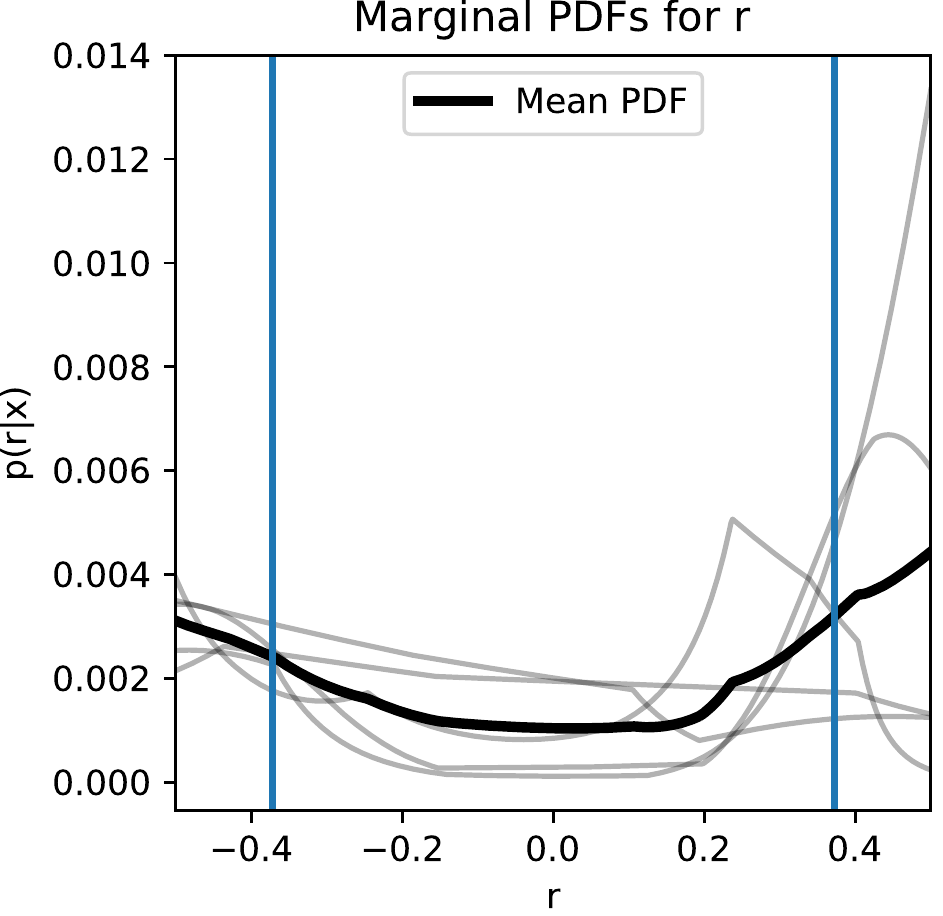}
    \caption{\textsc{lenet} ratio estimator}
    \vspace{.2cm}
  \end{subfigure}
  \begin{subfigure}{.8\linewidth}
    \centering
    \includegraphics[height=.3\linewidth]{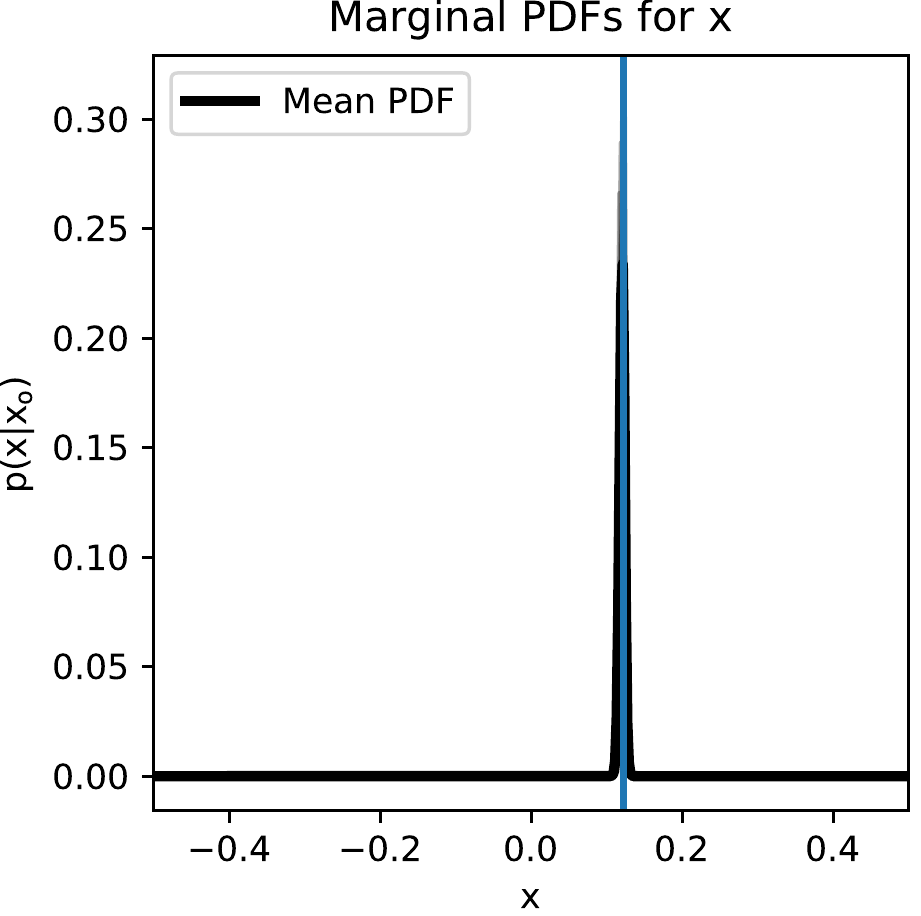}
    \includegraphics[height=.3\linewidth]{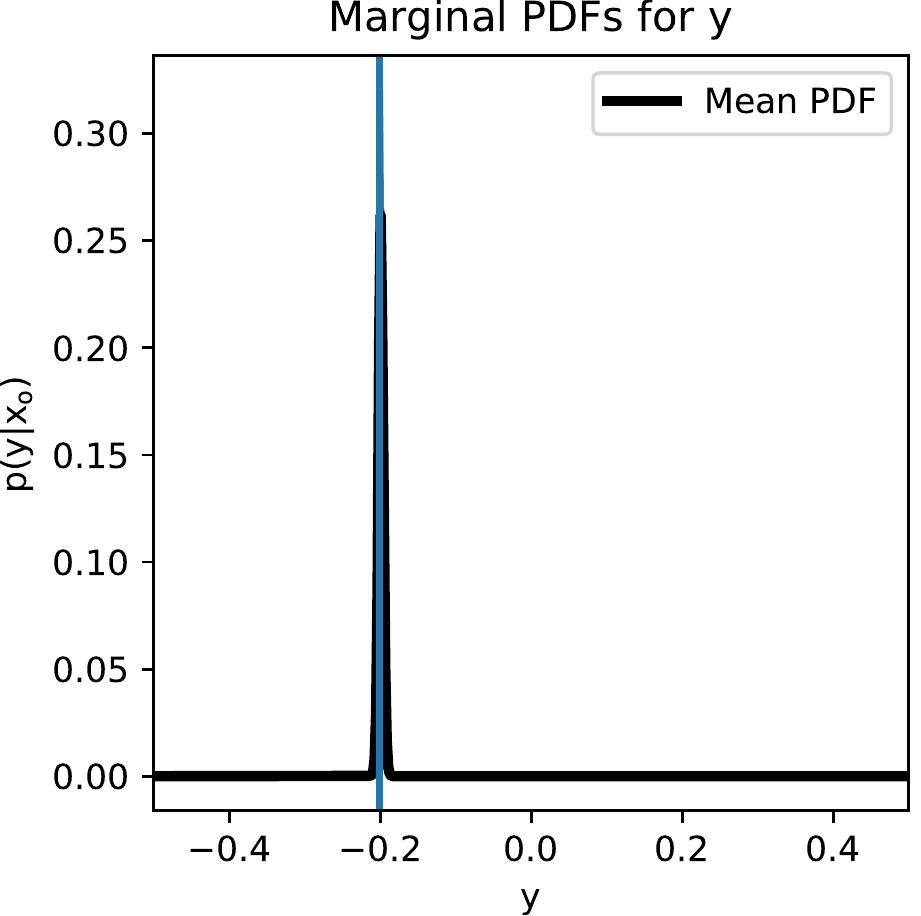}
    \includegraphics[height=.3\linewidth]{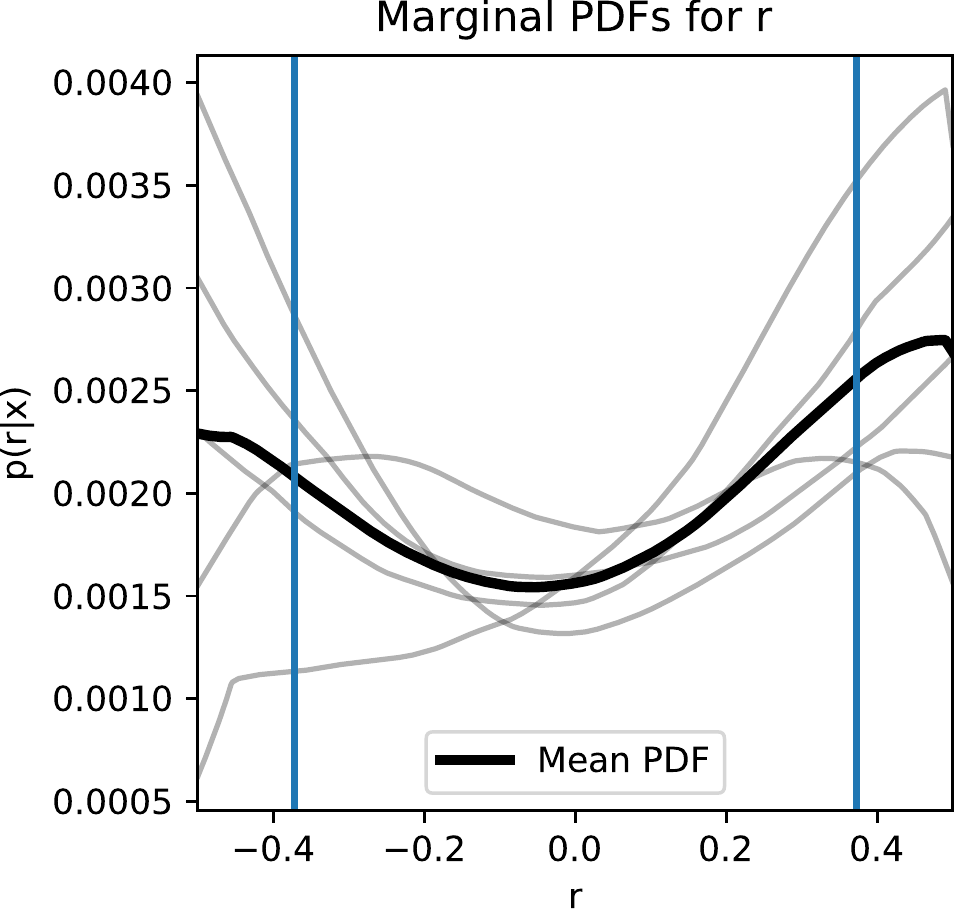}
    \caption{\textsc{resnet-18} ratio estimator}
  \end{subfigure}
  \caption{Marginals of the posteriors for the every ratio estimator architecture. The bold dark line shows the mean \textsc{pdf}, while a gray line shows the \textsc{pdf} of a single ratio estimator. The variance for the parameter $r$ shows that the ratio estimators were not able to properly estimate the ratio $r(\bx\vert\btheta)$. This indicates an issue with the capacity, as the other parameters are properly estimated. Otherwise, the training hyperparameters could be at fault.}
  \label{fig:circle_marginals}
\end{figure}

\begin{figure}
  \centering
  \begin{subfigure}{.8\linewidth}
    \centering
    \includegraphics[height=.3\linewidth]{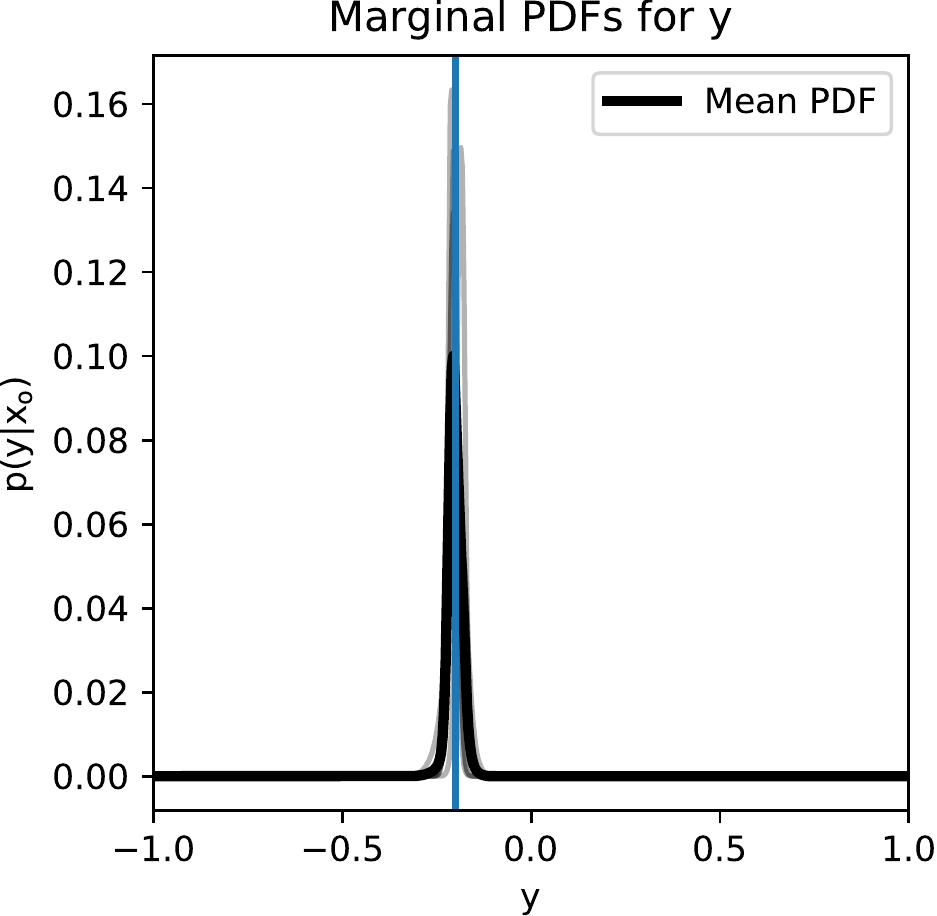}
    \includegraphics[height=.3\linewidth]{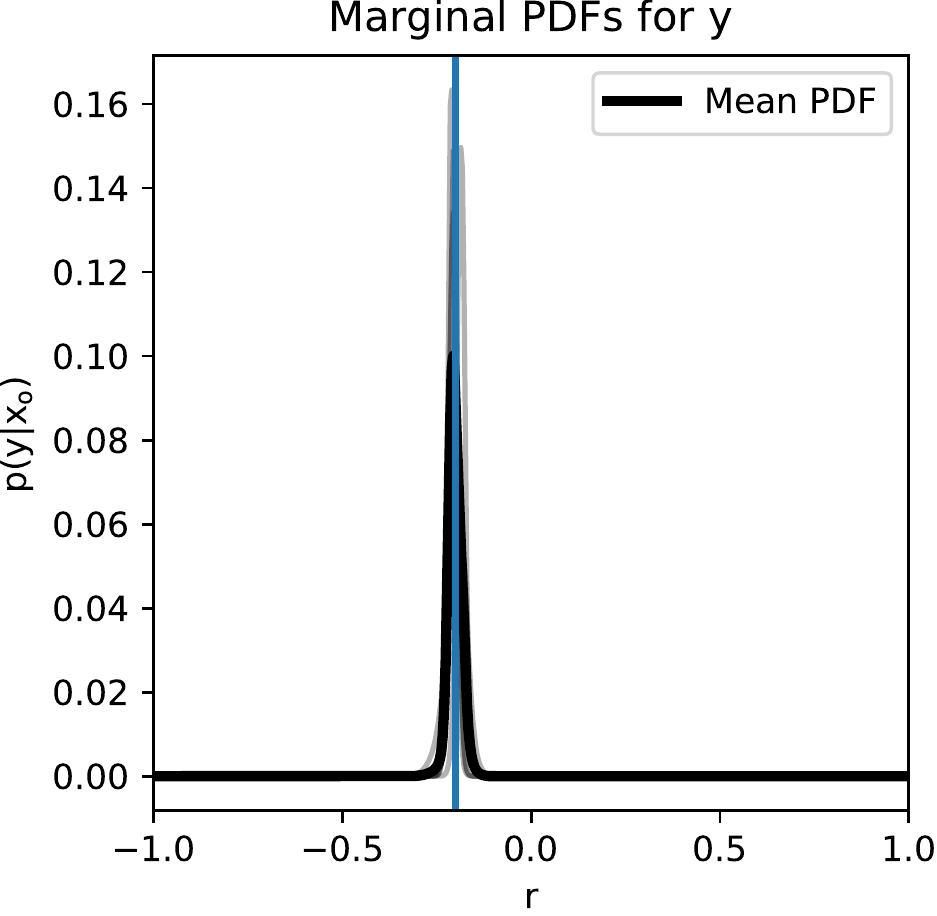}
    \includegraphics[height=.3\linewidth]{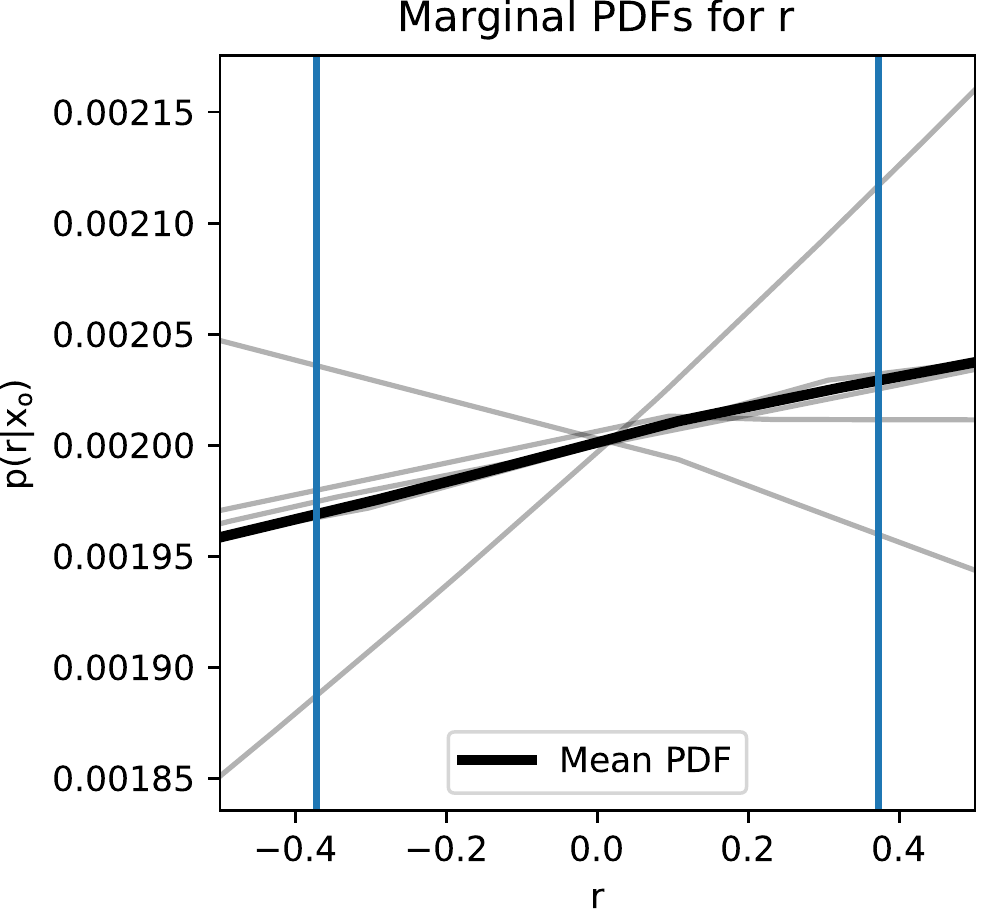}
    \caption{Fully connected ratio estimator}
    \vspace{.2cm}
  \end{subfigure}
  \begin{subfigure}{.8\linewidth}
    \centering
    \includegraphics[height=.3\linewidth]{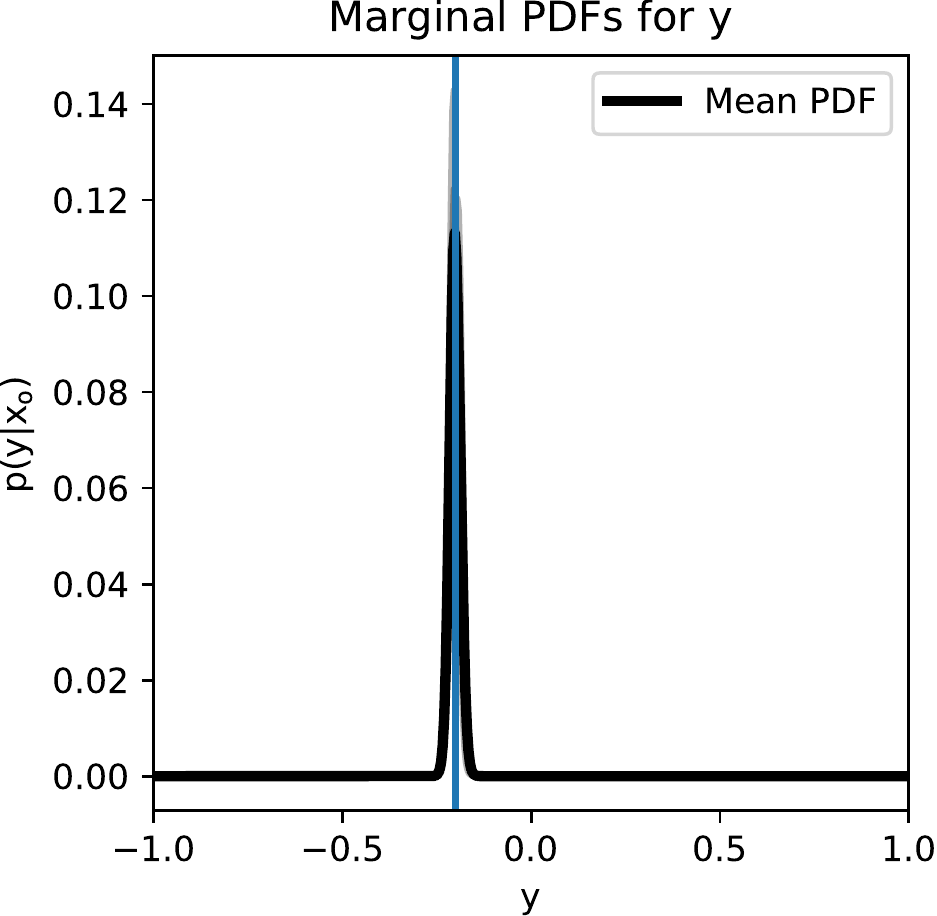}
    \includegraphics[height=.3\linewidth]{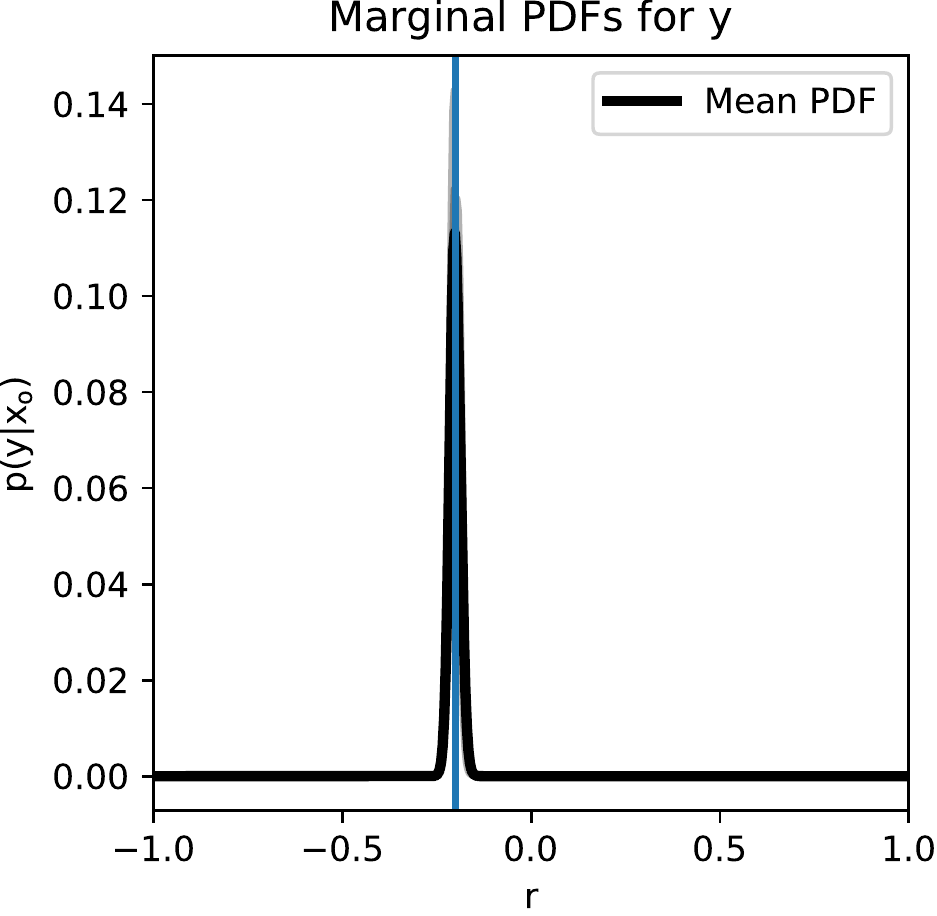}
    \includegraphics[height=.3\linewidth]{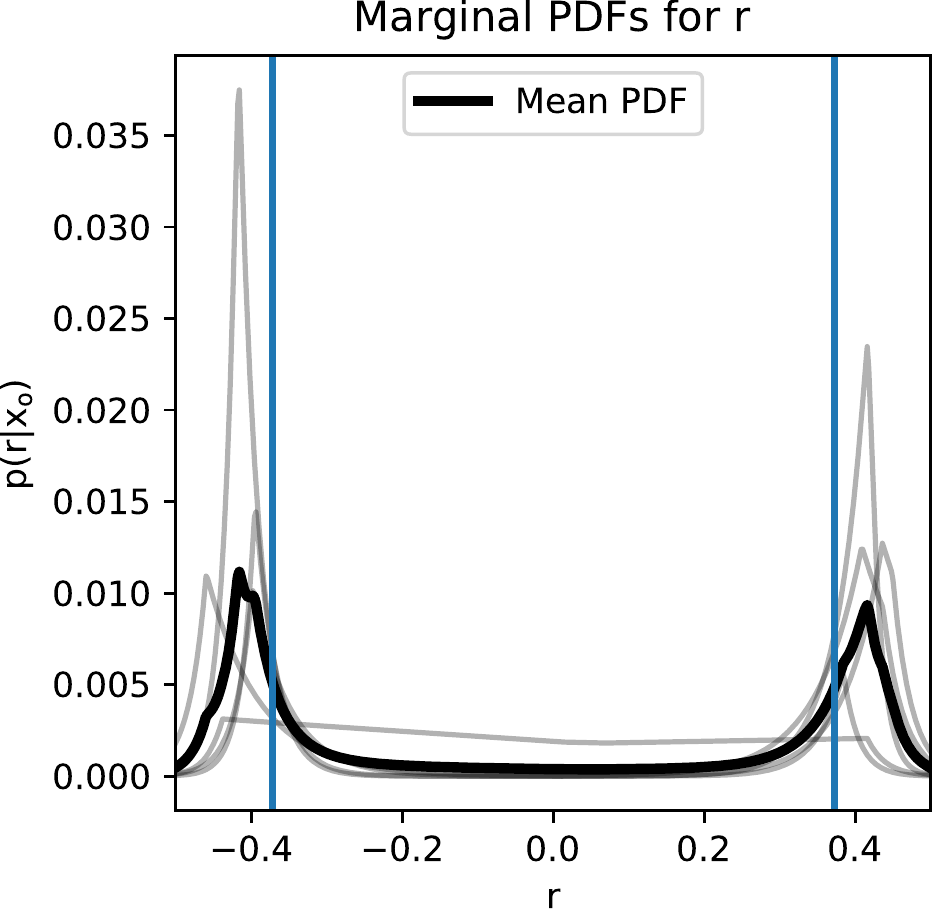}
    \caption{\textsc{lenet} ratio estimator}
    \vspace{.2cm}
  \end{subfigure}
  \begin{subfigure}{.8\linewidth}
    \centering
    \includegraphics[height=.3\linewidth]{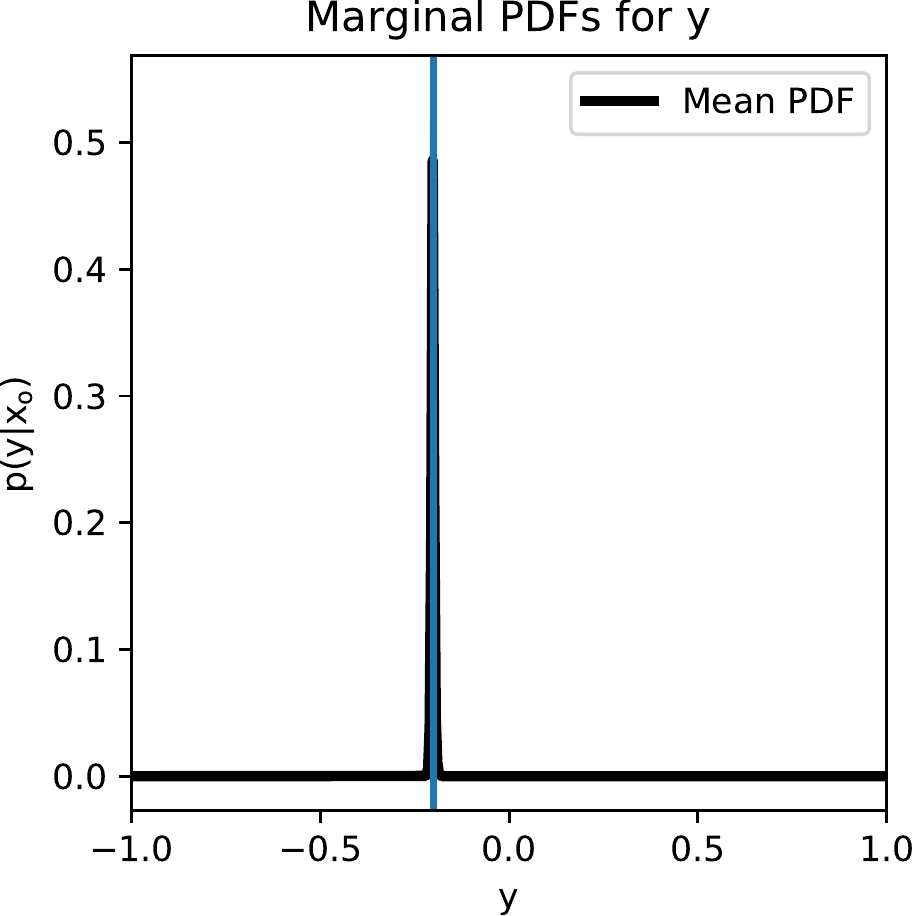}
    \includegraphics[height=.3\linewidth]{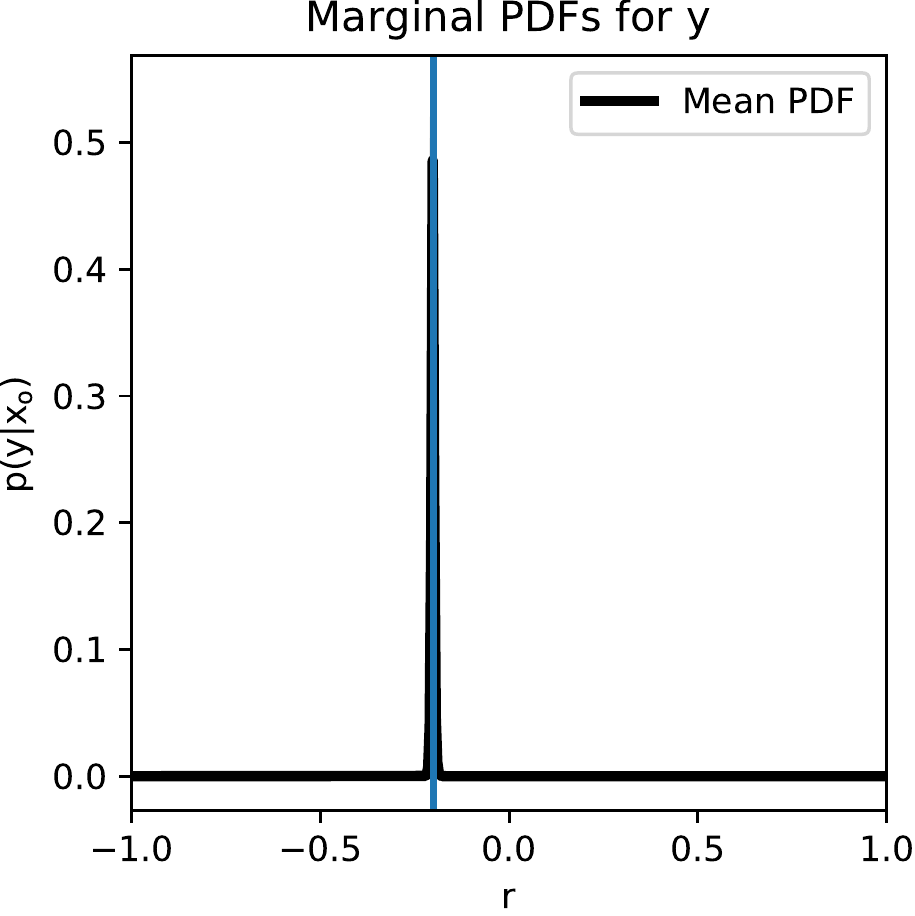}
    \includegraphics[height=.3\linewidth]{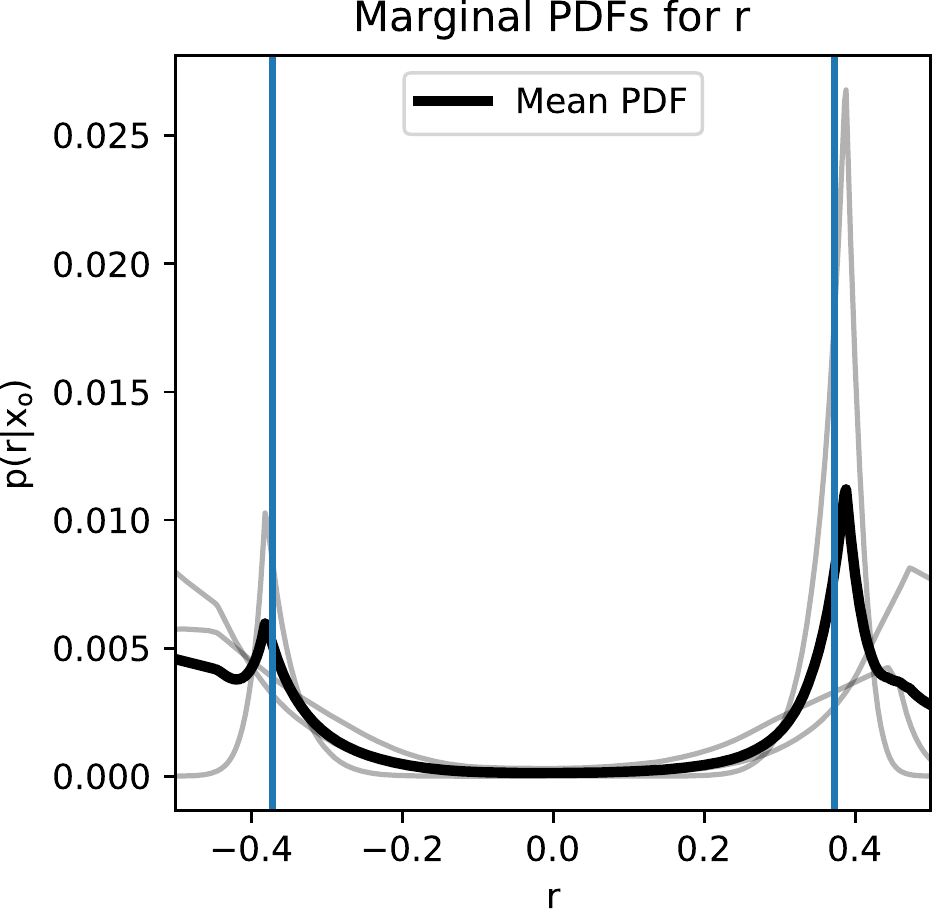}
    \caption{\textsc{resnet-18} ratio estimator}
  \end{subfigure}
  \caption{Marginals of the posteriors for every ratio estimator architecture using \textsc{relu} activations. The bold dark line shows the mean \textsc{pdf}, while a gray line shows the \textsc{pdf} of a single ratio estimator.}
  \label{fig:circle_marginals_relu}
\end{figure}

As noted above, the deterministic nature of the simulation model should generate posteriors which are sharp. We compute $\nabla_{\btheta} \bs(\bx,\btheta)$ and $\nabla_{\btheta}\log\hat{r}(\bx\vert\btheta)$ to investigate how the gradients behave in $p(\btheta)$. We expect the posteriors to be unimodal for the $x$ and $y$ parameters. As a result, the gradient field should converge to the generating parameter $\btheta^*$. Figure~\ref{fig:circle_field_maps} shows the gradient fields across the different architectures. All use the same observation. The left-hand side of the figure shows $\nabla_{\btheta} \bs(\bx,\btheta)$, and the right-hand side $\nabla_{\btheta}\log\hat{r}(\bx\vert\btheta)$. The saturation of the sigmoid operation is clearly visible as the gradients tend to 0. This is not the case for $\log\hat{r}(\bx\vert\btheta)$, demonstrating the effectiveness of the improvements put forward in Section~\ref{sec:improve_approx_lr}. This behavior is preferable for
Hamiltonian Monte Carlo, which relies on $\nabla_{\btheta}\log p(\bx\vert\btheta)$ to generate proposal states.
As expected, the gradients show that the sharpest posterior was obtained by the \textsc{resnet} estimator. However, the estimation of radius $r$ seems to be problematic. The variance of the \textsc{pdf} among the ratio estimators indicates that ratio estimators are not sufficient to accurately approximate the radius. While the general structure is present in \textsc{resnet-18}, the posterior for $r$ is not sharp. A strategy to resolve this would be to increase the capacity of the neural network by adding more parameters (weights), or by modifying the architecture to exploit some structure in the data (e.g., a \textsc{lstm} for time-series). Alternatively, other activation functions could be explored. Experiments indicate that \textsc{elu}~\cite{elu} and \textsc{selu}~\cite{selu} activation functions are good initial choices. For sharp posteriors, we found that \textsc{relu} activations worked best, as demonstrated in Figure~\ref{fig:circle_marginals_relu} and Figure~\ref{fig:circle_marginals} (which uses \textsc{selu} activations). Interestingly, even though the capacity is insufficient to capture all parameters, it seems that the solution is always included (the ratio estimator is not able to minimize the loss by excluding observations drawn from the generating parameter). This is a desirable property as true model parameters are not excluded, which could be beneficial in a Bayesian filtering setting.

To conclude, the amortization of our ratio estimator requires sufficient representational power to accurately approximate $r(\bx\vert\btheta)$. The complexity of the task at hand determines whether the ratio estimator is able to exploit some structure in observations $\bx$ and model parameters $\btheta$, thereby potentially reducing the necessary amount of parameters.


\end{document}